%% file: sample-sigconf.tex
\documentclass[conference]{IEEEtran}

\usepackage[utf8]{inputenc}
\usepackage[T1]{fontenc}

\usepackage{cite}
\usepackage{amsmath,amssymb}
\usepackage{graphicx}
\usepackage{booktabs}
\usepackage{multirow}
\usepackage{xcolor}
\usepackage{eso-pic}

\usepackage{tikz}
\usetikzlibrary{arrows.meta,positioning,calc,fit,backgrounds}
\usepackage{pgfplots}
\usepgfplotslibrary{groupplots,fillbetween}
\pgfplotsset{compat=1.18}
\definecolor{cofacolor}{HTML}{A23B72}%
\definecolor{selfcolor}{HTML}{6B6B6B}
\usepackage{colortbl}
\usepackage{algorithm}
\usepackage{algorithmic}%
\definecolor{decoshade}{HTML}{F6ECF2}%
\input{figures/palette}\usepackage{url}

\begin{document}
\AddToShipoutPictureFG*{\AtPageLowerLeft{\raisebox{0.42in}{\makebox[\paperwidth]{\parbox{\textwidth}{\centering\scriptsize This work has been submitted to the IEEE for possible publication. Copyright may be transferred without notice, after which this version may no longer be accessible.}}}}}

\title{DeCo: Decoupled Traffic State Modeling with Cross-Variable Coordination}

\author{\IEEEauthorblockN{Du Yin\textsuperscript{1,2,5}, Lindong Xie\textsuperscript{3,1}, Xiachong Lin\textsuperscript{1}, Ruiyi Yang\textsuperscript{1,5}, Yifan Duan\textsuperscript{1}, Hao Xue\textsuperscript{1,2,4,5}, Flora Salim\textsuperscript{1,5}}
\IEEEauthorblockA{\textsuperscript{1}University of New South Wales, Sydney, Australia\\
\textsuperscript{2}Hong Kong University of Science and Technology (Guangzhou), China\\
\textsuperscript{3}Hong Kong Polytechnic University, Hong Kong SAR, China\\
\textsuperscript{4}Hong Kong University of Science and Technology, Hong Kong SAR, China\\
\textsuperscript{5}DataCruiser, Sydney, Australia\\
\{du.yin, yifan.duan3, flora.salim\}@unsw.edu.au, \{dawn.lin, ruiyi.yang\}@student.unsw.edu.au,\\
lindong.xie@connect.polyu.hk, haoxue@hkust-gz.edu.cn}}

\maketitle

\input{sections/abstract_intro}

\input{sections/related_data}

\input{sections/method}

\input{sections/experiments}

\input{sections/conclusion}

\section*{Acknowledgments}
This work was supported by the ARC Centre of Excellence for Automated Decision-Making and Society (CE200100005). We acknowledge the resources and services provided by the National Computational Infrastructure (NCI), which is supported by the Australian Government. Experiments were performed on the UNSW Compute Cluster Katana (DOI: 10.26190/669X-A286). This research was also partially supported by the ARC Training Centre for Whole Life Design of Carbon Neutral Infrastructure (IC230100015).

Anthropic's Claude and OpenAI's Codex assisted with experiment code and execution, data analysis, and figures and tables (Sections IV--V). Claude polished drafts across Sections I--VI, and Codex assisted with numerical and reference checks and manuscript revisions. OpenAI's ChatGPT was also used to discuss study design. The authors directed the work and take responsibility for the submitted content.

\bibliographystyle{IEEEtran}
\bibliography{references}

\end{document}

%% file: figures/palette.tex
\definecolor{ds03}{HTML}{1B4F9C}
\definecolor{ds04}{HTML}{B8860B}
\definecolor{ds07}{HTML}{009E73}
\definecolor{ds08}{HTML}{CC3311}

%% file: sections/abstract_intro.tex
\begin{abstract}
Accurate forecasts of traffic flow, occupancy, and speed are essential for congestion management and travel planning. These variables exhibit distinct spatiotemporal patterns, and their coupled dynamics characterize the evolution of traffic conditions, such as the onset and dissipation of congestion. Existing methods often either neglect this coupling or struggle to capture the distinct dynamics of all three variables within a unified model. In this paper, we propose DeCo, a \underline{de}coupled traffic state modeling framework with cross-variable \underline{co}ordination for forecasting flow, occupancy, and speed. Specifically, DeCo first trains a dedicated forecaster for each variable to specialize in its unique spatiotemporal dynamics. It then estimates a residual correction for each forecast by jointly leveraging the predicted trajectories and recent observations of all three variables, thereby exploiting cross-variable coupling and mitigating systematic prediction errors. We extend the flow-only PeMS03 and PeMS07 benchmarks to all three variables on their original sensors and release the aligned data. Experiments on PeMS03, PeMS04, PeMS07, and PeMS08 show that DeCo achieves the lowest mean absolute error for every variable on every dataset, up to 4.9\% below RAGC, the strongest baseline. Code is available at \url{https://github.com/lixus7/DeCo}.
\end{abstract}

\begin{IEEEkeywords}
Traffic forecasting, spatiotemporal sensor data, multivariate time series, benchmark datasets, cross-variable coordination
\end{IEEEkeywords}

\section{Introduction}
\label{sec:introduction}
\label{sec:intro}

Traffic state forecasting underpins proactive traffic management, allowing operators to anticipate congestion and intervene before it builds~\cite{yin2022trafficreview}. Flow, occupancy, and speed are the three key traffic state variables, characterizing traffic throughput, road utilization, and travel velocity. Their joint variations reflect traffic conditions and their evolution, including transitions between free-flowing and congested regimes and the formation and dissipation of congestion~\cite{li2011fundamental}. Road sensor networks report the three variables together, yet two of the four standard PeMS benchmarks distribute flow only.

Existing traffic state forecasting methods can be broadly divided into three paradigms according to how they utilize multiple traffic variables. The first paradigm, single-variable forecasting, models each variable in isolation, as exemplified by DCRNN~\cite{li2018dcrnn} and STGCN~\cite{yu2018stgcn}. Such methods capture the spatiotemporal dynamics of an individual variable but ignore the information carried by the other variables, which is often essential for identifying the prevailing traffic state. An identical flow, for example, can be observed under both free-flowing and congested conditions~\cite{li2011fundamental}, and only occupancy or speed can distinguish the two. The second paradigm, auxiliary-input forecasting, addresses this limitation by feeding other variables' histories into a target variable's forecaster, as in DMSTGCN~\cite{han2021dynamic}. These additional inputs help distinguish traffic states, and each model remains dedicated to its own target. The three variables, however, are forecast separately. Each forecaster sees the other variables only through their past, without access to the other forecasters' outputs. The third paradigm, joint-output forecasting, removes this separation by forecasting all variables together through shared representations and cross-variable interactions, as in AutoSTL~\cite{zhang2023autostl} and DG-STMTL~\cite{dgstmtl2025}. Sharing lets the variables inform one another but also exposes them to interference from their distinct dynamics and objectives, leading to uneven gains across variables~\cite{dgstmtl2025}.

We compare single-variable, auxiliary-input, and joint-output versions of the same forecasting architectures, such as STAEformer~\cite{liu2023staeformer}, HimNet~\cite{dong2024himnet}, and D$^2$STGNN~\cite{shao2022d2stgnn}, under a common protocol. Auxiliary inputs reduce the test mean absolute error (MAE) of the single-variable forecasts by a median of $0.7\%$, whereas joint outputs increase it by a median of $1.2\%$ (Fig.~\ref{fig:intro}(c)). A broader comparison of 20 architectures on PeMS04 and PeMS08 points the same way. The joint-output form is the most accurate of the trained forms in fewer than one tenth of the cases (Table~\ref{tab:extra}). These results favor a dedicated model for each variable, which raises the question of how the three forecasts can still benefit from one another.

\input{figures/intro_figure}

To coordinate these forecasts, we propose DeCo, a decoupled traffic state modeling framework with cross-variable coordination. Specifically, DeCo first trains a dedicated forecaster for each of flow, occupancy, and speed, and then applies a coordination module that estimates a residual correction for each forecast (Fig.~\ref{fig:intro}(b)). The module takes as input the predicted trajectories and recent observations of all three variables. Because these trajectories span the same future horizon, the module can adjust each forecast according to the anticipated evolution of the other variables, not only their past observations. The forecasters are kept fixed while the module is trained. Each specializes in its own variable, and the module learns only how cross-variable information should adjust the forecasts. Following stacked generalization~\cite{wolpert1992stacked}, the module is trained on out-of-fold forecasts from $K$ fold forecasters per variable, so that it learns from errors resembling those the forecasters make on unseen data. At inference, DeCo averages the corrected forecasts of the fold forecasters and blends them with an early-fusion forecaster that reads all three variables, so that input-level fusion, which our comparison found helpful, complements forecast-level coordination. Existing correction methods for traffic forecasters, such as ResCAL~\cite{kim2022rescal} and RiskTraf~\cite{wang2026risktraf}, correct a single variable from its own past errors or from the histories of the other variables, whereas DeCo corrects all three variables from the forecasts of all of them. With a single forecaster per variable (DeCo-S), coordination alone reduces the MAE of these forecasters by a median of $2.1\%$ (Fig.~\ref{fig:intro}(c)). The full system attains the lowest MAE for every variable on every dataset, up to $4.9\%$ below RAGC, the strongest baseline, and the module also applies to a published forecaster, RAGC~\cite{ragc}.

The main contributions are as follows:
\begin{itemize}
\item We propose DeCo, a decoupled traffic state modeling framework with cross-variable coordination for forecasting flow, occupancy, and speed. DeCo separates variable-specific forecasting from cross-variable coordination, so that each forecast benefits from the other variables without forecasting them jointly.
\item We design a lightweight coordination module, with 0.11 million parameters, that corrects each forecast from the predicted trajectories and recent observations of all three variables. It is trained on out-of-fold forecasts from $K$ fold forecasters per variable, cross-fitted over contiguous time blocks.
\item We extend the flow-only PeMS03 and PeMS07 benchmarks with aligned occupancy and speed on their original sensors, matching the benchmark flow in 99.956\% of PeMS03 entries and in every PeMS07 entry. We release the aligned data with validity masks, station identifiers, and source checksums, so that all four standard PeMS benchmarks support three-variable forecasting.
\item On all four PeMS benchmarks, DeCo achieves the lowest MAE for every variable and dataset, up to $4.9\%$ below RAGC, the strongest baseline. The coordination module lowers the MAE on every dataset and variable, and with four fold forecasters, DeCo trains faster than D$^2$STGNN on PeMS04 and PeMS08 while remaining below every baseline (Fig.~\ref{fig:cost}).
\end{itemize}

%% file: figures/intro_figure.tex
\begin{figure*}[t]
\input{figures/intro_top}\par
\vspace{7pt}
\caption{Multivariate traffic state forecasting. (a)~Flow (F), occupancy (O), and speed (S) at $N$ sensors are forecast $H$ steps ahead from the past $P$ steps. Existing methods forecast each variable alone, read the other variables as auxiliary input, or predict all variables jointly. Filled circles are histories and open circles forecasts. (b)~DeCo trains dedicated forecasters $f_q$ for the three variables, and its coordination module $g$ corrects their forecasts. Snowflakes mark the forecasters, which stay fixed while $g$ is trained. (c)~MAE reduction over the corresponding single-variable forecaster, with each group under the form it measures. Dashed lines and numbers give medians. (d)~Speed MAE reduction by traffic state on each dataset. The spokes of a sector are PeMS03, PeMS04, PeMS07, and PeMS08. Panels (c) and (d) use DeCo with one forecaster per variable (DeCo-S).}
\label{fig:intro}
\vspace{13.75pt}
\end{figure*}
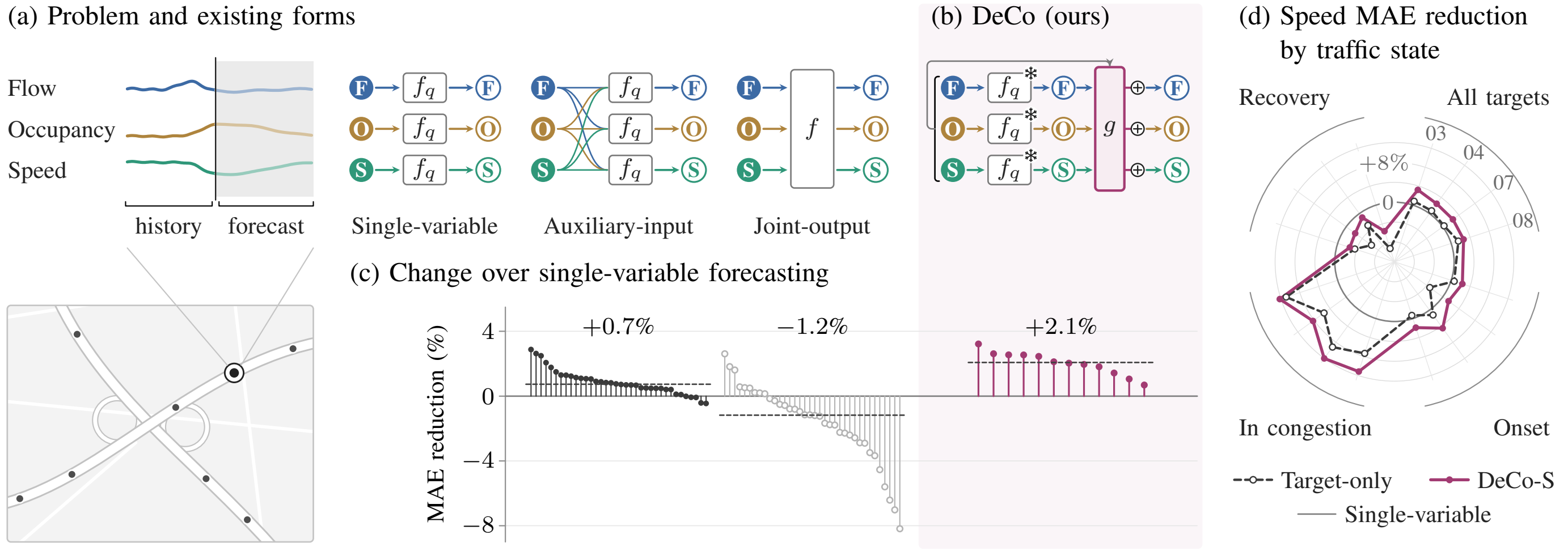

%% file: figures/intro_top.tex
\definecolor{vF}{HTML}{3C6AA4}\definecolor{vO}{HTML}{D4A14A}\definecolor{vS}{HTML}{2F9779}\definecolor{decoAccent}{HTML}{A23B72}\definecolor{inkMain}{HTML}{222222}\definecolor{inkMuted}{HTML}{6B6B6B}\definecolor{gD}{HTML}{4D4D4D}\definecolor{gL}{HTML}{A6A6A6}\definecolor{stroke}{HTML}{7A7A7A}\definecolor{rule}{HTML}{C4C4C4}\definecolor{paper}{HTML}{F3F3F3}\definecolor{zone}{HTML}{EBEBEB}\definecolor{mapCase}{HTML}{C2C2C2}\definecolor{gHist}{HTML}{3A3A3A}\definecolor{gFut}{HTML}{9A9A9A}
\definecolor{dTO}{HTML}{3A3A3A}%
\definecolor{dSV}{HTML}{B4B4B4}%
\centering
\begin{tikzpicture}[x=1cm, y=1cm, font=\footnotesize, line cap=round, line join=round]
\useasboundingbox (0,0.24) rectangle (18.1,6.5);
\fill[decoAccent!5, rounded corners=1.5pt] (10.63,0.12) rectangle (13.94,6.48);
\fill[vF] (11.03,5.5) circle (0.14);\node[font=\scriptsize\bfseries, text=white, inner sep=0] at (11.03,5.5) {F};
\draw[vF, line width=0.7pt, -{Stealth[length=2.6pt,width=2.4pt]}] (11.2,5.5) -- (11.4,5.5);
\draw[draw=stroke, fill=white, line width=0.5pt, rounded corners=1.2pt] (11.44,5.33) rectangle (11.94,5.67);
\node[font=\footnotesize, text=inkMain, inner sep=0] at (11.69,5.5) {$f_q$};
\fill[white] (11.94,5.67) circle (0.09);
\begin{scope}[shift={(11.94,5.67)}, draw=inkMain, line width=0.36pt, line cap=round]
\draw (90:0.072) -- (270:0.072);
\draw (90:0.043) -- ++(130:0.023) (90:0.043) -- ++(50:0.023);
\draw (270:0.043) -- ++(310:0.023) (270:0.043) -- ++(230:0.023);
\draw (30:0.072) -- (210:0.072);
\draw (30:0.043) -- ++(70:0.023) (30:0.043) -- ++(-10:0.023);
\draw (210:0.043) -- ++(250:0.023) (210:0.043) -- ++(170:0.023);
\draw (-30:0.072) -- (150:0.072);
\draw (-30:0.043) -- ++(10:0.023) (-30:0.043) -- ++(-70:0.023);
\draw (150:0.043) -- ++(190:0.023) (150:0.043) -- ++(110:0.023);
\end{scope}
\draw[vF, line width=0.7pt, -{Stealth[length=2.6pt,width=2.4pt]}] (11.98,5.5) -- (12.16,5.5);
\draw[vF, fill=white, line width=0.6pt] (12.32,5.5) circle (0.13);\node[font=\scriptsize\bfseries, text=vF, inner sep=0] at (12.32,5.5) {F};
\draw[vF, line width=0.7pt, -{Stealth[length=2.6pt,width=2.4pt]}] (12.48,5.5) -- (12.66,5.5);
\fill[vO!82!black] (11.03,5.02) circle (0.14);\node[font=\scriptsize\bfseries, text=white, inner sep=0] at (11.03,5.02) {O};
\draw[vO!82!black, line width=0.7pt, -{Stealth[length=2.6pt,width=2.4pt]}] (11.2,5.02) -- (11.4,5.02);
\draw[draw=stroke, fill=white, line width=0.5pt, rounded corners=1.2pt] (11.44,4.85) rectangle (11.94,5.19);
\node[font=\footnotesize, text=inkMain, inner sep=0] at (11.69,5.02) {$f_q$};
\fill[white] (11.94,5.19) circle (0.09);
\begin{scope}[shift={(11.94,5.19)}, draw=inkMain, line width=0.36pt, line cap=round]
\draw (90:0.072) -- (270:0.072);
\draw (90:0.043) -- ++(130:0.023) (90:0.043) -- ++(50:0.023);
\draw (270:0.043) -- ++(310:0.023) (270:0.043) -- ++(230:0.023);
\draw (30:0.072) -- (210:0.072);
\draw (30:0.043) -- ++(70:0.023) (30:0.043) -- ++(-10:0.023);
\draw (210:0.043) -- ++(250:0.023) (210:0.043) -- ++(170:0.023);
\draw (-30:0.072) -- (150:0.072);
\draw (-30:0.043) -- ++(10:0.023) (-30:0.043) -- ++(-70:0.023);
\draw (150:0.043) -- ++(190:0.023) (150:0.043) -- ++(110:0.023);
\end{scope}
\draw[vO!82!black, line width=0.7pt, -{Stealth[length=2.6pt,width=2.4pt]}] (11.98,5.02) -- (12.16,5.02);
\draw[vO!82!black, fill=white, line width=0.6pt] (12.32,5.02) circle (0.13);\node[font=\scriptsize\bfseries, text=vO!82!black, inner sep=0] at (12.32,5.02) {O};
\draw[vO!82!black, line width=0.7pt, -{Stealth[length=2.6pt,width=2.4pt]}] (12.48,5.02) -- (12.66,5.02);
\fill[vS] (11.03,4.54) circle (0.14);\node[font=\scriptsize\bfseries, text=white, inner sep=0] at (11.03,4.54) {S};
\draw[vS, line width=0.7pt, -{Stealth[length=2.6pt,width=2.4pt]}] (11.2,4.54) -- (11.4,4.54);
\draw[draw=stroke, fill=white, line width=0.5pt, rounded corners=1.2pt] (11.44,4.37) rectangle (11.94,4.71);
\node[font=\footnotesize, text=inkMain, inner sep=0] at (11.69,4.54) {$f_q$};
\fill[white] (11.94,4.71) circle (0.09);
\begin{scope}[shift={(11.94,4.71)}, draw=inkMain, line width=0.36pt, line cap=round]
\draw (90:0.072) -- (270:0.072);
\draw (90:0.043) -- ++(130:0.023) (90:0.043) -- ++(50:0.023);
\draw (270:0.043) -- ++(310:0.023) (270:0.043) -- ++(230:0.023);
\draw (30:0.072) -- (210:0.072);
\draw (30:0.043) -- ++(70:0.023) (30:0.043) -- ++(-10:0.023);
\draw (210:0.043) -- ++(250:0.023) (210:0.043) -- ++(170:0.023);
\draw (-30:0.072) -- (150:0.072);
\draw (-30:0.043) -- ++(10:0.023) (-30:0.043) -- ++(-70:0.023);
\draw (150:0.043) -- ++(190:0.023) (150:0.043) -- ++(110:0.023);
\end{scope}
\draw[vS, line width=0.7pt, -{Stealth[length=2.6pt,width=2.4pt]}] (11.98,4.54) -- (12.16,4.54);
\draw[vS, fill=white, line width=0.6pt] (12.32,4.54) circle (0.13);\node[font=\scriptsize\bfseries, text=vS, inner sep=0] at (12.32,4.54) {S};
\draw[vS, line width=0.7pt, -{Stealth[length=2.6pt,width=2.4pt]}] (12.48,4.54) -- (12.66,4.54);
\draw[draw=decoAccent, fill=decoAccent!9, line width=0.8pt, rounded corners=1.2pt] (12.69,4.3) rectangle (13.03,5.74);\node[font=\footnotesize, text=inkMain, inner sep=0] at (12.86,5.02) {$g$};
\draw[decoAccent, line width=0.7pt] (13.03,5.5) -- (13.11,5.5);
\draw[draw=inkMain, fill=white, line width=0.45pt] (13.185,5.5) circle (0.075);\draw[draw=inkMain, line width=0.45pt] (13.138,5.5) -- (13.231,5.5) (13.185,5.454) -- (13.185,5.546);
\draw[vF, line width=0.7pt, -{Stealth[length=2.6pt,width=2.4pt]}] (13.29,5.5) -- (13.47,5.5);
\draw[vF, fill=white, line width=0.6pt] (13.64,5.5) circle (0.14);\node[font=\scriptsize\bfseries, text=vF, inner sep=0] at (13.64,5.5) {F};
\draw[decoAccent, line width=0.7pt] (13.03,5.02) -- (13.11,5.02);
\draw[draw=inkMain, fill=white, line width=0.45pt] (13.185,5.02) circle (0.075);\draw[draw=inkMain, line width=0.45pt] (13.138,5.02) -- (13.231,5.02) (13.185,4.973) -- (13.185,5.066);
\draw[vO!82!black, line width=0.7pt, -{Stealth[length=2.6pt,width=2.4pt]}] (13.29,5.02) -- (13.47,5.02);
\draw[vO!82!black, fill=white, line width=0.6pt] (13.64,5.02) circle (0.14);\node[font=\scriptsize\bfseries, text=vO!82!black, inner sep=0] at (13.64,5.02) {O};
\draw[decoAccent, line width=0.7pt] (13.03,4.54) -- (13.11,4.54);
\draw[draw=inkMain, fill=white, line width=0.45pt] (13.185,4.54) circle (0.075);\draw[draw=inkMain, line width=0.45pt] (13.138,4.54) -- (13.231,4.54) (13.185,4.494) -- (13.185,4.587);
\draw[vS, line width=0.7pt, -{Stealth[length=2.6pt,width=2.4pt]}] (13.29,4.54) -- (13.47,4.54);
\draw[vS, fill=white, line width=0.6pt] (13.64,4.54) circle (0.14);\node[font=\scriptsize\bfseries, text=vS, inner sep=0] at (13.64,4.54) {S};
\draw[stroke, line width=0.5pt, rounded corners=1pt] (10.87,4.41) -- (10.82,4.41) -- (10.82,5.63) -- (10.87,5.63);
\draw[draw=stroke, line width=0.5pt, -{Stealth[length=2.6pt,width=2.4pt]}, rounded corners=1.5pt] (10.82,5.02) -- (10.73,5.02) -- (10.73,5.81) -- (12.86,5.81) -- (12.86,5.75);
\fill[vF] (4.13,5.5) circle (0.14);\node[font=\scriptsize\bfseries, text=white, inner sep=0] at (4.13,5.5) {F};
\draw[vF, line width=0.7pt, -{Stealth[length=2.6pt,width=2.4pt]}] (4.3,5.5) -- (4.58,5.5);
\draw[draw=stroke, fill=white, line width=0.5pt, rounded corners=1.2pt] (4.62,5.33) rectangle (5.12,5.67);
\node[font=\footnotesize, text=inkMain, inner sep=0] at (4.87,5.5) {$f_q$};
\draw[vF, line width=0.7pt, -{Stealth[length=2.6pt,width=2.4pt]}] (5.16,5.5) -- (5.44,5.5);
\draw[vF, fill=white, line width=0.6pt] (5.61,5.5) circle (0.14);\node[font=\scriptsize\bfseries, text=vF, inner sep=0] at (5.61,5.5) {F};
\fill[vO!82!black] (4.13,5.02) circle (0.14);\node[font=\scriptsize\bfseries, text=white, inner sep=0] at (4.13,5.02) {O};
\draw[vO!82!black, line width=0.7pt, -{Stealth[length=2.6pt,width=2.4pt]}] (4.3,5.02) -- (4.58,5.02);
\draw[draw=stroke, fill=white, line width=0.5pt, rounded corners=1.2pt] (4.62,4.85) rectangle (5.12,5.19);
\node[font=\footnotesize, text=inkMain, inner sep=0] at (4.87,5.02) {$f_q$};
\draw[vO!82!black, line width=0.7pt, -{Stealth[length=2.6pt,width=2.4pt]}] (5.16,5.02) -- (5.44,5.02);
\draw[vO!82!black, fill=white, line width=0.6pt] (5.61,5.02) circle (0.14);\node[font=\scriptsize\bfseries, text=vO!82!black, inner sep=0] at (5.61,5.02) {O};
\fill[vS] (4.13,4.54) circle (0.14);\node[font=\scriptsize\bfseries, text=white, inner sep=0] at (4.13,4.54) {S};
\draw[vS, line width=0.7pt, -{Stealth[length=2.6pt,width=2.4pt]}] (4.3,4.54) -- (4.58,4.54);
\draw[draw=stroke, fill=white, line width=0.5pt, rounded corners=1.2pt] (4.62,4.37) rectangle (5.12,4.71);
\node[font=\footnotesize, text=inkMain, inner sep=0] at (4.87,4.54) {$f_q$};
\draw[vS, line width=0.7pt, -{Stealth[length=2.6pt,width=2.4pt]}] (5.16,4.54) -- (5.44,4.54);
\draw[vS, fill=white, line width=0.6pt] (5.61,4.54) circle (0.14);\node[font=\scriptsize\bfseries, text=vS, inner sep=0] at (5.61,4.54) {S};
\fill[vF] (6.25,5.5) circle (0.14);\node[font=\scriptsize\bfseries, text=white, inner sep=0] at (6.25,5.5) {F};
\fill[vO!82!black] (6.25,5.02) circle (0.14);\node[font=\scriptsize\bfseries, text=white, inner sep=0] at (6.25,5.02) {O};
\fill[vS] (6.25,4.54) circle (0.14);\node[font=\scriptsize\bfseries, text=white, inner sep=0] at (6.25,4.54) {S};
\draw[vF, line width=0.5pt] (6.42,5.5) .. controls (6.722,5.5) and (6.698,5.5) .. (7,5.5);
\draw[vF, line width=0.5pt] (6.42,5.5) .. controls (6.722,5.5) and (6.698,5.02) .. (7,5.02);
\draw[vF, line width=0.5pt] (6.42,5.5) .. controls (6.722,5.5) and (6.698,4.54) .. (7,4.54);
\draw[vO!82!black, line width=0.5pt] (6.42,5.02) .. controls (6.722,5.02) and (6.698,5.5) .. (7,5.5);
\draw[vO!82!black, line width=0.5pt] (6.42,5.02) .. controls (6.722,5.02) and (6.698,5.02) .. (7,5.02);
\draw[vO!82!black, line width=0.5pt] (6.42,5.02) .. controls (6.722,5.02) and (6.698,4.54) .. (7,4.54);
\draw[vS, line width=0.5pt] (6.42,4.54) .. controls (6.722,4.54) and (6.698,5.5) .. (7,5.5);
\draw[vS, line width=0.5pt] (6.42,4.54) .. controls (6.722,4.54) and (6.698,5.02) .. (7,5.02);
\draw[vS, line width=0.5pt] (6.42,4.54) .. controls (6.722,4.54) and (6.698,4.54) .. (7,4.54);
\draw[draw=stroke, fill=white, line width=0.5pt, rounded corners=1.2pt] (7.02,5.33) rectangle (7.52,5.67);
\node[font=\footnotesize, text=inkMain, inner sep=0] at (7.27,5.5) {$f_q$};
\draw[vF, line width=0.7pt, -{Stealth[length=2.6pt,width=2.4pt]}] (7.56,5.5) -- (7.84,5.5);
\draw[vF, fill=white, line width=0.6pt] (8.01,5.5) circle (0.14);\node[font=\scriptsize\bfseries, text=vF, inner sep=0] at (8.01,5.5) {F};
\draw[draw=stroke, fill=white, line width=0.5pt, rounded corners=1.2pt] (7.02,4.85) rectangle (7.52,5.19);
\node[font=\footnotesize, text=inkMain, inner sep=0] at (7.27,5.02) {$f_q$};
\draw[vO!82!black, line width=0.7pt, -{Stealth[length=2.6pt,width=2.4pt]}] (7.56,5.02) -- (7.84,5.02);
\draw[vO!82!black, fill=white, line width=0.6pt] (8.01,5.02) circle (0.14);\node[font=\scriptsize\bfseries, text=vO!82!black, inner sep=0] at (8.01,5.02) {O};
\draw[draw=stroke, fill=white, line width=0.5pt, rounded corners=1.2pt] (7.02,4.37) rectangle (7.52,4.71);
\node[font=\footnotesize, text=inkMain, inner sep=0] at (7.27,4.54) {$f_q$};
\draw[vS, line width=0.7pt, -{Stealth[length=2.6pt,width=2.4pt]}] (7.56,4.54) -- (7.84,4.54);
\draw[vS, fill=white, line width=0.6pt] (8.01,4.54) circle (0.14);\node[font=\scriptsize\bfseries, text=vS, inner sep=0] at (8.01,4.54) {S};
\fill[vF] (8.65,5.5) circle (0.14);\node[font=\scriptsize\bfseries, text=white, inner sep=0] at (8.65,5.5) {F};
\draw[vF, line width=0.7pt, -{Stealth[length=2.6pt,width=2.4pt]}] (8.82,5.5) -- (9.1,5.5);
\fill[vO!82!black] (8.65,5.02) circle (0.14);\node[font=\scriptsize\bfseries, text=white, inner sep=0] at (8.65,5.02) {O};
\draw[vO!82!black, line width=0.7pt, -{Stealth[length=2.6pt,width=2.4pt]}] (8.82,5.02) -- (9.1,5.02);
\fill[vS] (8.65,4.54) circle (0.14);\node[font=\scriptsize\bfseries, text=white, inner sep=0] at (8.65,4.54) {S};
\draw[vS, line width=0.7pt, -{Stealth[length=2.6pt,width=2.4pt]}] (8.82,4.54) -- (9.1,4.54);
\draw[draw=stroke, fill=white, line width=0.5pt, rounded corners=1.2pt] (9.14,4.33) rectangle (9.64,5.71);
\node[font=\footnotesize, text=inkMain, inner sep=0] at (9.39,5.02) {$f$};
\draw[vF, line width=0.7pt, -{Stealth[length=2.6pt,width=2.4pt]}] (9.68,5.5) -- (9.96,5.5);
\draw[vF, fill=white, line width=0.6pt] (10.13,5.5) circle (0.14);\node[font=\scriptsize\bfseries, text=vF, inner sep=0] at (10.13,5.5) {F};
\draw[vO!82!black, line width=0.7pt, -{Stealth[length=2.6pt,width=2.4pt]}] (9.68,5.02) -- (9.96,5.02);
\draw[vO!82!black, fill=white, line width=0.6pt] (10.13,5.02) circle (0.14);\node[font=\scriptsize\bfseries, text=vO!82!black, inner sep=0] at (10.13,5.02) {O};
\draw[vS, line width=0.7pt, -{Stealth[length=2.6pt,width=2.4pt]}] (9.68,4.54) -- (9.96,4.54);
\draw[vS, fill=white, line width=0.6pt] (10.13,4.54) circle (0.14);\node[font=\scriptsize\bfseries, text=vS, inner sep=0] at (10.13,4.54) {S};
\node[font=\footnotesize, text=inkMain, anchor=base, inner sep=0] at (4.87,3.8) {Single-variable};
\node[font=\footnotesize, text=inkMain, anchor=base, inner sep=0] at (7.13,3.8) {Auxiliary-input};
\node[font=\footnotesize, text=inkMain, anchor=base, inner sep=0] at (9.39,3.8) {Joint-output};
\node[font=\small, text=black, anchor=base west, inner sep=0pt] at (10.77,6.24) {(b) DeCo (ours)};
\fill[zone] (2.43,4.23) rectangle (3.57,5.81);
\node[font=\footnotesize, text=inkMain, anchor=base west, inner sep=0] at (0,5.405) {Flow};
\draw[vF, line width=1.0pt] (1.4,5.486) -- (1.416,5.49) -- (1.432,5.493) -- (1.448,5.495) -- (1.464,5.495) -- (1.48,5.494) -- (1.497,5.491) -- (1.513,5.488) -- (1.529,5.484) -- (1.545,5.48) -- (1.561,5.477) -- (1.577,5.475) -- (1.593,5.475) -- (1.609,5.476) -- (1.625,5.478) -- (1.641,5.481) -- (1.657,5.484) -- (1.674,5.485) -- (1.69,5.486) -- (1.706,5.486) -- (1.722,5.484) -- (1.738,5.481) -- (1.754,5.477) -- (1.77,5.474) -- (1.786,5.472) -- (1.802,5.471) -- (1.818,5.472) -- (1.835,5.475) -- (1.851,5.479) -- (1.867,5.486) -- (1.883,5.493) -- (1.899,5.5) -- (1.915,5.507) -- (1.931,5.514) -- (1.947,5.519) -- (1.963,5.524) -- (1.979,5.528) -- (1.995,5.531) -- (2.012,5.534) -- (2.028,5.538) -- (2.044,5.543) -- (2.06,5.549) -- (2.076,5.555) -- (2.092,5.561) -- (2.108,5.567) -- (2.124,5.572) -- (2.14,5.575) -- (2.156,5.576) -- (2.172,5.574) -- (2.189,5.569) -- (2.205,5.562) -- (2.221,5.552) -- (2.237,5.541) -- (2.253,5.53) -- (2.269,5.519) -- (2.285,5.51) -- (2.301,5.501) -- (2.317,5.495) -- (2.333,5.49) -- (2.35,5.487) -- (2.366,5.484) -- (2.382,5.481) -- (2.398,5.478) -- (2.414,5.475) -- (2.43,5.471);
\draw[vF!50, line width=1.0pt] (2.43,5.471) -- (2.465,5.466) -- (2.5,5.461) -- (2.535,5.456) -- (2.57,5.451) -- (2.605,5.449) -- (2.64,5.447) -- (2.675,5.448) -- (2.71,5.451) -- (2.745,5.455) -- (2.78,5.459) -- (2.815,5.464) -- (2.85,5.468) -- (2.885,5.472) -- (2.92,5.474) -- (2.955,5.474) -- (2.99,5.474) -- (3.025,5.472) -- (3.06,5.47) -- (3.095,5.469) -- (3.13,5.469) -- (3.165,5.469) -- (3.2,5.471) -- (3.235,5.475) -- (3.27,5.479) -- (3.305,5.483) -- (3.34,5.487) -- (3.375,5.489) -- (3.41,5.491) -- (3.445,5.49) -- (3.48,5.489) -- (3.515,5.485) -- (3.55,5.481);
\node[font=\footnotesize, text=inkMain, anchor=base west, inner sep=0] at (0,4.925) {Occupancy};
\draw[vO!82!black, line width=1.0pt] (1.4,4.937) -- (1.416,4.939) -- (1.432,4.941) -- (1.448,4.942) -- (1.464,4.942) -- (1.48,4.941) -- (1.497,4.94) -- (1.513,4.938) -- (1.529,4.935) -- (1.545,4.933) -- (1.561,4.931) -- (1.577,4.93) -- (1.593,4.93) -- (1.609,4.931) -- (1.625,4.932) -- (1.641,4.934) -- (1.657,4.935) -- (1.674,4.936) -- (1.69,4.937) -- (1.706,4.936) -- (1.722,4.935) -- (1.738,4.933) -- (1.754,4.931) -- (1.77,4.929) -- (1.786,4.927) -- (1.802,4.926) -- (1.818,4.925) -- (1.835,4.926) -- (1.851,4.928) -- (1.867,4.93) -- (1.883,4.933) -- (1.899,4.935) -- (1.915,4.937) -- (1.931,4.939) -- (1.947,4.939) -- (1.963,4.939) -- (1.979,4.939) -- (1.995,4.938) -- (2.012,4.937) -- (2.028,4.937) -- (2.044,4.937) -- (2.06,4.939) -- (2.076,4.942) -- (2.092,4.946) -- (2.108,4.951) -- (2.124,4.957) -- (2.14,4.963) -- (2.156,4.968) -- (2.172,4.974) -- (2.189,4.98) -- (2.205,4.985) -- (2.221,4.991) -- (2.237,4.996) -- (2.253,5.003) -- (2.269,5.009) -- (2.285,5.017) -- (2.301,5.025) -- (2.317,5.034) -- (2.333,5.042) -- (2.35,5.05) -- (2.366,5.058) -- (2.382,5.064) -- (2.398,5.069) -- (2.414,5.072) -- (2.43,5.074);
\draw[vO!82!black!50, line width=1.0pt] (2.43,5.074) -- (2.465,5.074) -- (2.5,5.073) -- (2.535,5.072) -- (2.57,5.07) -- (2.605,5.068) -- (2.64,5.066) -- (2.675,5.064) -- (2.71,5.063) -- (2.745,5.062) -- (2.78,5.061) -- (2.815,5.059) -- (2.85,5.057) -- (2.885,5.052) -- (2.92,5.047) -- (2.955,5.04) -- (2.99,5.032) -- (3.025,5.023) -- (3.06,5.014) -- (3.095,5.005) -- (3.13,4.996) -- (3.165,4.989) -- (3.2,4.983) -- (3.235,4.978) -- (3.27,4.974) -- (3.305,4.97) -- (3.34,4.968) -- (3.375,4.965) -- (3.41,4.962) -- (3.445,4.958) -- (3.48,4.954) -- (3.515,4.949) -- (3.55,4.945);
\node[font=\footnotesize, text=inkMain, anchor=base west, inner sep=0] at (0,4.445) {Speed};
\draw[vS, line width=1.0pt] (1.4,4.634) -- (1.416,4.637) -- (1.432,4.638) -- (1.448,4.64) -- (1.464,4.64) -- (1.48,4.639) -- (1.497,4.637) -- (1.513,4.635) -- (1.529,4.633) -- (1.545,4.631) -- (1.561,4.629) -- (1.577,4.628) -- (1.593,4.628) -- (1.609,4.628) -- (1.625,4.63) -- (1.641,4.631) -- (1.657,4.633) -- (1.674,4.634) -- (1.69,4.634) -- (1.706,4.634) -- (1.722,4.632) -- (1.738,4.63) -- (1.754,4.628) -- (1.77,4.626) -- (1.786,4.624) -- (1.802,4.623) -- (1.818,4.622) -- (1.835,4.623) -- (1.851,4.624) -- (1.867,4.627) -- (1.883,4.629) -- (1.899,4.631) -- (1.915,4.633) -- (1.931,4.634) -- (1.947,4.634) -- (1.963,4.633) -- (1.979,4.631) -- (1.995,4.629) -- (2.012,4.627) -- (2.028,4.624) -- (2.044,4.623) -- (2.06,4.621) -- (2.076,4.621) -- (2.092,4.62) -- (2.108,4.62) -- (2.124,4.619) -- (2.14,4.617) -- (2.156,4.613) -- (2.172,4.608) -- (2.189,4.602) -- (2.205,4.593) -- (2.221,4.584) -- (2.237,4.573) -- (2.253,4.562) -- (2.269,4.552) -- (2.285,4.542) -- (2.301,4.533) -- (2.317,4.526) -- (2.333,4.519) -- (2.35,4.515) -- (2.366,4.51) -- (2.382,4.507) -- (2.398,4.504) -- (2.414,4.501) -- (2.43,4.497);
\draw[vS!50, line width=1.0pt] (2.43,4.497) -- (2.465,4.494) -- (2.5,4.491) -- (2.535,4.488) -- (2.57,4.487) -- (2.605,4.486) -- (2.64,4.487) -- (2.675,4.489) -- (2.71,4.493) -- (2.745,4.498) -- (2.78,4.505) -- (2.815,4.511) -- (2.85,4.519) -- (2.885,4.525) -- (2.92,4.532) -- (2.955,4.538) -- (2.99,4.544) -- (3.025,4.55) -- (3.06,4.555) -- (3.095,4.561) -- (3.13,4.568) -- (3.165,4.574) -- (3.2,4.581) -- (3.235,4.589) -- (3.27,4.596) -- (3.305,4.604) -- (3.34,4.61) -- (3.375,4.615) -- (3.41,4.619) -- (3.445,4.622) -- (3.48,4.623) -- (3.515,4.623) -- (3.55,4.623);
\draw[inkMain, line width=0.45pt] (2.43,4.19) -- (2.43,5.85);
\draw[stroke, line width=0.4pt] (1.38,4.17) -- (1.38,4.12) -- (2.39,4.12) -- (2.39,4.17);
\node[font=\footnotesize, text=inkMain, anchor=base, inner sep=0] at (1.885,3.8) {history};
\draw[stroke, line width=0.4pt] (2.47,4.17) -- (2.47,4.12) -- (3.57,4.12) -- (3.57,4.17);
\node[font=\footnotesize, text=inkMain, anchor=base, inner sep=0] at (3.02,3.8) {forecast};
\node[font=\small, text=black, anchor=base west, inner sep=0pt] at (0,6.24) {(a) Problem and existing forms};
\begin{scope}
\clip[rounded corners=1.2pt] (0,0.2) rectangle (3.57,2.96);
\fill[paper] (0,0.2) rectangle (3.57,2.96);
\draw[white, line width=1.1pt] (-0.178,1.856) -- (3.748,1.47);
\draw[white, line width=1.1pt] (2.642,3.098) -- (2.999,0.062);
\draw[white, line width=1.1pt] (-0.178,2.794) -- (1.785,3.098);
\draw[white, line width=1.1pt] (0,0.062) -- (1.499,1.359);
\draw[mapCase, line width=2.1pt] (2.074,1.54) circle (0.257);
\draw[white, line width=1.3pt] (2.074,1.54) circle (0.257);
\draw[mapCase, line width=2.1pt] (1.282,1.62) circle (0.257);
\draw[white, line width=1.3pt] (1.282,1.62) circle (0.257);
\draw[mapCase, line width=4.0pt] (0.5,3.236) .. controls (1.071,1.856) and (2.213,1.194) .. (3.284,-0.076);
\draw[white, line width=2.9pt] (0.5,3.236) .. controls (1.071,1.856) and (2.213,1.194) .. (3.284,-0.076);
\draw[mapCase, line width=4.0pt] (-0.357,0.586) .. controls (1.071,0.752) and (2.213,2.408) .. (3.927,2.574);
\draw[white, line width=2.9pt] (-0.357,0.586) .. controls (1.071,0.752) and (2.213,2.408) .. (3.927,2.574);
\end{scope}
\draw[rule, line width=0.5pt, rounded corners=1.2pt] (0,0.2) rectangle (3.57,2.96);
\fill[white] (0.146,0.705) circle (0.058); \fill[gD] (0.146,0.705) circle (0.04);
\fill[white] (0.754,0.988) circle (0.058); \fill[gD] (0.754,0.988) circle (0.04);
\fill[white] (1.965,1.77) circle (0.058); \fill[gD] (1.965,1.77) circle (0.04);
\fill[white] (3.333,2.455) circle (0.058); \fill[gD] (3.333,2.455) circle (0.04);
\fill[white] (0.815,2.623) circle (0.058); \fill[gD] (0.815,2.623) circle (0.04);
\fill[white] (2.319,0.938) circle (0.058); \fill[gD] (2.319,0.938) circle (0.04);
\fill[white] (2.8,0.459) circle (0.058); \fill[gD] (2.8,0.459) circle (0.04);
\draw[inkMain, fill=white, line width=0.55pt] (2.647,2.172) circle (0.11);
\fill[inkMain] (2.647,2.172) circle (0.055);
\draw[rule, line width=0.4pt] (2.567,2.262) -- (1.4,3.63);
\draw[rule, line width=0.4pt] (2.727,2.262) -- (3.57,3.63);
\draw[black!10, line width=0.3pt] (5.81,0.389) -- (13.88,0.389);
\draw[black!10, line width=0.3pt] (5.81,1.145) -- (13.88,1.145);
\draw[black!10, line width=0.3pt] (5.81,2.658) -- (13.88,2.658);
\draw[black!45, line width=0.4pt] (5.81,0.2) -- (5.81,2.94);
\draw[black!45, line width=0.4pt] (5.81,0.389) -- (5.74,0.389);
\node[font=\footnotesize, anchor=east, inner sep=0] at (5.69,0.389) {$-8$};
\draw[black!45, line width=0.4pt] (5.81,1.145) -- (5.74,1.145);
\node[font=\footnotesize, anchor=east, inner sep=0] at (5.69,1.145) {$-4$};
\draw[black!45, line width=0.4pt] (5.81,1.901) -- (5.74,1.901);
\node[font=\footnotesize, anchor=east, inner sep=0] at (5.69,1.901) {$0$};
\draw[black!45, line width=0.4pt] (5.81,2.658) -- (5.74,2.658);
\node[font=\footnotesize, anchor=east, inner sep=0] at (5.69,2.658) {$4$};
\node[font=\footnotesize, rotate=90, anchor=base, inner sep=0] at (5.09,1.58) {MAE reduction (\%)};
\draw[black!50, line width=0.5pt] (5.81,1.901) -- (13.88,1.901);
\draw[dTO, line width=0.4pt] (6.109,1.901) -- (6.109,2.445);
\fill[dTO] (6.109,2.445) circle (0.95pt);
\draw[dTO, line width=0.4pt] (6.168,1.901) -- (6.168,2.398);
\fill[dTO] (6.168,2.398) circle (0.95pt);
\draw[dTO, line width=0.4pt] (6.226,1.901) -- (6.226,2.372);
\fill[dTO] (6.226,2.372) circle (0.95pt);
\draw[dTO, line width=0.4pt] (6.284,1.901) -- (6.284,2.292);
\fill[dTO] (6.284,2.292) circle (0.95pt);
\draw[dTO, line width=0.4pt] (6.343,1.901) -- (6.343,2.236);
\fill[dTO] (6.343,2.236) circle (0.95pt);
\draw[dTO, line width=0.4pt] (6.401,1.901) -- (6.401,2.184);
\fill[dTO] (6.401,2.184) circle (0.95pt);
\draw[dTO, line width=0.4pt] (6.459,1.901) -- (6.459,2.147);
\fill[dTO] (6.459,2.147) circle (0.95pt);
\draw[dTO, line width=0.4pt] (6.518,1.901) -- (6.518,2.145);
\fill[dTO] (6.518,2.145) circle (0.95pt);
\draw[dTO, line width=0.4pt] (6.576,1.901) -- (6.576,2.134);
\fill[dTO] (6.576,2.134) circle (0.95pt);
\draw[dTO, line width=0.4pt] (6.634,1.901) -- (6.634,2.118);
\fill[dTO] (6.634,2.118) circle (0.95pt);
\draw[dTO, line width=0.4pt] (6.692,1.901) -- (6.692,2.109);
\fill[dTO] (6.692,2.109) circle (0.95pt);
\draw[dTO, line width=0.4pt] (6.751,1.901) -- (6.751,2.104);
\fill[dTO] (6.751,2.104) circle (0.95pt);
\draw[dTO, line width=0.4pt] (6.809,1.901) -- (6.809,2.101);
\fill[dTO] (6.809,2.101) circle (0.95pt);
\draw[dTO, line width=0.4pt] (6.867,1.901) -- (6.867,2.073);
\fill[dTO] (6.867,2.073) circle (0.95pt);
\draw[dTO, line width=0.4pt] (6.926,1.901) -- (6.926,2.068);
\fill[dTO] (6.926,2.068) circle (0.95pt);
\draw[dTO, line width=0.4pt] (6.984,1.901) -- (6.984,2.059);
\fill[dTO] (6.984,2.059) circle (0.95pt);
\draw[dTO, line width=0.4pt] (7.042,1.901) -- (7.042,2.056);
\fill[dTO] (7.042,2.056) circle (0.95pt);
\draw[dTO, line width=0.4pt] (7.101,1.901) -- (7.101,2.044);
\fill[dTO] (7.101,2.044) circle (0.95pt);
\draw[dTO, line width=0.4pt] (7.159,1.901) -- (7.159,2.037);
\fill[dTO] (7.159,2.037) circle (0.95pt);
\draw[dTO, line width=0.4pt] (7.218,1.901) -- (7.218,2.032);
\fill[dTO] (7.218,2.032) circle (0.95pt);
\draw[dTO, line width=0.4pt] (7.276,1.901) -- (7.276,2.031);
\fill[dTO] (7.276,2.031) circle (0.95pt);
\draw[dTO, line width=0.4pt] (7.334,1.901) -- (7.334,2.028);
\fill[dTO] (7.334,2.028) circle (0.95pt);
\draw[dTO, line width=0.4pt] (7.393,1.901) -- (7.393,1.998);
\fill[dTO] (7.393,1.998) circle (0.95pt);
\draw[dTO, line width=0.4pt] (7.451,1.901) -- (7.451,1.996);
\fill[dTO] (7.451,1.996) circle (0.95pt);
\draw[dTO, line width=0.4pt] (7.509,1.901) -- (7.509,1.993);
\fill[dTO] (7.509,1.993) circle (0.95pt);
\draw[dTO, line width=0.4pt] (7.567,1.901) -- (7.567,1.992);
\fill[dTO] (7.567,1.992) circle (0.95pt);
\draw[dTO, line width=0.4pt] (7.626,1.901) -- (7.626,1.991);
\fill[dTO] (7.626,1.991) circle (0.95pt);
\draw[dTO, line width=0.4pt] (7.684,1.901) -- (7.684,1.979);
\fill[dTO] (7.684,1.979) circle (0.95pt);
\draw[dTO, line width=0.4pt] (7.742,1.901) -- (7.742,1.978);
\fill[dTO] (7.742,1.978) circle (0.95pt);
\draw[dTO, line width=0.4pt] (7.801,1.901) -- (7.801,1.923);
\fill[dTO] (7.801,1.923) circle (0.95pt);
\draw[dTO, line width=0.4pt] (7.859,1.901) -- (7.859,1.919);
\fill[dTO] (7.859,1.919) circle (0.95pt);
\draw[dTO, line width=0.4pt] (7.917,1.901) -- (7.917,1.899);
\fill[dTO] (7.917,1.899) circle (0.95pt);
\draw[dTO, line width=0.4pt] (7.976,1.901) -- (7.976,1.888);
\fill[dTO] (7.976,1.888) circle (0.95pt);
\draw[dTO, line width=0.4pt] (8.034,1.901) -- (8.034,1.886);
\fill[dTO] (8.034,1.886) circle (0.95pt);
\draw[dTO, line width=0.4pt] (8.092,1.901) -- (8.092,1.822);
\fill[dTO] (8.092,1.822) circle (0.95pt);
\draw[dTO, line width=0.4pt] (8.151,1.901) -- (8.151,1.816);
\fill[dTO] (8.151,1.816) circle (0.95pt);
\draw[dSV, line width=0.4pt] (8.369,1.901) -- (8.369,2.395);
\filldraw[fill=white, draw=dSV, line width=0.5pt] (8.369,2.395) circle (0.95pt);
\draw[dSV, line width=0.4pt] (8.427,1.901) -- (8.427,2.245);
\filldraw[fill=white, draw=dSV, line width=0.5pt] (8.427,2.245) circle (0.95pt);
\draw[dSV, line width=0.4pt] (8.486,1.901) -- (8.486,2.206);
\filldraw[fill=white, draw=dSV, line width=0.5pt] (8.486,2.206) circle (0.95pt);
\draw[dSV, line width=0.4pt] (8.544,1.901) -- (8.544,2.009);
\filldraw[fill=white, draw=dSV, line width=0.5pt] (8.544,2.009) circle (0.95pt);
\draw[dSV, line width=0.4pt] (8.602,1.901) -- (8.602,2);
\filldraw[fill=white, draw=dSV, line width=0.5pt] (8.602,2) circle (0.95pt);
\draw[dSV, line width=0.4pt] (8.661,1.901) -- (8.661,1.995);
\filldraw[fill=white, draw=dSV, line width=0.5pt] (8.661,1.995) circle (0.95pt);
\draw[dSV, line width=0.4pt] (8.719,1.901) -- (8.719,1.942);
\filldraw[fill=white, draw=dSV, line width=0.5pt] (8.719,1.942) circle (0.95pt);
\draw[dSV, line width=0.4pt] (8.777,1.901) -- (8.777,1.938);
\filldraw[fill=white, draw=dSV, line width=0.5pt] (8.777,1.938) circle (0.95pt);
\draw[dSV, line width=0.4pt] (8.836,1.901) -- (8.836,1.929);
\filldraw[fill=white, draw=dSV, line width=0.5pt] (8.836,1.929) circle (0.95pt);
\draw[dSV, line width=0.4pt] (8.894,1.901) -- (8.894,1.872);
\filldraw[fill=white, draw=dSV, line width=0.5pt] (8.894,1.872) circle (0.95pt);
\draw[dSV, line width=0.4pt] (8.952,1.901) -- (8.952,1.847);
\filldraw[fill=white, draw=dSV, line width=0.5pt] (8.952,1.847) circle (0.95pt);
\draw[dSV, line width=0.4pt] (9.011,1.901) -- (9.011,1.805);
\filldraw[fill=white, draw=dSV, line width=0.5pt] (9.011,1.805) circle (0.95pt);
\draw[dSV, line width=0.4pt] (9.069,1.901) -- (9.069,1.793);
\filldraw[fill=white, draw=dSV, line width=0.5pt] (9.069,1.793) circle (0.95pt);
\draw[dSV, line width=0.4pt] (9.127,1.901) -- (9.127,1.752);
\filldraw[fill=white, draw=dSV, line width=0.5pt] (9.127,1.752) circle (0.95pt);
\draw[dSV, line width=0.4pt] (9.186,1.901) -- (9.186,1.749);
\filldraw[fill=white, draw=dSV, line width=0.5pt] (9.186,1.749) circle (0.95pt);
\draw[dSV, line width=0.4pt] (9.244,1.901) -- (9.244,1.721);
\filldraw[fill=white, draw=dSV, line width=0.5pt] (9.244,1.721) circle (0.95pt);
\draw[dSV, line width=0.4pt] (9.302,1.901) -- (9.302,1.683);
\filldraw[fill=white, draw=dSV, line width=0.5pt] (9.302,1.683) circle (0.95pt);
\draw[dSV, line width=0.4pt] (9.361,1.901) -- (9.361,1.683);
\filldraw[fill=white, draw=dSV, line width=0.5pt] (9.361,1.683) circle (0.95pt);
\draw[dSV, line width=0.4pt] (9.419,1.901) -- (9.419,1.675);
\filldraw[fill=white, draw=dSV, line width=0.5pt] (9.419,1.675) circle (0.95pt);
\draw[dSV, line width=0.4pt] (9.477,1.901) -- (9.477,1.667);
\filldraw[fill=white, draw=dSV, line width=0.5pt] (9.477,1.667) circle (0.95pt);
\draw[dSV, line width=0.4pt] (9.536,1.901) -- (9.536,1.585);
\filldraw[fill=white, draw=dSV, line width=0.5pt] (9.536,1.585) circle (0.95pt);
\draw[dSV, line width=0.4pt] (9.594,1.901) -- (9.594,1.568);
\filldraw[fill=white, draw=dSV, line width=0.5pt] (9.594,1.568) circle (0.95pt);
\draw[dSV, line width=0.4pt] (9.652,1.901) -- (9.652,1.564);
\filldraw[fill=white, draw=dSV, line width=0.5pt] (9.652,1.564) circle (0.95pt);
\draw[dSV, line width=0.4pt] (9.711,1.901) -- (9.711,1.478);
\filldraw[fill=white, draw=dSV, line width=0.5pt] (9.711,1.478) circle (0.95pt);
\draw[dSV, line width=0.4pt] (9.769,1.901) -- (9.769,1.47);
\filldraw[fill=white, draw=dSV, line width=0.5pt] (9.769,1.47) circle (0.95pt);
\draw[dSV, line width=0.4pt] (9.827,1.901) -- (9.827,1.447);
\filldraw[fill=white, draw=dSV, line width=0.5pt] (9.827,1.447) circle (0.95pt);
\draw[dSV, line width=0.4pt] (9.886,1.901) -- (9.886,1.415);
\filldraw[fill=white, draw=dSV, line width=0.5pt] (9.886,1.415) circle (0.95pt);
\draw[dSV, line width=0.4pt] (9.944,1.901) -- (9.944,1.357);
\filldraw[fill=white, draw=dSV, line width=0.5pt] (9.944,1.357) circle (0.95pt);
\draw[dSV, line width=0.4pt] (10.002,1.901) -- (10.002,1.352);
\filldraw[fill=white, draw=dSV, line width=0.5pt] (10.002,1.352) circle (0.95pt);
\draw[dSV, line width=0.4pt] (10.061,1.901) -- (10.061,1.242);
\filldraw[fill=white, draw=dSV, line width=0.5pt] (10.061,1.242) circle (0.95pt);
\draw[dSV, line width=0.4pt] (10.119,1.901) -- (10.119,1.206);
\filldraw[fill=white, draw=dSV, line width=0.5pt] (10.119,1.206) circle (0.95pt);
\draw[dSV, line width=0.4pt] (10.177,1.901) -- (10.177,1.043);
\filldraw[fill=white, draw=dSV, line width=0.5pt] (10.177,1.043) circle (0.95pt);
\draw[dSV, line width=0.4pt] (10.236,1.901) -- (10.236,0.843);
\filldraw[fill=white, draw=dSV, line width=0.5pt] (10.236,0.843) circle (0.95pt);
\draw[dSV, line width=0.4pt] (10.294,1.901) -- (10.294,0.688);
\filldraw[fill=white, draw=dSV, line width=0.5pt] (10.294,0.688) circle (0.95pt);
\draw[dSV, line width=0.4pt] (10.352,1.901) -- (10.352,0.574);
\filldraw[fill=white, draw=dSV, line width=0.5pt] (10.352,0.574) circle (0.95pt);
\draw[dSV, line width=0.4pt] (10.411,1.901) -- (10.411,0.354);
\filldraw[fill=white, draw=dSV, line width=0.5pt] (10.411,0.354) circle (0.95pt);
\draw[decoAccent, line width=0.6pt] (11.328,1.901) -- (11.328,2.51);
\fill[decoAccent] (11.328,2.51) circle (1.15pt);
\draw[decoAccent, line width=0.6pt] (11.504,1.901) -- (11.504,2.396);
\fill[decoAccent] (11.504,2.396) circle (1.15pt);
\draw[decoAccent, line width=0.6pt] (11.68,1.901) -- (11.68,2.383);
\fill[decoAccent] (11.68,2.383) circle (1.15pt);
\draw[decoAccent, line width=0.6pt] (11.855,1.901) -- (11.855,2.382);
\fill[decoAccent] (11.855,2.382) circle (1.15pt);
\draw[decoAccent, line width=0.6pt] (12.031,1.901) -- (12.031,2.365);
\fill[decoAccent] (12.031,2.365) circle (1.15pt);
\draw[decoAccent, line width=0.6pt] (12.207,1.901) -- (12.207,2.302);
\fill[decoAccent] (12.207,2.302) circle (1.15pt);
\draw[decoAccent, line width=0.6pt] (12.383,1.901) -- (12.383,2.286);
\fill[decoAccent] (12.383,2.286) circle (1.15pt);
\draw[decoAccent, line width=0.6pt] (12.559,1.901) -- (12.559,2.271);
\fill[decoAccent] (12.559,2.271) circle (1.15pt);
\draw[decoAccent, line width=0.6pt] (12.735,1.901) -- (12.735,2.244);
\fill[decoAccent] (12.735,2.244) circle (1.15pt);
\draw[decoAccent, line width=0.6pt] (12.91,1.901) -- (12.91,2.171);
\fill[decoAccent] (12.91,2.171) circle (1.15pt);
\draw[decoAccent, line width=0.6pt] (13.086,1.901) -- (13.086,2.101);
\fill[decoAccent] (13.086,2.101) circle (1.15pt);
\draw[decoAccent, line width=0.6pt] (13.262,1.901) -- (13.262,2.031);
\fill[decoAccent] (13.262,2.031) circle (1.15pt);
\draw[black!75, line width=0.5pt, dash pattern=on 1.4pt off 0.9pt] (6.05,2.04) -- (8.21,2.04);
\node[font=\footnotesize, anchor=south, inner sep=1pt] at (7.13,2.554) {$+0.7$\%};
\draw[black!75, line width=0.5pt, dash pattern=on 1.4pt off 0.9pt] (8.31,1.679) -- (10.47,1.679);
\node[font=\footnotesize, anchor=south, inner sep=1pt] at (9.39,2.554) {${-}1.2$\%};
\draw[black!75, line width=0.5pt, dash pattern=on 1.4pt off 0.9pt] (11.21,2.294) -- (13.38,2.294);
\node[font=\footnotesize, anchor=south, inner sep=1pt] at (12.295,2.554) {$+2.1$\%};
\node[font=\small, text=black, anchor=base west, inner sep=0pt] at (3.99,3.24) {(c) Change over single-variable forecasting};
\begin{scope}[shift={(16.18,3.468)}]
\draw[black!14, line width=0.25pt] (0,0) circle (0.232);
\draw[black!14, line width=0.25pt] (0,0) circle (0.464);
\draw[black!14, line width=0.25pt] (0,0) circle (0.928);
\draw[black!14, line width=0.25pt] (0,0) circle (1.160);
\draw[black!14, line width=0.25pt] (0,0) circle (1.392);
\draw[black!10, line width=0.25pt] (0,0) -- (72:1.450);
\draw[black!10, line width=0.25pt] (0,0) -- (54:1.450);
\draw[black!10, line width=0.25pt] (0,0) -- (36:1.450);
\draw[black!10, line width=0.25pt] (0,0) -- (18:1.450);
\draw[black!10, line width=0.25pt] (0,0) -- (-18:1.450);
\draw[black!10, line width=0.25pt] (0,0) -- (-36:1.450);
\draw[black!10, line width=0.25pt] (0,0) -- (-54:1.450);
\draw[black!10, line width=0.25pt] (0,0) -- (-72:1.450);
\draw[black!10, line width=0.25pt] (0,0) -- (-108:1.450);
\draw[black!10, line width=0.25pt] (0,0) -- (-126:1.450);
\draw[black!10, line width=0.25pt] (0,0) -- (-144:1.450);
\draw[black!10, line width=0.25pt] (0,0) -- (-162:1.450);
\draw[black!10, line width=0.25pt] (0,0) -- (-198:1.450);
\draw[black!10, line width=0.25pt] (0,0) -- (-216:1.450);
\draw[black!10, line width=0.25pt] (0,0) -- (-234:1.450);
\draw[black!10, line width=0.25pt] (0,0) -- (-252:1.450);
\draw[black!50, line width=0.5pt] (0,0) circle (0.696);
\draw[dTO, line width=0.8pt, line join=round, dash pattern=on 2.2pt off 1.4pt] (72:0.743) -- (54:0.737) -- (36:0.715) -- (18:0.784) -- (-18:0.733) -- (-36:0.510) -- (-54:0.766) -- (-72:0.657) -- (-108:1.123) -- (-126:1.229) -- (-144:1.015) -- (-162:1.328) -- (-198:0.486) -- (-216:0.331) -- (-234:0.526) -- (-252:0.164) -- cycle;
\draw[decoAccent, line width=1.0pt, line join=round] (72:0.883) -- (54:0.838) -- (36:0.843) -- (18:0.848) -- (-18:0.833) -- (-36:0.780) -- (-54:0.957) -- (-72:0.807) -- (-108:1.346) -- (-126:1.393) -- (-144:1.174) -- (-162:1.406) -- (-198:0.547) -- (-216:0.566) -- (-234:0.639) -- (-252:0.376) -- cycle;
\filldraw[fill=white, draw=dTO, line width=0.45pt] (72:0.743) circle (1.0pt);
\fill[decoAccent] (72:0.883) circle (1.1pt);
\filldraw[fill=white, draw=dTO, line width=0.45pt] (54:0.737) circle (1.0pt);
\fill[decoAccent] (54:0.838) circle (1.1pt);
\filldraw[fill=white, draw=dTO, line width=0.45pt] (36:0.715) circle (1.0pt);
\fill[decoAccent] (36:0.843) circle (1.1pt);
\filldraw[fill=white, draw=dTO, line width=0.45pt] (18:0.784) circle (1.0pt);
\fill[decoAccent] (18:0.848) circle (1.1pt);
\filldraw[fill=white, draw=dTO, line width=0.45pt] (-18:0.733) circle (1.0pt);
\fill[decoAccent] (-18:0.833) circle (1.1pt);
\filldraw[fill=white, draw=dTO, line width=0.45pt] (-36:0.510) circle (1.0pt);
\fill[decoAccent] (-36:0.780) circle (1.1pt);
\filldraw[fill=white, draw=dTO, line width=0.45pt] (-54:0.766) circle (1.0pt);
\fill[decoAccent] (-54:0.957) circle (1.1pt);
\filldraw[fill=white, draw=dTO, line width=0.45pt] (-72:0.657) circle (1.0pt);
\fill[decoAccent] (-72:0.807) circle (1.1pt);
\filldraw[fill=white, draw=dTO, line width=0.45pt] (-108:1.123) circle (1.0pt);
\fill[decoAccent] (-108:1.346) circle (1.1pt);
\filldraw[fill=white, draw=dTO, line width=0.45pt] (-126:1.229) circle (1.0pt);
\fill[decoAccent] (-126:1.393) circle (1.1pt);
\filldraw[fill=white, draw=dTO, line width=0.45pt] (-144:1.015) circle (1.0pt);
\fill[decoAccent] (-144:1.174) circle (1.1pt);
\filldraw[fill=white, draw=dTO, line width=0.45pt] (-162:1.328) circle (1.0pt);
\fill[decoAccent] (-162:1.406) circle (1.1pt);
\filldraw[fill=white, draw=dTO, line width=0.45pt] (-198:0.486) circle (1.0pt);
\fill[decoAccent] (-198:0.547) circle (1.1pt);
\filldraw[fill=white, draw=dTO, line width=0.45pt] (-216:0.331) circle (1.0pt);
\fill[decoAccent] (-216:0.566) circle (1.1pt);
\filldraw[fill=white, draw=dTO, line width=0.45pt] (-234:0.526) circle (1.0pt);
\fill[decoAccent] (-234:0.639) circle (1.1pt);
\filldraw[fill=white, draw=dTO, line width=0.45pt] (-252:0.164) circle (1.0pt);
\fill[decoAccent] (-252:0.376) circle (1.1pt);
\node[font=\scriptsize, text=inkMuted, fill=white, inner sep=0.8pt] at (96:0.696) {$0$};
\node[font=\scriptsize, text=inkMuted, fill=white, inner sep=0.8pt] at (96:1.160) {$+8\%$};
\node[font=\scriptsize, text=inkMuted, inner sep=0.5pt] at (72:1.580) {03};
\node[font=\scriptsize, text=inkMuted, inner sep=0.5pt] at (54:1.580) {04};
\node[font=\scriptsize, text=inkMuted, inner sep=0.5pt] at (36:1.580) {07};
\node[font=\scriptsize, text=inkMuted, inner sep=0.5pt] at (18:1.580) {08};
\draw[black!45, line width=0.4pt] (78:1.720) arc[start angle=78, end angle=12, radius=1.720cm];
\draw[black!45, line width=0.4pt] (-12:1.720) arc[start angle=-12, end angle=-78, radius=1.720cm];
\draw[black!45, line width=0.4pt] (-102:1.720) arc[start angle=-102, end angle=-168, radius=1.720cm];
\draw[black!45, line width=0.4pt] (-192:1.720) arc[start angle=-192, end angle=-258, radius=1.720cm];
\end{scope}
\node[font=\footnotesize, text=inkMain, anchor=base west, inner sep=0] at (14.36,5.3) {Recovery};
\node[font=\footnotesize, text=inkMain, anchor=base east, inner sep=0] at (18,5.3) {All targets};
\node[font=\footnotesize, text=inkMain, anchor=base west, inner sep=0] at (14.36,1.445) {In congestion};
\node[font=\footnotesize, text=inkMain, anchor=base east, inner sep=0] at (18,1.445) {Onset};
\node[font=\small, text=black, anchor=base west, inner sep=0pt] at (14.36,6.24) {(d) Speed MAE reduction};
\node[font=\small, text=black, anchor=base west, inner sep=0pt] at (14.84,5.84) {by traffic state};
\draw[dTO, line width=0.8pt, dash pattern=on 2.2pt off 1.4pt] (14.312,0.925) -- (14.752,0.925);\filldraw[fill=white, draw=dTO, line width=0.45pt] (14.532,0.925) circle (1.05pt);
\node[font=\footnotesize, text=inkMain, anchor=base west, inner sep=0] at (14.852,0.83) {Target-only};
\draw[decoAccent, line width=1.0pt] (16.602,0.925) -- (17.042,0.925);\fill[decoAccent] (16.822,0.925) circle (1.15pt);
\node[font=\footnotesize, text=inkMain, anchor=base west, inner sep=0] at (17.142,0.83) {DeCo-S};
\draw[black!50, line width=0.5pt] (15.055,0.525) -- (15.495,0.525);
\node[font=\footnotesize, text=inkMain, anchor=base west, inner sep=0] at (15.595,0.43) {Single-variable};
\end{tikzpicture}

%% file: sections/related_data.tex
\section{Related Work}
\label{sec:related-work}

\subsection{Spatiotemporal Traffic Forecasting}
Deep spatiotemporal forecasters combine temporal modeling with message passing over road graphs that are either given or learned~\cite{li2018dcrnn,yu2018stgcn,wu2019graphwavenet,guo2019astgcn,song2020stsgcn}. Subsequent work has refined this design along several directions: separating diffused from inherent signals (D$^2$STGNN~\cite{shao2022d2stgnn}), adapting parameters to spatial and temporal heterogeneity (ST-WA~\cite{cirstea2022stwa}, HimNet~\cite{dong2024himnet}), enriching Transformers with spatiotemporal embeddings (STAEformer~\cite{liu2023staeformer}), modeling traffic in the spectral domain (STWave~\cite{fang2023stwave}), and scaling to large road networks (BigST~\cite{han2024bigst}, RAGC~\cite{ragc}). Pre-trained and universal models further learn spatiotemporal representations that transfer across predictors and datasets~\cite{gao2024stdmae,yuan2024unist}. Benchmarks such as DL-Traff~\cite{jiang2021dltraff} and TFB~\cite{qiu2024tfb} support fair comparison across these models. All of them improve the forecaster itself and are complementary to DeCo, which can take such models as its variable-specific forecasters (Section~\ref{sec:ragc}). Its default forecaster, DecoupleST, follows D$^2$STGNN in separating a network component from a local one (Section~\ref{sec:decouplest}).

\begin{table}[t]
\centering
\caption{Representative traffic forecasting approaches.}
\label{tab:related}
\scriptsize
\setlength{\tabcolsep}{2.2pt}
\begin{tabular}{llcccccc}
\toprule
\multirow{2}{*}{Type} & \multirow{2}{*}{Approach} & \multicolumn{3}{c}{Variables} & Decoupled & Residual & Cross-variable\\
\cmidrule(lr){3-5}
 & & F & O & S & forecasting & correction & coordination\\
\midrule
\multirow{4}{*}{Single} & DCRNN~\cite{li2018dcrnn} & -- & -- & $\bullet$ & \checkmark & -- & --\\
 & STGCN~\cite{yu2018stgcn} & -- & -- & $\bullet$ & \checkmark & -- & --\\
 & STSGCN~\cite{song2020stsgcn} & $\bullet$ & -- & -- & \checkmark & -- & --\\
 & RAGC~\cite{ragc} & $\bullet$ & -- & -- & \checkmark & -- & --\\
\midrule
\multirow{4}{*}{Auxiliary} & ASTGCN~\cite{guo2019astgcn} & $\bullet$ & $\circ$ & $\circ$ & \checkmark & -- & --\\
 & DMSTGCN~\cite{han2021dynamic} & $\circ$ & -- & $\bullet$ & \checkmark & -- & --\\
 & MS-GAT~\cite{huang2022multiaspect} & $\bullet$ & $\circ$ & $\circ$ & \checkmark & -- & --\\
 & MVC-STNet~\cite{miao2024mvcstnet} & $\bullet$ & $\circ$ & $\circ$ & \checkmark & -- & --\\
\midrule
\multirow{4}{*}{Joint} & AutoSTL~\cite{zhang2023autostl} & $\bullet$ & -- & $\bullet$ & -- & -- & --\\
 & DG-STMTL~\cite{dgstmtl2025} & $\bullet$ & -- & $\bullet$ & -- & -- & --\\
 & MTNet~\cite{zhang2026mtnet} & $\bullet$ & -- & $\bullet$ & -- & -- & --\\
 & MCST-Mamba~\cite{hamad2025mcstmamba} & $\bullet$ & $\bullet$ & $\bullet$ & -- & -- & --\\
\midrule
\multirow{3}{*}{Correction} & ResCAL~\cite{kim2022rescal} & -- & -- & $\bullet$ & \checkmark & \checkmark & --\\
 & ADCSD~\cite{guo2025adcsd} & $\bullet$ & -- & -- & \checkmark & \checkmark & --\\
 & RiskTraf~\cite{wang2026risktraf} & $\bullet$ & $\circ$ & $\circ$ & \checkmark & \checkmark & --\\
\midrule
 & \textbf{DeCo (ours)} & $\bullet$ & $\bullet$ & $\bullet$ & \checkmark & \checkmark & \checkmark\\
\bottomrule
\multicolumn{8}{l}{F/O/S: flow, occupancy, speed. $\bullet$ forecast, $\circ$ auxiliary input only, -- not used.}\\
\multicolumn{8}{l}{Decoupled forecasting: one forecaster per variable.}\\
\multicolumn{8}{l}{Residual correction: a learned correction is added to fixed forecasts.}\\
\multicolumn{8}{l}{Cross-variable coordination: each variable is corrected with the others' forecasts.}\\
\end{tabular}
\end{table}

\subsection{Forecasting with Multiple Traffic Variables}
Traffic forecasters differ in the variables they read and predict (Fig.~\ref{fig:intro}(a)), and Table~\ref{tab:related} lists the settings reported in the original studies. Most spatiotemporal forecasters predict one variable, flow or speed, from its own history~\cite{li2018dcrnn,yu2018stgcn,song2020stsgcn,ragc}. Auxiliary-input models read several variables to predict one: ASTGCN, MS-GAT, and MVC-STNet predict flow from flow, occupancy, and speed~\cite{guo2019astgcn,huang2022multiaspect,miao2024mvcstnet}, and DMSTGCN fuses hidden states learned from traffic volume into those learned from speed~\cite{han2021dynamic}. Beyond traffic, TimeXer forecasts an endogenous series with the help of exogenous ones~\cite{wang2024timexer}. Joint-output models predict several variables with one network: AutoSTL, DG-STMTL, and MTNet predict flow and speed with shared representations and cross-task interactions~\cite{zhang2023autostl,dgstmtl2025,zhang2026mtnet}, and MCST-Mamba models all three variables at once~\cite{hamad2025mcstmamba}. Sharing can exploit the relations between tasks~\cite{caruana1997multitask} but can also cause interference between them~\cite{dgstmtl2025}, which multi-task learning mitigates through loss weighting~\cite{kendall2018uncertainty,chen2018gradnorm} or gradient surgery~\cite{yu2020pcgrad}. In general multivariate forecasting, the same question appears as channel dependence: PatchTST processes each channel independently with shared weights~\cite{nie2023patchtst}, iTransformer applies attention across variate tokens~\cite{liu2024itransformer}, and Han et al.\ analyze the trade-off between the capacity and robustness of the two strategies~\cite{han2024capacity}. DeCo combines the two strategies in separate stages. Its fold forecasters treat each variable independently, and a residual coordination stage introduces cross-variable dependence through their forecasts rather than through shared histories or hidden representations.

\subsection{Forecast Combination and Correction}
Forecast combination pools predictions of the same quantity from multiple models~\cite{bates1969combination}, and stacked generalization learns a second-stage predictor from held-out first-stage outputs~\cite{wolpert1992stacked}. In multi-target regression, stacked single-target regression extends this idea by feeding the first-stage predictions of all targets into the second stage~\cite{spyromitros2016mtr}. In traffic forecasting, residual corrections have been attached to trained forecasters: ResCAL estimates a forecaster's future errors from its past errors and graph signals~\cite{kim2022rescal}, ADCSD adapts a light network online to correct a fixed forecaster's output~\cite{guo2025adcsd}, EvoST calibrates a forecaster per sensor from delayed labels~\cite{yin2026evost}, and RiskTraf learns a flow correction from historical speed and occupancy~\cite{wang2026risktraf}. Each of these methods corrects a single variable from its own forecast, its past errors, or the histories of the other variables. DeCo instead corrects all three variables from the forecasts of all of them. Its module corrects whole horizons with parameters shared across sensors, and its cross-fitting uses contiguous time blocks because training windows overlap. As Table~\ref{tab:related} summarizes, DeCo alone combines decoupled forecasting, residual correction, and cross-variable coordination.

\section{Data and Problem}
\label{sec:problem}

\subsection{Three-Variable PeMS Benchmarks}
\label{sec:data}

\begin{table}[t]
\centering
\caption{Datasets at five-minute resolution.}
\label{tab:data}
\footnotesize
\setlength{\tabcolsep}{3.5pt}
\begin{tabular}{lrrlrrr}
\toprule
 & & & & \multicolumn{2}{c}{Invalid (\%)} & No-speed\\
\cmidrule(lr){5-6}
Dataset & Sensors & Days & Start & Occ. & Speed & sensors\\
\midrule
PeMS03 & 358 & 91 & 2018-09-01 & 0.48 & 2.62 & 8\\
PeMS04 & 307 & 59 & 2018-01-01 & $0^\dagger$ & $0^\dagger$ & $0^\dagger$\\
PeMS07 & 883 & 98 & 2017-05-01 & 0.20 & 1.25 & 11\\
PeMS08 & 170 & 62 & 2016-07-01 & $0^\dagger$ & $0^\dagger$ & $0^\dagger$\\
\bottomrule
\multicolumn{7}{l}{\scriptsize $\dagger$ Every finite entry of the PeMS04/08 releases is treated as valid.}\\
\multicolumn{7}{l}{\scriptsize No-speed sensors: sensors without any valid speed record.}\\
\end{tabular}
\end{table}

We use five-minute station-level records of flow, occupancy, and speed from the Caltrans Performance Measurement System (PeMS)~\cite{chen2001pems}. Of the four benchmarks most used for spatiotemporal forecasting~\cite{guo2019astgcn,song2020stsgcn}, PeMS04 and PeMS08 are distributed with all three variables, whereas the PeMS03 and PeMS07 files distributed with STSGCN~\cite{song2020stsgcn} contain flow only, with $26{,}208\times358$ and $28{,}224\times883$ time--sensor entries. We extend PeMS03 and PeMS07 to three variables and keep their sensors, periods, and flow values, so that all four benchmarks support three-variable forecasting on their original sensors and splits (Table~\ref{tab:data}). Other three-variable PeMS releases re-select sensors and therefore do not extend the standard benchmarks. RiskTraf rebuilds Caltrans districts 3, 4, 7, and 8 from mainline and HOV detectors that record all three variables~\cite{wang2026risktraf}, and TraffiDent covers 16,972 PeMS traffic nodes~\cite{gou2025traffident}.

\textbf{Alignment.} Occupancy and speed come from the five-minute PeMS station records of the same periods. PeMS03 sensors follow the benchmark's published station list. The nodes of PeMS07 are anonymous, so each node is matched to a station by comparing its flow with the PeMS records over all 98 days. Where PeMS reports flow, it agrees with the benchmark flow to within $10^{-6}$ in 99.956\% of the PeMS03 entries and in every PeMS07 entry. Where they differ, we keep the benchmark value. Both datasets keep the benchmark road graphs and node order.

\textbf{Validity and release.} Flow is valid everywhere. An occupancy or speed entry is valid if the station record has a value. Otherwise it holds zero and is masked. We release the aligned arrays with the validity masks, the station identifiers, and the checksums of every source file. Their use remains subject to the terms of the upstream PeMS data and benchmark files. Flow counts vehicles per five-minute interval, speed is in miles per hour, and occupancy is stored as a fraction, with errors reported in percentage points.

\subsection{Problem Definition}

\textbf{Multi-variable sensor data.} A road sensor network has $N$ sensors. For every sensor and five-minute step $t$, we represent $Q=3$ variables, flow, occupancy, and speed, collected in $x_t\in\mathbb{R}^{N\times Q}$, with a validity mask $v_t\in\{0,1\}^{N\times Q}$. An invalid input is replaced by the last valid value of the same sensor and variable, or by zero before the first valid value. We write $\tilde{x}_t$ for these causally filled inputs, which use only values up to step $t$.

\textbf{Forecasting task.} At origin $t$, a forecaster receives the last $P=12$ steps $\tilde{X}_t=(\tilde{x}_{t-P+1},\dots,\tilde{x}_t)$ and forecasts the targets $y_{t+h}=x_{t+h}$ for leads $h=1,\dots,H$, with $H=12$. We write $\hat{y}^{(q)}_{t,h,i}$ for the forecast of variable $q$ at sensor $i$, and $X^{(q)}$, $Y^{(q)}$, $X_i$, and $Y_i$ for the slices of a variable or a sensor.

\textbf{Input and output forms.} A forecaster can use the variables in three forms, named by the variables it reads and predicts. In the single-input single-output form (SISO, 1$\to$1), $\hat{Y}^{(q)}=f_q(\tilde{X}^{(q)})$. In the multi-input single-output form (MISO, 3$\to$1), $\hat{Y}^{(q)}=f_q(\tilde{X})$. In the multi-input multi-output form (MIMO, 3$\to$3), $\hat{Y}=f(\tilde{X})$. The forms differ only in the traffic variables they read and predict, and every form returns a twelve-step forecast. The first two forms train one model per variable, and the third trains one model for all variables. SISO, MISO, and MIMO correspond to the single-variable, auxiliary-input, and joint-output paradigms of Section~\ref{sec:intro}.

\textbf{Evaluation.} We evaluate each variable separately by the MAE over valid targets,
\begin{equation}
\mathrm{MAE}^{(q)}=\frac{\sum_{t,h,i} v^{(q)}_{t+h,i}\,\bigl|\hat{y}^{(q)}_{t,h,i}-y^{(q)}_{t+h,i}\bigr|}{\sum_{t,h,i} v^{(q)}_{t+h,i}},
\end{equation}
over all test origins, leads, and sensors. Valid zero targets are included. The masked MAE commonly used on these benchmarks excludes all zero targets, so our errors are not directly comparable with published values. RMSE is defined analogously.

%% file: sections/method.tex
\section{DeCo}
\label{sec:method}

\subsection{Overview}
\label{sec:overview}

\input{figures/overview}

DeCo separates the forecasting of each traffic variable from the modeling of cross-variable dependence (Fig.~\ref{fig:overview}). In Stage~1, a forecaster $f_q$ is trained for each variable $q$, by default from that variable alone. In Stage~2, a coordination module $g$ reads the histories and forecasts of all three variables at a sensor and corrects each forecast:
\begin{equation}
\hat{Y}'_i=\hat{Y}_i+g\bigl(\tilde{X}_i,\hat{Y}_i\bigr),
\label{eq:correction}
\end{equation}
where $\tilde{X}_i\in\mathbb{R}^{P\times Q}$ and $\hat{Y}_i\in\mathbb{R}^{H\times Q}$ are the histories and the forecasts at sensor $i$. The forecasters remain fixed while $g$ is trained, and $g$ shares its parameters across sensors. DeCo rests on three design choices, namely decoupled forecasters, coordination on forecasts, and out-of-fold training. Fig.~\ref{fig:detail} shows the main flow in detail and expands the coordination module, its variable mixing, and cross-fitting.

\input{figures/detail}

\textbf{Decoupled forecasters.} Our comparison in Section~\ref{sec:intro} shows that reading the other variables helps a forecaster, whereas predicting them jointly increases the error in most settings (Fig.~\ref{fig:intro}(c)). DeCo therefore trains a separate forecaster for each variable, optimized only for its own target. Cross-variable information enters through the coordination module rather than through the forecaster inputs. With the same forecasters, this outperforms early fusion (Section~\ref{sec:ablation}). As a complement, DeCo blends one early-fusion forecaster per variable into the output with a small fixed weight (Section~\ref{sec:inference}).

\textbf{Coordination on forecasts.} Each forecaster reads the history of its variable across the whole network and condenses it into an $H$-step forecast for every sensor, so spatial information reaches the module through the forecasts. The module can therefore operate at a single sensor with few parameters, relating the three forecasts to one another and to the recent observations.

\textbf{Out-of-fold training.} A module trained on the forecasters' in-sample predictions would learn to correct errors that are smaller than those at test time. Following stacked generalization~\cite{wolpert1992stacked}, DeCo trains the module on out-of-fold forecasts, whose targets are excluded from the corresponding forecaster fits (Section~\ref{sec:training}). All $K$ forecasters produced by this cross-fitting are reused at inference (Section~\ref{sec:inference}).

\subsection{Variable-Specific Forecasters}
\label{sec:decouplest}

In Stage~1, DeCo uses DecoupleST by default, an MLP-based forecaster that separates a network component from a local one. It can be replaced by any forecaster that outputs the forecast of one variable, such as RAGC in Section~\ref{sec:ragc}.

\textbf{Input embedding.} For variable $q$, DecoupleST embeds each of the $P$ entries of the recent window $\tilde{X}^{(q)}_i$ of sensor $i$ as
\begin{equation}
h_{i,\tau}=W_v\tilde{X}^{(q)}_{i,\tau}+b_v+c_i(t)+p_\tau,
\end{equation}
where $p_\tau$ is a learned step embedding and the context $c_i(t)$ combines learned embeddings of the sensor, the time of day, and the day of the week at the last observed step~$t$. A residual MLP then mixes the $P$ steps, yielding $\mathbf{H}\in\mathbb{R}^{N\times P\times d}$.

\textbf{Network and local branches.} The network branch aggregates $\mathbf{H}$ over sensors with a learned, input-independent adjacency $A$ constructed from sensor embeddings, and the local branch models the remainder:
\begin{equation}
\mathbf{Z}=\mathrm{FF}_{\mathrm{net}}(A\mathbf{H}),\qquad
\mathbf{L}=\mathrm{Local}\bigl(\mathbf{H}-R(\mathbf{Z})\bigr),
\end{equation}
where $\mathrm{FF}_{\mathrm{net}}$ is a residual feed-forward layer, $R$ is a linear map, and $\mathrm{Local}$ stacks two pairs of step-mixing and feed-forward layers. Subtracting $R(\mathbf{Z})$ leaves the local branch the part of each representation that the network branch does not reconstruct. Two MLP heads and a linear map from the input give the forecast in standardized units,
\begin{equation}
\bar{Y}^{(q)}=\mathrm{Head}_{\mathrm{net}}(\mathbf{Z})+\mathrm{Head}_{\mathrm{loc}}(\mathbf{L})+\mathrm{Direct}(\tilde{X}^{(q)}),
\end{equation}
which the mean and standard deviation of the forecaster's training data convert to $\hat{Y}^{(q)}$ in the original units.

\subsection{Coordination Module}
\label{sec:module}

The module corrects the three forecasts at one sensor (Fig.~\ref{fig:detail}), so this subsection omits the sensor index $i$. For variable $q$, let $\mu_q$ and $\sigma_q$ be the mean and standard deviation of its valid values in the training split. The module first encodes each variable from its standardized history and forecast,
\begin{equation}
z^{(q)}=\Bigl[\tfrac{\tilde{X}^{(q)}-\mu_q}{\sigma_q};\,\tfrac{\hat{Y}^{(q)}-\mu_q}{\sigma_q}\Bigr],\qquad
e^{(q)}=W_q z^{(q)}+b_q\in\mathbb{R}^{d}.
\end{equation}
Two mixer blocks with separate parameters follow. Each block first updates every variable with its own residual MLP,
\begin{equation}
e^{(q)}\leftarrow e^{(q)}+\mathrm{MLP}_q\bigl(\mathrm{LN}(e^{(q)})\bigr),
\end{equation}
and then mixes the three variables feature by feature:
\begin{equation}
e^{(q)}_j\leftarrow e^{(q)}_j+\phi_q\bigl(\bar{e}^{(q)}_j,\bar{e}^{(q')}_j,\bar{e}^{(q'')}_j\bigr),\qquad j=1,\dots,d,
\label{eq:cross}
\end{equation}
where $\bar{e}=\mathrm{LN}(e)$ is the layer-normalized representation, $q'$ and $q''$ are the other two variables, and $\phi_q:\mathbb{R}^3\to\mathbb{R}$ is a small MLP shared across features. A linear head per variable maps $e^{(q)}$ to a standardized correction $\delta^{(q)}\in\mathbb{R}^{H}$, which is rescaled to the original units:
\begin{equation}
\hat{Y}'^{(q)}=\hat{Y}^{(q)}+\sigma_q\,\delta^{(q)}.
\end{equation}
The heads are initialized at zero, so the untrained module leaves the forecasts unchanged.

\textbf{Target-only module.} Equation~\eqref{eq:cross} is the only place where information passes between variables. Replacing its three inputs with three copies of the target's own representation, $\phi_q(\bar{e}^{(q)}_j,\bar{e}^{(q)}_j,\bar{e}^{(q)}_j)$, yields a target-only module of the same size and training, in which no history or forecast of another variable reaches the target. Section~\ref{sec:experiments} uses this variant to isolate the contribution of the other variables.

\begin{algorithm}[t]
\caption{DeCo}
\label{alg:deco}
\begin{algorithmic}[1]
\REQUIRE training windows $\mathcal{D}$, validation windows $\mathcal{V}$, number of folds $K$, weight $w$
\STATE \textit{// training}
\STATE split $\mathcal{D}$ into contiguous blocks $\mathcal{D}_1,\dots,\mathcal{D}_K$
\FOR{each variable $q$ and fold $k$}
\STATE train $f^{(k)}_q$ on $\mathcal{D}\setminus\mathcal{D}_k$ \COMMENT{overlapping windows removed}
\STATE forecast $\mathcal{D}_k$ with $f^{(k)}_q$ \COMMENT{out-of-fold}
\ENDFOR
\STATE train the early-fusion forecasters $f^{\mathrm{e}}_q$ on $\mathcal{D}$
\STATE train $g$ on the out-of-fold forecasts with $\mathcal{L}$ in~\eqref{eq:loss} and select its checkpoint on $\mathcal{V}$
\STATE \textit{// inference at origin $t$}
\FOR{$k=1,\dots,K$}
\STATE $\hat{Y}^{(k)}\gets$ forecasts of $f^{(k)}_1,\dots,f^{(k)}_Q$
\STATE $\hat{Y}'^{(k)}\gets\hat{Y}^{(k)}+g\bigl(\tilde{X},\hat{Y}^{(k)}\bigr)$
\ENDFOR
\STATE $\hat{Y}^{\mathrm{e}}\gets$ forecasts of $f^{\mathrm{e}}_1,\dots,f^{\mathrm{e}}_Q$
\RETURN $(1-w)\,\frac{1}{K}\sum_{k}\hat{Y}'^{(k)}+w\,\hat{Y}^{\mathrm{e}}$
\end{algorithmic}
\end{algorithm}

\subsection{Training on Out-of-Fold Forecasts}
\label{sec:training}

\textbf{Cross-fitting.} The training windows are split into $K$ contiguous blocks (Fig.~\ref{fig:detail}). For each block $k$ and variable $q$, a fold forecaster $f^{(k)}_q$ is trained on the other blocks and forecasts the windows of block $k$. Every training window thus receives one forecast from a forecaster that has not seen its targets.

\textbf{Leakage control.} Because consecutive windows overlap in time, windows that overlap the held-out target times are removed from the fold's training set. The last 10\% of the remaining windows form an inner validation set for early stopping, and windows that overlap its target times are removed as well. When a fold fills missing inputs, it does not use values at held-out target times, and its training inputs also exclude inner-validation target times. Its scalers use only the observations permitted for fitting. The out-of-fold forecasts themselves are computed from the regular causally filled inputs.

\textbf{Objective.} Let $m^{(q)}_{t,h,i}=v^{(q)}_{t+h,i}\,\mathbf{1}[y^{(q)}_{t+h,i}\neq0]$ mark valid nonzero targets. With the out-of-fold forecasts as inputs, the module minimizes the MAE in standardized units, averaged over the variables,
\begin{equation}
\mathcal{L}=\frac{1}{Q}\sum_{q=1}^{Q}\frac{\sum_{t,h,i}m^{(q)}_{t,h,i}\,\bigl|\hat{y}'^{(q)}_{t,h,i}-y^{(q)}_{t+h,i}\bigr|/\sigma_q}{\sum_{t,h,i}m^{(q)}_{t,h,i}},
\label{eq:loss}
\end{equation}
over mini-batches of origin--sensor pairs.

\input{tables/main_table}

\subsection{Inference}
\label{sec:inference}

Cross-fitting yields $K$ trained forecasters per variable, and DeCo uses all of them. The module corrects the forecasts of every fold, and the corrected forecasts are averaged:
\begin{equation}
\hat{Y}^{\mathrm{avg}}=\frac{1}{K}\sum_{k=1}^{K}\Bigl[\hat{Y}^{(k)}+g\bigl(\tilde{X},\hat{Y}^{(k)}\bigr)\Bigr],
\label{eq:deploy}
\end{equation}
where $\hat{Y}^{(k)}$ collects the forecasts of the fold-$k$ forecasters of the three variables. Because the fold forecasters are already trained, the average adds no training.

\textbf{Early fusion.} The module coordinates the variables at each sensor. For a network-wide view of the raw inputs, DeCo also trains an early-fusion forecaster per variable, a DecoupleST in the MISO form that reads the histories of all three variables at all sensors. With its forecast $\hat{Y}^{\mathrm{e}}$, DeCo outputs
\begin{equation}
\hat{Y}^{\mathrm{DeCo}}=(1-w)\,\hat{Y}^{\mathrm{avg}}+w\,\hat{Y}^{\mathrm{e}},
\label{eq:blend}
\end{equation}
with a fixed weight $w$. Algorithm~\ref{alg:deco} summarizes training and inference, and Section~\ref{sec:sensitivity} varies $K$ and $w$.

\textbf{Efficiency.} The forecasters are independent and can be trained in parallel, and the coordination module is small. Section~\ref{sec:cost} reports training hours and latency.

%% file: figures/overview.tex
\begin{figure}[t]
\centering
\definecolor{iLflow}{HTML}{3C6AA4}
\definecolor{iLocc}{HTML}{D4A14A}
\definecolor{iLspeed}{HTML}{2F9779}
\definecolor{iLplum}{HTML}{A23B72}
\definecolor{iLink}{HTML}{222222}
\definecolor{iLmuted}{HTML}{6B6B6B}
\definecolor{iLred}{HTML}{C00000}
\definecolor{iLblue}{HTML}{2E75B6}
\definecolor{iLgreen}{HTML}{548235}
\definecolor{iLtitle}{HTML}{336699}
\definecolor{iLmodB}{HTML}{F0F3FA}
\definecolor{iLmodC}{HTML}{FFFCF4}
\definecolor{iLmodD}{HTML}{F5F5F5}
\definecolor{iLopc}{HTML}{FEF7E3}
\definecolor{iLfit}{HTML}{E6EAEF}
\tikzset{
  iLar/.style={-{Stealth[length=3.2pt,width=2.6pt]}, line width=0.5pt, draw=iLink},
  iLln/.style={line width=0.5pt, draw=iLink},
  iLdash/.style={dash pattern=on 2pt off 1.4pt},
  iLband/.style={draw=iLink!32, line width=0.6pt, rounded corners=3pt},
  iLtrain/.style={draw=iLred!42, fill=iLred!6, line width=0.5pt, rounded corners=2.5pt},
  iLinfer/.style={draw=iLgreen!55, fill=iLgreen!9, line width=0.5pt, rounded corners=2.5pt},
  iLpanel/.style={draw=iLink!20, fill=iLink!4, line width=0.4pt, rounded corners=2.5pt},
  iLinset/.style={draw=iLink!26, fill=white, line width=0.4pt, rounded corners=2.5pt},
  iLzoom/.style={draw=iLmuted!75, line width=0.4pt},
  iLtrn/.style={-{Stealth[length=3.2pt,width=2.6pt]}, draw=iLred, line width=0.55pt, dash pattern=on 2pt off 1.3pt},
  iLbox/.style={draw=iLink, line width=0.55pt, fill=white, align=center, inner xsep=2pt, inner ysep=1.5pt, minimum height=0.44cm},
  iLchip/.style={rounded corners=2pt, inner xsep=2pt, inner ysep=1.5pt, align=center, minimum height=0.36cm},
  iLop/.style={draw=iLink, line width=0.55pt, fill=white, align=center, inner sep=1.2pt, minimum height=0.34cm, minimum width=0.86cm},
  iLtab/.style={draw=iLink, line width=0.4pt, align=center, inner xsep=1.6pt, inner ysep=1.3pt, font=\fontsize{6}{7}\selectfont\itshape},
  iLlet/.style={circle, draw=iLink, line width=0.4pt, inner sep=0pt, minimum size=0.25cm, font=\fontsize{6}{7}\selectfont},
  iLtt/.style={font=\fontsize{7.5}{9}\selectfont\bfseries, inner sep=0.5pt},
  iLit/.style={font=\fontsize{7.5}{9}\selectfont\bfseries\itshape, inner sep=0.5pt},
  iLlab/.style={font=\scriptsize, inner sep=0.5pt},
  iLnote/.style={font=\fontsize{6}{7}\selectfont, inner sep=0.5pt},
  iLsum/.style={draw=iLink, circle, inner sep=0pt, minimum size=0.25cm, line width=0.5pt, fill=iLopc,
    path picture={\draw[iLink, line width=0.5pt] (path picture bounding box.north) -- (path picture bounding box.south)
      (path picture bounding box.west) -- (path picture bounding box.east);}},
  pics/iLflame/.style={code={\begin{scope}[scale=0.75]
    \fill[iLred] (0,0) .. controls (-0.15,0) and (-0.19,0.13) .. (-0.12,0.23)
      .. controls (-0.08,0.29) and (-0.05,0.31) .. (-0.055,0.38)
      .. controls (0.03,0.34) and (0.075,0.28) .. (0.07,0.21)
      .. controls (0.1,0.235) and (0.115,0.27) .. (0.11,0.3)
      .. controls (0.2,0.2) and (0.16,0.0) .. (0,0) -- cycle;
    \fill[white] (0.005,0.025) .. controls (-0.07,0.025) and (-0.09,0.09) .. (-0.055,0.14)
      .. controls (-0.035,0.17) and (-0.015,0.185) .. (-0.01,0.22)
      .. controls (0.05,0.18) and (0.085,0.12) .. (0.065,0.07)
      .. controls (0.055,0.04) and (0.03,0.025) .. (0.005,0.025) -- cycle;\end{scope}}},
  pics/iLsnowH/.style={code={\filldraw[fill=white, draw=iLblue!55, line width=0.35pt] (0,0) circle[radius=0.155]; \pic {iLsnow};}},
  pics/iLsnow/.style={code={
    \foreach \a in {90,150,210,270,330,30} {
      \draw[iLblue, line width=0.5pt, line cap=round] (0,0) -- (\a:0.13) (\a:0.075) -- ++(\a+50:0.045) (\a:0.075) -- ++(\a-50:0.045);}}},
  pics/iLsf/.style={code={\draw[#1] plot[smooth, tension=0.6] coordinates {(0,0.30) (0.12,0.36) (0.25,0.55) (0.38,0.80) (0.5,0.70) (0.62,0.50) (0.75,0.56) (0.88,0.74) (1,0.62)};}},
  pics/iLso/.style={code={\draw[#1] plot[smooth, tension=0.6] coordinates {(0,0.20) (0.12,0.24) (0.25,0.40) (0.38,0.64) (0.5,0.58) (0.62,0.36) (0.75,0.40) (0.88,0.56) (1,0.46)};}},
  pics/iLss/.style={code={\draw[#1] plot[smooth, tension=0.6] coordinates {(0,0.80) (0.12,0.78) (0.25,0.62) (0.38,0.38) (0.5,0.44) (0.62,0.66) (0.75,0.60) (0.88,0.44) (1,0.52)};}},
  pics/iLcub/.style={code={
    \fill[#1!55!white] (0,0.19) -- ++(0.07,0.055) -- ++(1.68,0) -- ++(-0.07,-0.055) -- cycle;
    \fill[#1!70!black!60] (1.68,0) -- ++(0.07,0.055) -- ++(0,0.19) -- ++(-0.07,-0.055) -- cycle;
    \foreach \j in {0,...,6} {\filldraw[fill=#1!35, draw=iLink, line width=0.4pt] (\j*0.24,0) rectangle ++(0.24,0.19);}
    \draw[iLink, line width=0.4pt] (0,0.19) -- ++(0.07,0.055) -- ++(1.68,0) -- ++(-0.07,-0.055) (1.68,0) -- ++(0.07,0.055) -- ++(0,0.19);}},
}
\tikzset{
  lB/.style={draw=iLink!32, line width=0.6pt, rounded corners=3pt},
  lC/.style={draw=iLink!32, line width=0.6pt, rounded corners=3pt},
  lE/.style={draw=iLink!32, line width=0.6pt, rounded corners=3pt},
  iLtrn/.style={-{Stealth[length=3.2pt,width=2.6pt]}, draw=iLink!75, line width=0.55pt, dash pattern=on 2pt off 1.3pt}}
\colorlet{iLdc}{iLred!78!black}
\begin{tikzpicture}[x=1cm, y=1cm, font=\footnotesize, line join=round, text=iLink,
  sTT/.style={font=\fontsize{8.5}{10}\selectfont\bfseries, inner sep=0.5pt},
  sHist/.style={draw=iLmuted, line width=0.5pt},
  sHistA/.style={sHist, -{Stealth[length=3.2pt,width=2.6pt]}},
  sReuse/.style={draw=iLink!65, line width=0.5pt, -{Stealth[length=3.0pt,width=2.4pt]}},
  pics/sBun/.style n args={3}{code={
    \draw[iLar, draw=iLflow, line width=0.55pt] (#1,#3+0.10) -- (#2,#3+0.10);
    \draw[iLar, draw=iLocc, line width=0.55pt] (#1,#3) -- (#2,#3);
    \draw[iLar, draw=iLspeed, line width=0.55pt] (#1,#3-0.10) -- (#2,#3-0.10);}},
  pics/sBunV/.style n args={3}{code={
    \draw[iLar, draw=iLflow, line width=0.55pt] (#1+0.10,#2) -- (#1+0.10,#3);
    \draw[iLar, draw=iLocc, line width=0.55pt] (#1,#2) -- (#1,#3);
    \draw[iLar, draw=iLspeed, line width=0.55pt] (#1-0.10,#2) -- (#1-0.10,#3);}},
  pics/sCards/.style n args={2}{code={
    \draw[iLspeed, line width=0.55pt, fill=white] (#1+0.14,#2+0.14) rectangle ++(0.84,0.44);
    \draw[iLocc, line width=0.55pt, fill=white] (#1+0.07,#2+0.07) rectangle ++(0.84,0.44);
    \draw[iLflow, line width=0.55pt, fill=white] (#1,#2) rectangle ++(0.84,0.44);
    \pic[shift={(#1+0.07,#2+0.03)}, xscale=0.70, yscale=0.40] {iLsf={iLflow, line width=0.8pt}};}},
  pics/sFore/.style n args={2}{code={
    \draw[iLspeed, line width=0.55pt, fill=white] (#1-0.65,#2-0.35) rectangle ++(1.44,0.78);
    \draw[iLocc, line width=0.55pt, fill=white] (#1-0.685,#2-0.39) rectangle ++(1.44,0.78);
    \draw[iLflow, line width=0.55pt, fill=white] (#1-0.72,#2-0.43) rectangle ++(1.44,0.78);
    \node[inner sep=0pt, align=center] at (#1,#2-0.04) {Forecasters\\[1pt]$f^{(k)}_q$};}}]
\useasboundingbox (0,0) rectangle (8.85,5.12);
\colorlet{iLopc}{white}\colorlet{iLmodC}{white}
\draw[iLflow, line width=0.7pt] (0.12,5.030) -- ++(0.42,0);
\draw[iLocc, line width=0.7pt] (0.12,4.950) -- ++(0.42,0);
\draw[iLspeed, line width=0.7pt] (0.12,4.870) -- ++(0.42,0);
\node[font=\scriptsize, anchor=west, inner sep=0pt] at (0.62,4.950) {flow, occupancy, speed};
\pic at (3.50,4.835) {iLflame};
\node[font=\scriptsize, anchor=west, inner sep=0pt] at (3.67,4.950) {trained};
\pic at (5.00,4.950) {iLsnow};
\node[font=\scriptsize, anchor=west, inner sep=0pt] at (5.19,4.950) {fixed};
\draw[iLtrn] (6.22,4.950) -- ++(0.50,0);
\node[font=\scriptsize, anchor=west, inner sep=0pt] at (6.80,4.950) {training only};
\draw[lB] (0.03,2.76) rectangle (8.82,4.73);
\draw[lE] (0.03,0.03) rectangle (8.82,2.61);
\node[sTT, text=iLink, anchor=base west] at (0.15,4.41) {Training};
\node[sTT, text=iLgreen, anchor=base west] at (0.15,2.29) {Inference};
\foreach \y in {3.75, 1.74} {
  \pic {sCards={0.18}{\y-0.29}};
  \pic {sBun={1.22}{1.66}{\y}};
  \pic {sFore={2.38}{\y}};
  \pic {sBun={3.22}{4.60}{\y}};
  \pic {sBun={6.25}{6.61}{\y}};
  \fill[iLmuted] (1.40,\y) circle[radius=0.04];
}
\draw[sHistA, rounded corners=1.5pt] (1.40,3.62) -- (1.40,3.04) -- (5.00,3.04) -- (5.00,3.32);
\node[fill=white, inner xsep=2pt, inner ysep=0pt, text=iLmuted] at (3.89,3.04) {histories $\tilde X$};
\node[anchor=south, inner sep=0pt] at (3.89,3.91) {out-of-fold};
\node[anchor=north, inner sep=0pt] at (3.89,3.60) {$\hat Y$};
\node[iLbox, fill=iLplum!14, draw=iLplum, line width=0.8pt, minimum width=1.65cm, minimum height=0.86cm] (sG1) at (5.425,3.75) {Coordination\\module $g$};
\pic at (1.63,4.14) {iLflame};
\node[iLtab, fill=white, draw=iLtitle, text=iLtitle, font=\scriptsize\itshape] at (2.415,4.46) {Stage 1};
\pic at (4.57,4.21) {iLflame};
\node[iLtab, fill=white, draw=iLplum, text=iLplum, font=\scriptsize\itshape] at (5.425,4.46) {Stage 2};
\filldraw[fill=white, draw=iLink, line width=0.5pt] (6.575,3.53) -- (7.115,3.53) -- (7.195,3.97) -- (6.655,3.97) -- cycle;
\node[inner sep=0pt] at (6.885,3.75) {$\hat Y'$};
\node[iLchip, fill=white, draw=iLink!70, line width=0.5pt, minimum width=0.40cm] (sL) at (8.22,3.75) {$\mathcal{L}$};
\draw[iLtrn] (7.175,3.75) -- (sL.west);
\node[inner sep=0.5pt] (sY) at (8.22,3.06) {$Y$};
\draw[iLtrn] (sY.north) -- (sL.south);
\draw[iLtrn] (sL.north) -- (8.22,4.46) -- (6.02,4.46) -- (6.02,4.18);
\draw[sHist, rounded corners=1.5pt] (1.40,1.61) -- (1.40,1.01) -- (5.00,1.01);
\draw[sHistA] (5.00,1.01) -- (5.00,1.31);
\draw[sHistA] (5.00,1.01) -- (5.00,0.69);
\fill[iLmuted] (5.00,1.01) circle[radius=0.04];
\node[iLbox, fill=iLplum!14, draw=iLplum, line width=0.8pt, minimum width=1.65cm, minimum height=0.86cm] (sG2) at (5.425,1.74) {Coordination\\module $g$};
\pic at (1.61,2.15) {iLsnowH};
\pic at (4.55,2.23) {iLsnowH};
\node[iLbox, fill=iLopc, minimum width=1.36cm, minimum height=0.72cm] (sM) at (7.29,1.74) {Mean over\\$K$ folds};
\node[iLsum, minimum size=0.42cm] (sP) at (7.29,0.47) {};
\pic {sBunV={7.29}{1.38}{0.72}};
\node[anchor=west, inner sep=0pt] at (7.46,1.06) {$(1{-}w)$};
\node[iLbox, fill=white, minimum width=1.96cm, minimum height=0.44cm] (sE) at (5.58,0.47) {Early fusion $f^{\mathrm{e}}_q$};
\pic at (4.54,0.76) {iLsnowH};
\pic {sBun={6.56}{7.04}{0.47}};
\node[anchor=south, inner sep=0pt] at (6.80,0.62) {$w$};
\pic {sBun={7.54}{7.79}{0.47}};
\node[anchor=west, inner sep=0pt] at (7.82,0.47) {$\hat Y^{\mathrm{DeCo}}$};
\draw[white, line width=2.4pt] (2.38,3.11) -- (2.38,2.97);
\draw[sReuse] (2.38,3.32) -- (2.38,2.20);
\draw[sReuse] (5.425,3.32) -- (5.425,2.19);
\end{tikzpicture}
\caption{Overview of DeCo. Stage~1 trains $K$ fold forecasters per variable, Stage~2 trains the coordination module $g$ on their out-of-fold forecasts, and inference averages the corrected forecasts over the folds and blends them with the early-fusion forecasts.}
\label{fig:overview}
\end{figure}
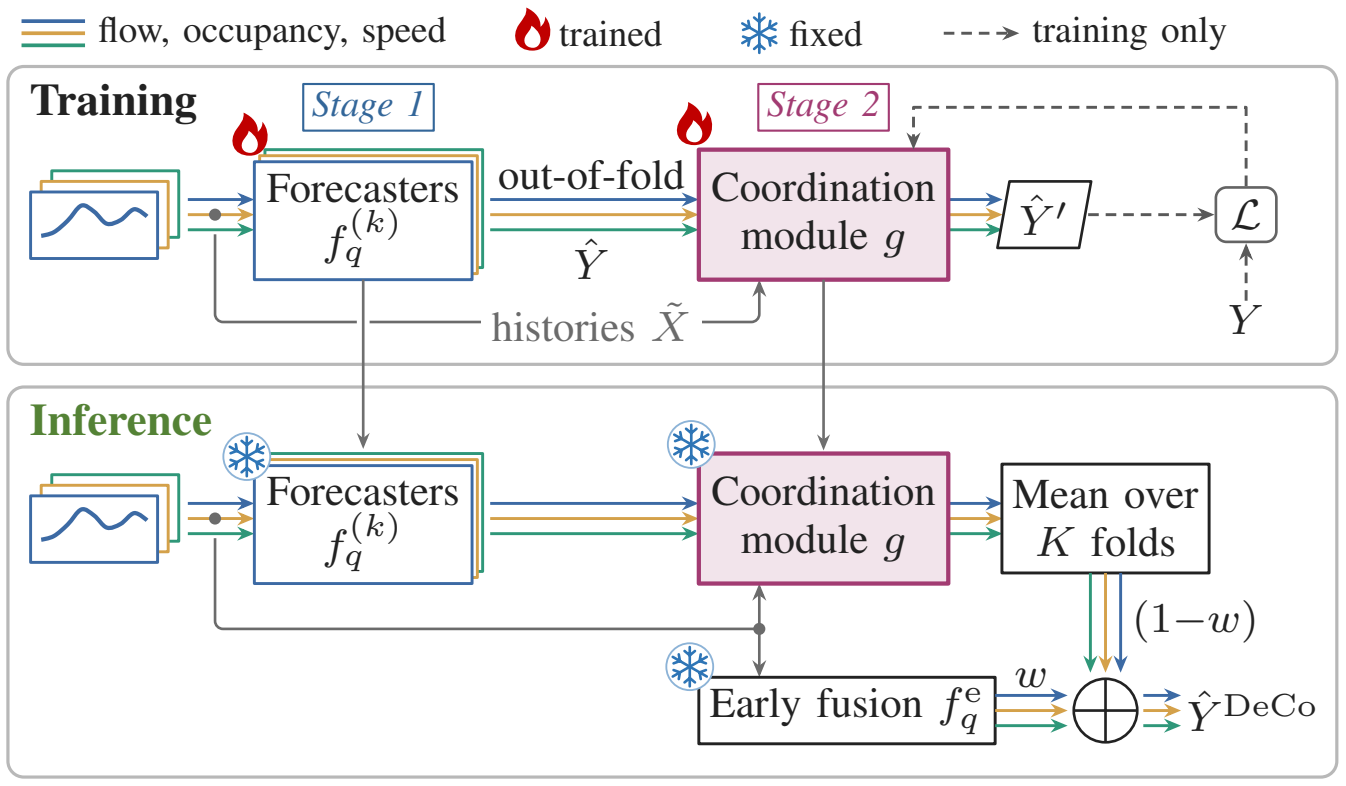

%% file: figures/detail.tex
\begin{figure*}[t]
\centering
\definecolor{iLflow}{HTML}{3C6AA4}
\definecolor{iLocc}{HTML}{D4A14A}
\definecolor{iLspeed}{HTML}{2F9779}
\definecolor{iLplum}{HTML}{A23B72}
\definecolor{iLink}{HTML}{222222}
\definecolor{iLmuted}{HTML}{6B6B6B}
\definecolor{iLred}{HTML}{C00000}
\definecolor{iLblue}{HTML}{2E75B6}
\definecolor{iLgreen}{HTML}{548235}
\definecolor{iLtitle}{HTML}{336699}
\definecolor{iLmodB}{HTML}{F0F3FA}
\definecolor{iLmodC}{HTML}{FFFCF4}
\definecolor{iLmodD}{HTML}{F5F5F5}
\definecolor{iLopc}{HTML}{FEF7E3}
\definecolor{iLfit}{HTML}{E6EAEF}
\tikzset{
  iLar/.style={-{Stealth[length=3.2pt,width=2.6pt]}, line width=0.5pt, draw=iLink},
  iLln/.style={line width=0.5pt, draw=iLink},
  iLdash/.style={dash pattern=on 2pt off 1.4pt},
  iLband/.style={draw=iLink!32, line width=0.6pt, rounded corners=3pt},
  iLtrain/.style={draw=iLred!42, fill=iLred!6, line width=0.5pt, rounded corners=2.5pt},
  iLinfer/.style={draw=iLgreen!55, fill=iLgreen!9, line width=0.5pt, rounded corners=2.5pt},
  iLpanel/.style={draw=iLink!20, fill=iLink!4, line width=0.4pt, rounded corners=2.5pt},
  iLinset/.style={draw=iLink!26, fill=white, line width=0.4pt, rounded corners=2.5pt},
  iLzoom/.style={draw=iLmuted!75, line width=0.4pt},
  iLtrn/.style={-{Stealth[length=3.2pt,width=2.6pt]}, draw=iLred, line width=0.55pt, dash pattern=on 2pt off 1.3pt},
  iLbox/.style={draw=iLink, line width=0.55pt, fill=white, align=center, inner xsep=2pt, inner ysep=1.5pt, minimum height=0.44cm},
  iLchip/.style={rounded corners=2pt, inner xsep=2pt, inner ysep=1.5pt, align=center, minimum height=0.36cm},
  iLop/.style={draw=iLink, line width=0.55pt, fill=white, align=center, inner sep=1.2pt, minimum height=0.34cm, minimum width=0.86cm},
  iLtab/.style={draw=iLink, line width=0.4pt, align=center, inner xsep=1.6pt, inner ysep=1.3pt, font=\fontsize{6}{7}\selectfont\itshape},
  iLlet/.style={circle, draw=iLink, line width=0.4pt, inner sep=0pt, minimum size=0.25cm, font=\fontsize{6}{7}\selectfont},
  iLtt/.style={font=\fontsize{7.5}{9}\selectfont\bfseries, inner sep=0.5pt},
  iLit/.style={font=\fontsize{7.5}{9}\selectfont\bfseries\itshape, inner sep=0.5pt},
  iLlab/.style={font=\scriptsize, inner sep=0.5pt},
  iLnote/.style={font=\fontsize{6}{7}\selectfont, inner sep=0.5pt},
  iLsum/.style={draw=iLink, circle, inner sep=0pt, minimum size=0.25cm, line width=0.5pt, fill=iLopc,
    path picture={\draw[iLink, line width=0.5pt] (path picture bounding box.north) -- (path picture bounding box.south)
      (path picture bounding box.west) -- (path picture bounding box.east);}},
  pics/iLflame/.style={code={\begin{scope}[scale=0.75]
    \fill[iLred] (0,0) .. controls (-0.15,0) and (-0.19,0.13) .. (-0.12,0.23)
      .. controls (-0.08,0.29) and (-0.05,0.31) .. (-0.055,0.38)
      .. controls (0.03,0.34) and (0.075,0.28) .. (0.07,0.21)
      .. controls (0.1,0.235) and (0.115,0.27) .. (0.11,0.3)
      .. controls (0.2,0.2) and (0.16,0.0) .. (0,0) -- cycle;
    \fill[white] (0.005,0.025) .. controls (-0.07,0.025) and (-0.09,0.09) .. (-0.055,0.14)
      .. controls (-0.035,0.17) and (-0.015,0.185) .. (-0.01,0.22)
      .. controls (0.05,0.18) and (0.085,0.12) .. (0.065,0.07)
      .. controls (0.055,0.04) and (0.03,0.025) .. (0.005,0.025) -- cycle;\end{scope}}},
  pics/iLsnowH/.style={code={\filldraw[fill=white, draw=iLblue!55, line width=0.35pt] (0,0) circle[radius=0.155]; \pic {iLsnow};}},
  pics/iLsnow/.style={code={
    \foreach \a in {90,150,210,270,330,30} {
      \draw[iLblue, line width=0.5pt, line cap=round] (0,0) -- (\a:0.13) (\a:0.075) -- ++(\a+50:0.045) (\a:0.075) -- ++(\a-50:0.045);}}},
  pics/iLsf/.style={code={\draw[#1] plot[smooth, tension=0.6] coordinates {(0,0.30) (0.12,0.36) (0.25,0.55) (0.38,0.80) (0.5,0.70) (0.62,0.50) (0.75,0.56) (0.88,0.74) (1,0.62)};}},
  pics/iLso/.style={code={\draw[#1] plot[smooth, tension=0.6] coordinates {(0,0.20) (0.12,0.24) (0.25,0.40) (0.38,0.64) (0.5,0.58) (0.62,0.36) (0.75,0.40) (0.88,0.56) (1,0.46)};}},
  pics/iLss/.style={code={\draw[#1] plot[smooth, tension=0.6] coordinates {(0,0.80) (0.12,0.78) (0.25,0.62) (0.38,0.38) (0.5,0.44) (0.62,0.66) (0.75,0.60) (0.88,0.44) (1,0.52)};}},
  pics/iLcub/.style={code={
    \fill[#1!55!white] (0,0.19) -- ++(0.07,0.055) -- ++(1.68,0) -- ++(-0.07,-0.055) -- cycle;
    \fill[#1!70!black!60] (1.68,0) -- ++(0.07,0.055) -- ++(0,0.19) -- ++(-0.07,-0.055) -- cycle;
    \foreach \j in {0,...,6} {\filldraw[fill=#1!35, draw=iLink, line width=0.4pt] (\j*0.24,0) rectangle ++(0.24,0.19);}
    \draw[iLink, line width=0.4pt] (0,0.19) -- ++(0.07,0.055) -- ++(1.68,0) -- ++(-0.07,-0.055) (1.68,0) -- ++(0.07,0.055) -- ++(0,0.19);}},
}
\begin{tikzpicture}[x=1cm, y=1cm, font=\scriptsize, line join=round, text=iLink,
  pP/.style={draw=iLink!32, line width=0.6pt, rounded corners=3pt},
  pD/.style={draw=iLink!45, fill=iLxf, line width=0.5pt, dash pattern=on 2.4pt off 1.4pt, rounded corners=2.5pt},
  lG/.style={fill=iLmodG, draw=iLplum, line width=0.8pt},
  lCm/.style={draw=iLplum, fill=iLmodG, line width=0.8pt},
  lPhi/.style={fill=iLphi, draw=iLvio, line width=0.6pt, rounded corners=2pt},
  lCp/.style={draw=iLvio, fill=iLphi, line width=0.6pt, rounded corners=2.5pt},
  lMlp/.style={fill=white},
  lXtab/.style={draw=iLink!75, line width=0.5pt, dash pattern=on 2pt off 1.2pt, fill=iLxf, text=iLink, align=center},
  lTtl/.style={font=\fontsize{7.5}{9}\selectfont\bfseries, inner xsep=3pt, inner ysep=1.2pt, minimum height=0.32cm},
  iLtrn/.style={-{Stealth[length=3.2pt,width=2.6pt]}, draw=iLink!75, line width=0.55pt, dash pattern=on 2pt off 1.3pt}]
\definecolor{iLvio}{HTML}{6F58B0}\definecolor{iLxf}{HTML}{EFEBE0}
\colorlet{iLmodG}{iLplum!14}\colorlet{iLphi}{iLvio!17}
\colorlet{iLopc}{white}\colorlet{iLmodB}{white}\colorlet{iLmodC}{white}\colorlet{iLmodD}{white}
\useasboundingbox (0,0) rectangle (17.80,7.49);
\pic at (6.60,7.185) {iLflame};
\node[iLnote, anchor=west] at (6.76,7.300) {trained};
\pic at (8.10,7.300) {iLsnow};
\node[iLnote, anchor=west] at (8.28,7.300) {fixed};
\draw[iLtrn] (9.35,7.300) -- (9.84,7.300);
\node[iLnote, anchor=west] at (9.90,7.300) {training only};
\begin{scope}[shift={(0.00,0.00)}]
\draw[iLband] (0.03,2.48) rectangle (2.25,7.07);%
\draw[pP] (2.40,5.02) rectangle (12.50,7.07);%
\draw[pP] (2.40,2.48) rectangle (6.90,4.87);%
\draw[pP] (7.05,2.48) rectangle (12.50,4.87);%
\fill[iLplum!8] (10.86,6.295) -- (12.66,7.02) -- (12.66,5.08) -- (10.86,5.705) -- cycle;
\fill[iLxf] (5.87,2.92) -- (6.63,2.92) -- (6.86,2.31) -- (5.38,2.31) -- cycle;
\foreach \nm/\q/\c/\s/\yc/\xb/\yr in {Flow/1/iLflow/iLsf/6.20/2.12/4.50, Occupancy/2/iLocc/iLso/4.88/2.00/4.03, Speed/3/iLspeed/iLss/3.56/1.88/3.56} {
  \node[iLlab, anchor=west] at (0.14,\yc+0.60) {\nm\ $\tilde X^{(\q)}$};
  \foreach \o in {0.14,0.07,0} {
    \draw[\c, line width=0.55pt, fill=white] (0.20+\o,\yc-0.31+\o) rectangle (1.62+\o,\yc+0.31+\o);}
  \fill[white] (0.20,\yc-0.31) rectangle (1.62,\yc+0.31);
  \draw[\c, line width=0.55pt] (0.20,\yc-0.31) rectangle (1.62,\yc+0.31);
  \pic[shift={(0.32,\yc-0.27)}, xscale=1.18, yscale=0.52] {\s={\c, line width=0.8pt}};
  \draw[iLar, draw=\c, line width=0.6pt] (1.76,\yc) -- (\xb,\yc) -- (\xb,\yr) -- (2.75,\yr);
}
\node[iLtt, text=iLink, anchor=west] at (0.13,2.655) {Input};

\node[iLtt, text=iLplum, anchor=west] at (2.52,6.87) {Stage 2: coordination};
\filldraw[fill=white, draw=iLink, line width=0.5pt] (2.62,6.18) -- (4.82,6.18) -- (4.97,6.62) -- (2.77,6.62) -- cycle;
\node[iLlab] at (3.80,6.40) {Histories $\tilde X_i$};
\filldraw[fill=white, draw=iLink, line width=0.5pt] (4.40,5.38) -- (6.85,5.38) -- (7.00,5.84) -- (4.55,5.84) -- cycle;
\node[iLlab] at (5.70,5.61) {Out-of-fold $\hat Y_i$};
\node[draw=iLink, circle, inner sep=0pt, minimum size=0.25cm, line width=0.5pt, fill=iLopc] (iLcat) at (7.30,6.00) {};
\draw[iLln] (7.275,5.93) -- (7.275,6.07) (7.325,5.93) -- (7.325,6.07);
\draw[iLar] (4.90,6.40) -- (7.30,6.40) -- (iLcat.north);
\draw[iLar] (6.93,5.61) -- (7.30,5.61) -- (iLcat.south);
\draw[iLar] (iLcat.east) -- (7.69,6.00);
\foreach \y/\c/\l in {6.22/iLflow/F, 6.00/iLocc/O, 5.78/iLspeed/S} {
  \fill[\c!70] (7.72,\y-0.085) rectangle (8.27,\y+0.085);
  \fill[\c!25] (8.27,\y-0.085) rectangle (8.82,\y+0.085);
  \draw[iLink, line width=0.4pt] (7.72,\y-0.085) rectangle (8.82,\y+0.085) (8.27,\y-0.085) -- (8.27,\y+0.085);
  \node[iLnote, anchor=west] at (8.84,\y) {\l};
}
\node[iLnote, anchor=base east] at (8.24,6.38) {history};
\node[iLnote, anchor=base west] at (8.31,6.38) {forecast};
\node[iLbox, lG, minimum width=1.60cm, minimum height=0.59cm] (iLgT) at (10.06,6.00) {Coordination\\[-2pt]module $g$};
\draw[iLar] (8.99,6.00) -- (iLgT.west);
\pic at (9.22,6.25) {iLflame};
\node[iLlab] (iLyp) at (11.33,6.00) {$\hat Y'_i$};
\draw[iLar] (iLgT.east) -- (iLyp.west);
\node[iLchip, fill=white, draw=iLink!70, line width=0.5pt, minimum width=0.36cm] (iLL) at (12.08,6.00) {$\mathcal{L}$};
\draw[iLtrn] (iLyp.east) -- (iLL.west);
\node[iLlab] (iLy) at (12.08,5.40) {$Y_i$};
\draw[iLtrn] (iLy.north) -- (iLL.south);
\draw[iLtrn] (iLL.north) -- (12.08,6.588) -- (10.55,6.588) -- (10.55,6.305);
\node[iLtt, text=iLtitle, anchor=west] at (2.52,2.655) {Stage 1: forecasting};
\foreach \nm/\q/\c/\y in {Flow/1/iLflow/4.50, Occupancy/2/iLocc/4.03, Speed/3/iLspeed/3.56} {
  \draw[\c, line width=0.55pt, fill=white] (2.83,\y-0.102) rectangle (4.78,\y+0.242);
  \draw[\c, line width=0.55pt, fill=white] (2.79,\y-0.137) rectangle (4.74,\y+0.207);
  \draw[\c, line width=0.55pt, fill=white] (2.75,\y-0.172) rectangle (4.70,\y+0.172);
  \node[iLlab] at (3.725,\y) {\nm\ $f_{\q}$};
  \draw[iLar, draw=\c, line width=0.6pt] (4.78,\y) -- (5.07,\y);
  \filldraw[fill=white, draw=\c, line width=0.55pt] (5.05,\y-0.172) -- (5.77,\y-0.172) -- (5.85,\y+0.172) -- (5.13,\y+0.172) -- cycle;
  \node[iLlab] at (5.45,\y) {$\hat Y^{(\q)}$};
  \draw[iLln] (5.81,\y) -- (6.25,\y);
}
\draw[iLln] (6.25,3.56) -- (6.25,4.50);
\draw[iLtrn] (6.25,4.50) -- (6.25,5.38);
\pic at (2.64,3.08) {iLflame};
\node[iLnote, anchor=west] at (2.82,3.21) {$K$ folds per variable};
\node[iLtab, lXtab] at (6.25,3.10) {Cross-\\fitting};

\node[iLtt, text=iLgreen, anchor=west] at (7.17,2.655) {Inference};
\draw[iLspeed, line width=0.55pt, fill=white] (7.34,4.155) rectangle (8.56,4.745);
\draw[iLocc, line width=0.55pt, fill=white] (7.295,4.11) rectangle (8.515,4.70);
\node[iLbox, draw=iLflow, fill=white, minimum width=1.22cm, minimum height=0.59cm] (iLff) at (7.86,4.36) {Fold\\[-2pt]forecasters};
\pic at (7.23,4.645) {iLsnowH};
\node[iLbox, lG, minimum width=1.60cm, minimum height=0.59cm] (iLgI) at (10.06,4.36) {Coordination\\[-2pt]module $g$};
\pic at (9.22,4.645) {iLsnowH};
\draw[iLar] (8.56,4.36) -- node[iLlab, anchor=north, yshift=-1.5pt] {$\hat Y^{(k)}$} (iLgI.west);
\node[iLbox, fill=iLmodC, minimum width=0.76cm, minimum height=0.59cm] (iLmean) at (12.00,4.36) {Mean\\[-2pt]over $k$};
\draw[iLar] (iLgI.east) -- node[iLlab, anchor=south, yshift=1.5pt] {$\hat Y'^{(k)}$} (iLmean.west);
\draw[iLar] (iLgT.south) -- (iLgI.north);
\node[iLnote, anchor=west] at (10.14,5.22) {trained $g$};
\filldraw[fill=white, draw=iLink, line width=0.5pt] (7.28,3.135) -- (7.84,3.135) -- (7.92,3.495) -- (7.36,3.495) -- cycle;
\node[iLlab] at (7.60,3.315) {$\tilde X$};
\draw[iLar] (7.60,3.495) -- (7.60,4.055);
\fill[iLink] (7.60,3.805) circle[radius=0.035];
\draw[iLar] (7.60,3.805) -- (10.06,3.805) -- (iLgI.south);
\node[iLbox, fill=white, minimum width=1.65cm, minimum height=0.59cm] (iLfe) at (9.05,3.315) {Early-fusion\\[-2pt]forecasters $f^{\mathrm{e}}_q$};
\pic at (8.185,3.60) {iLsnowH};
\draw[iLar] (7.88,3.315) -- (iLfe.west);
\node[iLsum] (iLplus) at (11.05,3.315) {};
\draw[iLar] (iLfe.east) -- (iLplus.west);
\node[iLlab, anchor=south] at (10.10,3.345) {$\hat Y^{\mathrm{e}}$};
\node[iLlab, anchor=south] at (10.62,3.345) {$\times w$};
\draw[iLar] (iLmean.south) -- (12.00,3.805) -- (11.05,3.805) -- (iLplus.north);
\node[iLlab, anchor=south] at (11.34,3.835) {$\hat Y^{\mathrm{avg}}$};
\node[iLlab, anchor=west] at (11.13,3.615) {$\times(1{-}w)$};
\draw[iLar] (iLplus.east) -- (11.50,3.315);
\node[iLlab, anchor=west] at (11.51,3.315) {$\hat Y^{\mathrm{DeCo}}$};
\end{scope}
\draw[lCm] (12.66,2.49) rectangle (17.76,7.06);
\begin{scope}[shift={(0.00,0.00)}]
\node[iLit, text=iLplum, anchor=west] at (12.80,6.86) {Coordination module $g$};
\filldraw[fill=white, draw=iLink, line width=0.45pt] (12.87,6.40) rectangle (13.59,6.66);
\filldraw[fill=white, draw=iLink, line width=0.45pt] (13.59,6.40) rectangle (14.39,6.66);
\node[iLnote] at (13.23,6.53) {history};
\node[iLnote] at (13.99,6.53) {forecast};
\node[iLbox, fill=iLmodB, minimum width=1.27cm, minimum height=0.36cm] (iLW) at (13.48,5.97) {Linear $W_q$};
\draw[iLar] (13.48,6.40) -- (iLW.north);
\draw[iLinset] (12.80,3.06) rectangle (15.40,5.38);
\node[iLsum] (iLs1) at (13.48,4.62) {};
\node[iLsum] (iLs2) at (13.48,3.72) {};
\draw[iLar] (iLW.south) -- (iLs1.north);
\draw[iLar] (iLs1.south) -- (iLs2.north);
\foreach \y/\yr/\s/\lab/\fc in {5.12/4.62/iLs1/{$\mathrm{MLP}_q$}/lMlp, 4.22/3.72/iLs2/{$\phi_q$}/lPhi} {
  \fill[iLink] (13.48,\y) circle[radius=0.035];
  \node[iLbox, font=\fontsize{6}{7}\selectfont, fill=iLmodD, minimum width=0.40cm, minimum height=0.24cm, inner xsep=1pt, inner ysep=1pt] (iLln\s) at (13.96,\y) {LN};
  \node[iLbox, \fc, minimum width=0.86cm, minimum height=0.36cm] (iLf\s) at (14.85,\y) {\lab};
  \draw[iLar] (13.48,\y) -- (iLln\s.west);
  \draw[iLar] (iLln\s.east) -- (iLf\s.west);
  \draw[iLar] (iLf\s.south) -- (14.85,\yr) -- (\s.east);
}
\node[iLnote, anchor=east] at (15.32,3.30) {Mixer block $\times2$};
\node[iLbox, fill=iLmodB, minimum width=1.33cm, minimum height=0.36cm] (iLhd) at (13.48,2.74) {Linear head};
\draw[iLar] (iLs2.south) -- (iLhd.north);
\node[iLsum] (iLs3) at (16.35,2.74) {};
\draw[iLar] (iLhd.east) -- (iLs3.west);
\node[iLlab] (iLyq) at (16.35,3.29) {$\hat Y^{(q)}$};
\draw[iLar] (16.35,3.16) -- (iLs3.north);
\draw[iLar] (iLs3.east) -- (16.84,2.74);
\node[iLlab, anchor=west] at (16.86,2.74) {$\hat Y'^{(q)}$};
\draw[iLinset] (15.52,5.03) rectangle (17.64,6.27);
\node[iLnote] at (16.58,5.16) {within a variable};
\draw[iLinset, draw=iLvio!70] (15.52,3.51) rectangle (17.64,4.79);
\node[iLnote] at (16.58,3.65) {across variables};
\foreach \ya/\yb/\c/\l in {5.99/4.51/iLflow/F, 5.70/4.22/iLocc/O, 5.41/3.93/iLspeed/S} {
  \foreach \y in {\ya,\yb} {
    \node[iLlet, fill=\c!35] at (15.74,\y) {\l};
    \pic[shift={(15.93,\y-0.095)}, xscale=0.92] {iLcub=\c};
  }
}
\draw[iLzoom] (iLfiLs1.east) -- (15.52,5.12);
\draw[iLzoom] (iLfiLs2.east) -- (15.52,4.22);
\draw[iLink!75, line width=0.8pt, rounded corners=1.5pt] (15.895,5.865) rectangle (17.575,6.165);
\draw[iLvio, line width=0.8pt, rounded corners=1.5pt] (17.01,3.81) rectangle (17.28,4.69);
\end{scope}
\draw[pD] (0.03,0.03) rectangle (8.55,2.33);
\draw[lCp] (8.70,0.03) rectangle (17.76,2.33);
\begin{scope}[shift={(0.00,0.00)}]
\node[lXtab, lTtl, font=\fontsize{7.5}{9}\selectfont\bfseries\itshape, minimum height=0.30cm, anchor=west] at (0.16,2.12) {Cross-fitting};
\node[iLlab, anchor=west] at (0.20,1.73) {fold forecaster $f^{(k)}_q$};
\node[iLbox, font=\fontsize{6}{7}\selectfont, fill=iLfit, minimum width=3.19cm, minimum height=0.32cm] (iLr1) at (1.795,1.30) {fit on the other blocks};
\node[iLbox, font=\fontsize{6}{7}\selectfont, fill=white, minimum width=3.19cm, minimum height=0.32cm] (iLr3) at (1.795,0.62) {forecast block $k$};
\draw[iLar] (iLr1.south) -- (iLr3.north);
\node[iLnote, anchor=east] at (8.26,2.13) {training windows};
\draw[iLink, line width=0.6pt] plot[smooth, tension=0.6] coordinates {(4.42,1.92) (4.74,1.97) (5.06,1.90) (5.38,1.94) (5.70,2.00) (6.02,1.92) (6.34,1.89) (6.66,1.97) (6.98,1.99) (7.30,1.91) (7.62,1.94) (7.94,2.00) (8.26,1.93)};
\foreach \k in {1,...,12} {
  \draw[iLink, line width=0.4pt, fill=white] (4.10+0.32*\k,1.60) rectangle ++(0.32,0.20);
}
\node[iLnote, anchor=east] at (4.34,1.70) {block};
\node[iLnote] at (4.58,1.70) {1};
\node[iLnote] at (4.90,1.70) {2};
\node[iLnote] at (8.10,1.70) {$K$};
\foreach \y/\h in {1.42/1, 1.21/2, 0.88/12} {\draw[iLar, line width=0.4pt] (4.26+0.32*\h,\y-0.08) -- (4.26+0.32*\h,0.55);}
\foreach \y/\h/\lab in {1.42/1/{fold $k{=}1$}, 1.21/2/{$k{=}2$}, 0.88/12/{$k{=}K$}} {
  \foreach \k in {1,...,12} {
    \draw[iLmuted, line width=0.35pt, fill=iLfit] (4.10+0.32*\k,\y-0.08) rectangle ++(0.32,0.16);}
  \draw[iLmuted, line width=0.35pt, fill=iLink!58] (4.10+0.32*\h,\y-0.08) rectangle ++(0.32,0.16);
  \node[iLnote, anchor=east] at (4.34,\y) {\lab};
}
\node[iLnote] at (6.34,1.045) {$\cdots$};
\foreach \k in {1,...,12} {
  \draw[iLink, line width=0.4pt, fill=white] (4.10+0.32*\k,0.37) rectangle ++(0.32,0.18);}
\node[iLnote] at (6.32,0.22) {out-of-fold forecasts};

\node[iLbox, lPhi, minimum width=0.56cm, minimum height=0.30cm, inner ysep=1pt, anchor=west] at (8.840,2.110) {$\phi_q$};
\node[iLit, text=iLvio!80!black, anchor=west] at (9.520,2.110) {Variable mixing};
\begin{scope}[yshift=-0.18cm]
\foreach \y/\c/\l in {1.42/iLflow/F, 1.03/iLocc/O, 0.64/iLspeed/S} {
  \node[iLlet, fill=\c!35] (iLc\l) at (9.02,\y) {\l};
  \node[iLbox, font=\fontsize{6}{7}\selectfont, fill=iLmodD, minimum width=0.40cm, minimum height=0.24cm, inner xsep=1pt, inner ysep=1pt] (iLn\l) at (9.62,\y) {LN};
  \draw[iLar] (iLc\l.east) -- (iLn\l.west);
  \draw[iLar] (iLn\l.east) -- (10.08,\y);
  \pic at (10.08,\y-0.095) {iLcub=\c};
  \filldraw[fill=\c!35, draw=iLink, line width=0.4pt] (12.25,\y-0.095) rectangle ++(0.24,0.19);
  \draw[iLar] (12.53,\y) -- (13.00,\y);
  \fill[\c] (13.06,\y) circle[radius=0.06];
}
\draw[iLvio, line width=0.8pt, rounded corners=1.5pt] (11.255,0.50) rectangle (11.545,1.60);
\draw[iLvio, line width=0.8pt, rounded corners=1.5pt] (12.205,0.50) rectangle (12.535,1.60);
\draw[iLar, draw=iLvio, line width=0.6pt] (11.40,1.60) -- (11.40,1.88) -- (12.37,1.88) -- (12.37,1.60);
\node[iLnote, text=iLvio!80!black] at (11.885,1.71) {feature $j$};
\foreach \yh in {1.51,1.27,1.03,0.79,0.55} {
  \foreach \yi in {1.42,1.03,0.64} {\draw[iLmuted, line width=0.35pt] (13.06,\yi) -- (13.80,\yh);}
  \draw[iLmuted, line width=0.35pt] (13.80,\yh) -- (14.54,1.03);
}
\foreach \yh in {1.51,1.27,1.03,0.79,0.55} {\filldraw[fill=white, draw=iLink, line width=0.4pt] (13.80,\yh) circle[radius=0.055];}
\fill[iLvio] (14.54,1.03) circle[radius=0.06];
\node[iLsum] (iLp2) at (15.20,1.03) {};
\draw[iLar] (14.60,1.03) -- (iLp2.west);
\node[iLnote, anchor=west] (iLej) at (15.50,0.58) {own feature $j$};
\draw[iLar] (15.50,0.58) -- (15.20,0.58) -- (iLp2.south);
\draw[iLar] (iLp2.east) -- (15.75,1.03);
\pic at (15.75,0.935) {iLcub=iLmuted};
\filldraw[fill=iLvio!55, draw=iLink, line width=0.4pt] (16.95,0.935) rectangle ++(0.24,0.19);
\end{scope}
\end{scope}
\end{tikzpicture}
\caption{DeCo in detail. White panels show the main flow. Each tinted panel expands the box of the same color, namely the coordination module, its variable mixing $\phi_q$, and cross-fitting.}
\label{fig:detail}
\end{figure*}
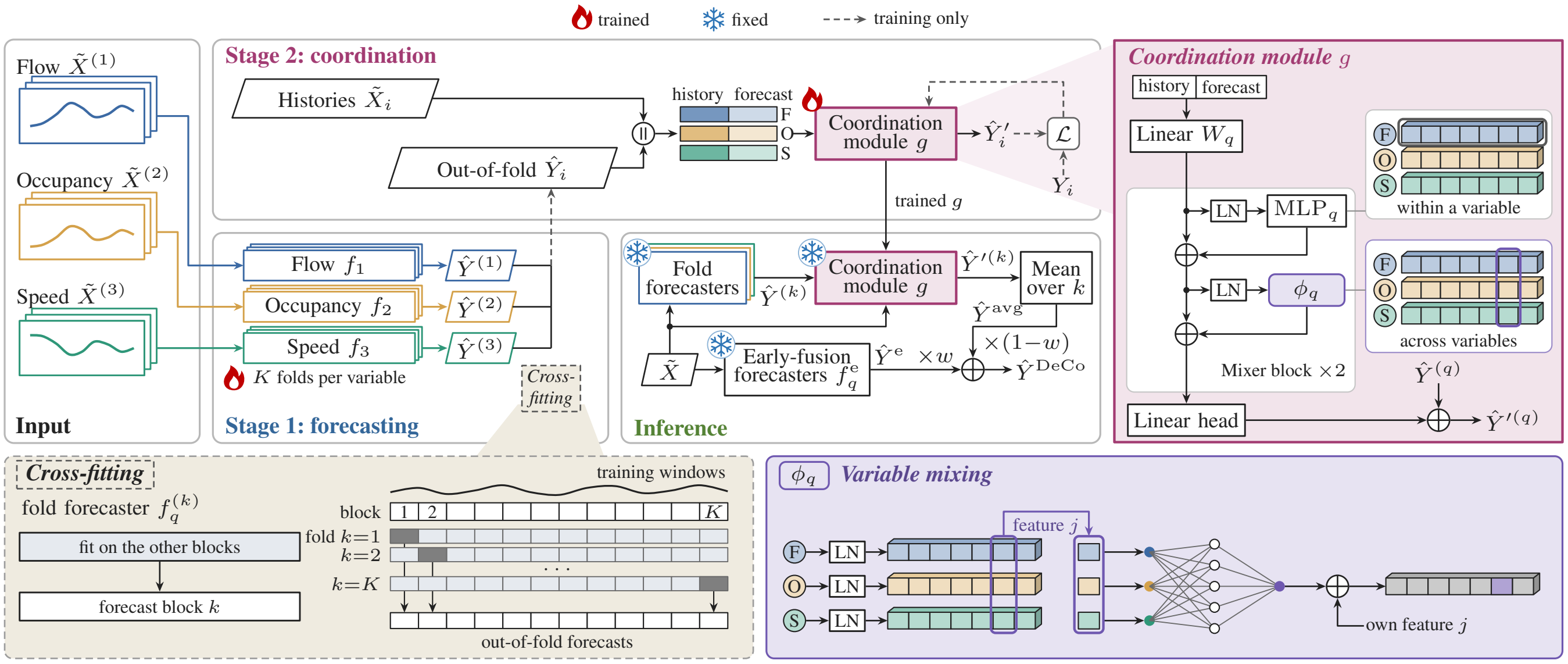

%% file: tables/main_table.tex
\begin{table*}[t]
\centering
\caption{Test MAE and RMSE (mean $\pm$ SD over three seeds). Bold marks the best value and underlining the second best. Ties at the displayed precision share a rank.}
\label{tab:main}
\scriptsize
\setlength{\tabcolsep}{1.5pt}
\setlength{\aboverulesep}{0.25ex}
\setlength{\belowrulesep}{0.25ex}
\begin{tabular}{llcccccccccccc}
\toprule
 & & \multicolumn{6}{c}{PeMS03} & \multicolumn{6}{c}{PeMS04}\\
\cmidrule(lr){3-8}\cmidrule(lr){9-14}
 & & \multicolumn{2}{c}{Flow} & \multicolumn{2}{c}{Occ.} & \multicolumn{2}{c}{Speed} & \multicolumn{2}{c}{Flow} & \multicolumn{2}{c}{Occ.} & \multicolumn{2}{c}{Speed}\\
\cmidrule(lr){3-4}\cmidrule(lr){5-6}\cmidrule(lr){7-8}\cmidrule(lr){9-10}\cmidrule(lr){11-12}\cmidrule(lr){13-14}
Model & Variant & MAE & RMSE & MAE & RMSE & MAE & RMSE & MAE & RMSE & MAE & RMSE & MAE & RMSE\\
\midrule
\multirow{3}{*}{STAEformer} & SISO & 15.42{\tiny$\pm$.13} & 26.65{\tiny$\pm$.11} & 0.958{\tiny$\pm$.006} & 2.519{\tiny$\pm$.025} & 1.970{\tiny$\pm$.032} & 4.048{\tiny$\pm$.044} & 19.13{\tiny$\pm$.06} & 31.64{\tiny$\pm$.17} & 0.744{\tiny$\pm$.002} & 1.977{\tiny$\pm$.006} & 1.586{\tiny$\pm$.004} & 3.627{\tiny$\pm$.031}\\
 & MISO & 15.22{\tiny$\pm$.05} & 26.52{\tiny$\pm$.08} & 0.954{\tiny$\pm$.013} & 2.541{\tiny$\pm$.043} & 1.913{\tiny$\pm$.008} & 3.927{\tiny$\pm$.019} & 19.05{\tiny$\pm$.06} & 31.45{\tiny$\pm$.16} & 0.743{\tiny$\pm$.001} & 1.982{\tiny$\pm$.004} & 1.562{\tiny$\pm$.002} & 3.580{\tiny$\pm$.015}\\
 & MIMO & 15.68{\tiny$\pm$.05} & 26.67{\tiny$\pm$.10} & 0.953{\tiny$\pm$.005} & 2.489{\tiny$\pm$.004} & 1.918{\tiny$\pm$.008} & 3.883{\tiny$\pm$.027} & 19.59{\tiny$\pm$.05} & 31.53{\tiny$\pm$.09} & 0.753{\tiny$\pm$.004} & 1.963{\tiny$\pm$.007} & 1.594{\tiny$\pm$.009} & 3.585{\tiny$\pm$.009}\\
\midrule
\multirow{3}{*}{HimNet} & SISO & 15.31{\tiny$\pm$.04} & 26.28{\tiny$\pm$.16} & 0.940{\tiny$\pm$.005} & 2.551{\tiny$\pm$.021} & 1.930{\tiny$\pm$.008} & 4.017{\tiny$\pm$.011} & 19.33{\tiny$\pm$.09} & 32.01{\tiny$\pm$.33} & 0.746{\tiny$\pm$.002} & 1.994{\tiny$\pm$.006} & 1.574{\tiny$\pm$.005} & 3.674{\tiny$\pm$.030}\\
 & MISO & 15.23{\tiny$\pm$.07} & 26.31{\tiny$\pm$.16} & 0.936{\tiny$\pm$.000} & 2.568{\tiny$\pm$.010} & 1.879{\tiny$\pm$.005} & 3.903{\tiny$\pm$.011} & 19.41{\tiny$\pm$.03} & 32.29{\tiny$\pm$.22} & 0.741{\tiny$\pm$.004} & 1.981{\tiny$\pm$.006} & 1.566{\tiny$\pm$.010} & 3.642{\tiny$\pm$.038}\\
 & MIMO & 15.49{\tiny$\pm$.15} & 26.26{\tiny$\pm$.31} & 0.938{\tiny$\pm$.006} & 2.517{\tiny$\pm$.030} & 1.894{\tiny$\pm$.009} & 3.890{\tiny$\pm$.033} & 19.82{\tiny$\pm$.11} & 32.19{\tiny$\pm$.11} & 0.750{\tiny$\pm$.003} & 1.982{\tiny$\pm$.008} & 1.571{\tiny$\pm$.002} & 3.606{\tiny$\pm$.014}\\
\midrule
\multirow{3}{*}{D$^2$STGNN} & SISO & 15.22{\tiny$\pm$.14} & 26.02{\tiny$\pm$.04} & 0.940{\tiny$\pm$.008} & 2.478{\tiny$\pm$.031} & 1.933{\tiny$\pm$.008} & 3.948{\tiny$\pm$.037} & 19.51{\tiny$\pm$.08} & 31.67{\tiny$\pm$.11} & 0.767{\tiny$\pm$.010} & 1.994{\tiny$\pm$.003} & 1.599{\tiny$\pm$.005} & 3.646{\tiny$\pm$.016}\\
 & MISO & 15.11{\tiny$\pm$.12} & 25.70{\tiny$\pm$.39} & 0.932{\tiny$\pm$.003} & 2.459{\tiny$\pm$.010} & 1.885{\tiny$\pm$.006} & 3.855{\tiny$\pm$.028} & 19.59{\tiny$\pm$.11} & 31.86{\tiny$\pm$.08} & 0.759{\tiny$\pm$.004} & 1.976{\tiny$\pm$.011} & 1.582{\tiny$\pm$.005} & 3.616{\tiny$\pm$.003}\\
 & MIMO & 15.75{\tiny$\pm$.24} & 26.46{\tiny$\pm$.40} & 0.935{\tiny$\pm$.008} & 2.451{\tiny$\pm$.021} & 1.902{\tiny$\pm$.020} & 3.865{\tiny$\pm$.050} & 20.76{\tiny$\pm$.16} & 32.97{\tiny$\pm$.19} & 0.769{\tiny$\pm$.003} & 1.989{\tiny$\pm$.003} & 1.627{\tiny$\pm$.009} & 3.636{\tiny$\pm$.020}\\
\midrule
\multirow{2}{*}{DecoupleST} & SISO & 14.80{\tiny$\pm$.02} & 24.07{\tiny$\pm$.05} & 0.928{\tiny$\pm$.001} & 2.507{\tiny$\pm$.011} & 1.935{\tiny$\pm$.006} & 4.011{\tiny$\pm$.023} & 19.10{\tiny$\pm$.10} & 31.57{\tiny$\pm$.14} & 0.739{\tiny$\pm$.001} & 1.959{\tiny$\pm$.005} & 1.574{\tiny$\pm$.005} & 3.592{\tiny$\pm$.016}\\
 & MISO & 14.71{\tiny$\pm$.08} & 24.08{\tiny$\pm$.42} & 0.922{\tiny$\pm$.003} & 2.489{\tiny$\pm$.021} & 1.894{\tiny$\pm$.003} & 3.900{\tiny$\pm$.007} & 18.95{\tiny$\pm$.03} & 31.20{\tiny$\pm$.09} & 0.736{\tiny$\pm$.001} & 1.948{\tiny$\pm$.011} & 1.560{\tiny$\pm$.007} & 3.539{\tiny$\pm$.011}\\
\midrule
RAGC & MISO & 14.56{\tiny$\pm$.05} & 23.10{\tiny$\pm$.03} & 0.888{\tiny$\pm$.002} & \textbf{2.352}{\tiny$\pm$.009} & 1.798{\tiny$\pm$.012} & \textbf{3.679}{\tiny$\pm$.031} & 19.00{\tiny$\pm$.04} & 30.88{\tiny$\pm$.05} & 0.737{\tiny$\pm$.003} & 1.950{\tiny$\pm$.014} & 1.529{\tiny$\pm$.001} & 3.539{\tiny$\pm$.007}\\
\midrule
\rowcolor{decoshade}\textbf{DeCo} & MIMO & \textbf{14.06}{\tiny$\pm$.02} & \textbf{22.94}{\tiny$\pm$.14} & \textbf{0.875}{\tiny$\pm$.001} & 2.400{\tiny$\pm$.007} & \textbf{1.791}{\tiny$\pm$.002} & 3.756{\tiny$\pm$.004} & \textbf{18.34}{\tiny$\pm$.04} & \underline{30.36}{\tiny$\pm$.10} & \textbf{0.709}{\tiny$\pm$.000} & \textbf{1.913}{\tiny$\pm$.001} & \textbf{1.491}{\tiny$\pm$.001} & \textbf{3.462}{\tiny$\pm$.003}\\
\rowcolor{decoshade}\textbf{DeCo (RAGC)} & MIMO & \underline{14.24}{\tiny$\pm$.03} & \underline{23.08}{\tiny$\pm$.17} & \underline{0.877}{\tiny$\pm$.002} & \underline{2.366}{\tiny$\pm$.007} & \underline{1.797}{\tiny$\pm$.007} & \underline{3.688}{\tiny$\pm$.008} & \underline{18.55}{\tiny$\pm$.01} & \textbf{30.33}{\tiny$\pm$.00} & \underline{0.716}{\tiny$\pm$.001} & \underline{1.926}{\tiny$\pm$.006} & \underline{1.508}{\tiny$\pm$.004} & \underline{3.478}{\tiny$\pm$.012}\\
\bottomrule
\end{tabular}
\par\medskip
\begin{tabular}{llcccccccccccc}
\toprule
 & & \multicolumn{6}{c}{PeMS07} & \multicolumn{6}{c}{PeMS08}\\
\cmidrule(lr){3-8}\cmidrule(lr){9-14}
 & & \multicolumn{2}{c}{Flow} & \multicolumn{2}{c}{Occ.} & \multicolumn{2}{c}{Speed} & \multicolumn{2}{c}{Flow} & \multicolumn{2}{c}{Occ.} & \multicolumn{2}{c}{Speed}\\
\cmidrule(lr){3-4}\cmidrule(lr){5-6}\cmidrule(lr){7-8}\cmidrule(lr){9-10}\cmidrule(lr){11-12}\cmidrule(lr){13-14}
Model & Variant & MAE & RMSE & MAE & RMSE & MAE & RMSE & MAE & RMSE & MAE & RMSE & MAE & RMSE\\
\midrule
\multirow{3}{*}{STAEformer} & SISO & 20.85{\tiny$\pm$.23} & 34.01{\tiny$\pm$.20} & 1.146{\tiny$\pm$.004} & 2.869{\tiny$\pm$.006} & 2.149{\tiny$\pm$.012} & 4.633{\tiny$\pm$.007} & \underline{14.18}{\tiny$\pm$.01} & 23.49{\tiny$\pm$.01} & 0.655{\tiny$\pm$.004} & 1.840{\tiny$\pm$.024} & 1.283{\tiny$\pm$.004} & 3.264{\tiny$\pm$.037}\\
 & MISO & 20.63{\tiny$\pm$.10} & 33.74{\tiny$\pm$.15} & 1.131{\tiny$\pm$.004} & 2.853{\tiny$\pm$.010} & 2.104{\tiny$\pm$.008} & 4.562{\tiny$\pm$.016} & 14.20{\tiny$\pm$.08} & 23.54{\tiny$\pm$.10} & 0.651{\tiny$\pm$.002} & 1.841{\tiny$\pm$.019} & 1.273{\tiny$\pm$.001} & 3.235{\tiny$\pm$.034}\\
 & MIMO & 21.45{\tiny$\pm$.24} & 34.58{\tiny$\pm$.33} & 1.160{\tiny$\pm$.011} & 2.880{\tiny$\pm$.017} & 2.228{\tiny$\pm$.027} & 4.680{\tiny$\pm$.035} & 15.18{\tiny$\pm$.09} & 24.51{\tiny$\pm$.11} & 0.661{\tiny$\pm$.000} & 1.800{\tiny$\pm$.007} & 1.306{\tiny$\pm$.007} & 3.233{\tiny$\pm$.024}\\
\midrule
\multirow{3}{*}{HimNet} & SISO & 20.82{\tiny$\pm$.02} & 33.89{\tiny$\pm$.03} & 1.137{\tiny$\pm$.002} & 2.875{\tiny$\pm$.007} & 2.168{\tiny$\pm$.003} & 4.663{\tiny$\pm$.010} & 14.71{\tiny$\pm$.06} & 23.83{\tiny$\pm$.03} & 0.662{\tiny$\pm$.003} & 1.876{\tiny$\pm$.033} & 1.309{\tiny$\pm$.007} & 3.377{\tiny$\pm$.075}\\
 & MISO & 20.63{\tiny$\pm$.06} & 33.72{\tiny$\pm$.15} & 1.131{\tiny$\pm$.004} & 2.874{\tiny$\pm$.009} & 2.130{\tiny$\pm$.006} & 4.611{\tiny$\pm$.020} & 14.69{\tiny$\pm$.10} & 23.76{\tiny$\pm$.12} & 0.657{\tiny$\pm$.002} & 1.848{\tiny$\pm$.021} & 1.299{\tiny$\pm$.005} & 3.333{\tiny$\pm$.025}\\
 & MIMO & 21.30{\tiny$\pm$.14} & 34.35{\tiny$\pm$.18} & 1.140{\tiny$\pm$.005} & 2.883{\tiny$\pm$.015} & 2.165{\tiny$\pm$.008} & 4.630{\tiny$\pm$.016} & 15.53{\tiny$\pm$.10} & 24.58{\tiny$\pm$.02} & 0.667{\tiny$\pm$.001} & 1.846{\tiny$\pm$.020} & 1.302{\tiny$\pm$.009} & 3.303{\tiny$\pm$.041}\\
\midrule
\multirow{3}{*}{D$^2$STGNN} & SISO & 21.17{\tiny$\pm$.05} & 34.41{\tiny$\pm$.15} & 1.144{\tiny$\pm$.006} & 2.854{\tiny$\pm$.013} & 2.186{\tiny$\pm$.008} & 4.606{\tiny$\pm$.019} & 15.19{\tiny$\pm$.03} & 24.44{\tiny$\pm$.08} & 0.666{\tiny$\pm$.007} & 1.837{\tiny$\pm$.017} & 1.324{\tiny$\pm$.008} & 3.323{\tiny$\pm$.043}\\
 & MISO & 20.91{\tiny$\pm$.16} & 33.91{\tiny$\pm$.17} & 1.135{\tiny$\pm$.002} & 2.835{\tiny$\pm$.016} & 2.171{\tiny$\pm$.012} & 4.561{\tiny$\pm$.020} & 15.19{\tiny$\pm$.09} & 24.40{\tiny$\pm$.07} & 0.667{\tiny$\pm$.009} & 1.816{\tiny$\pm$.010} & 1.310{\tiny$\pm$.010} & 3.224{\tiny$\pm$.022}\\
 & MIMO & 22.14{\tiny$\pm$.35} & 35.26{\tiny$\pm$.44} & 1.153{\tiny$\pm$.012} & 2.850{\tiny$\pm$.023} & 2.235{\tiny$\pm$.021} & 4.625{\tiny$\pm$.058} & 16.43{\tiny$\pm$.07} & 26.32{\tiny$\pm$.18} & 0.686{\tiny$\pm$.002} & 1.813{\tiny$\pm$.018} & 1.340{\tiny$\pm$.006} & 3.221{\tiny$\pm$.017}\\
\midrule
\multirow{2}{*}{DecoupleST} & SISO & 20.43{\tiny$\pm$.04} & 33.46{\tiny$\pm$.05} & 1.129{\tiny$\pm$.001} & 2.843{\tiny$\pm$.013} & 2.151{\tiny$\pm$.005} & 4.582{\tiny$\pm$.009} & 14.51{\tiny$\pm$.03} & 23.46{\tiny$\pm$.03} & 0.651{\tiny$\pm$.001} & 1.820{\tiny$\pm$.017} & 1.301{\tiny$\pm$.002} & 3.285{\tiny$\pm$.018}\\
 & MISO & 20.21{\tiny$\pm$.09} & 33.17{\tiny$\pm$.05} & 1.126{\tiny$\pm$.005} & 2.845{\tiny$\pm$.011} & 2.110{\tiny$\pm$.002} & 4.532{\tiny$\pm$.014} & 14.49{\tiny$\pm$.03} & 23.44{\tiny$\pm$.03} & 0.651{\tiny$\pm$.004} & 1.812{\tiny$\pm$.027} & 1.289{\tiny$\pm$.005} & 3.233{\tiny$\pm$.038}\\
\midrule
RAGC & MISO & 20.35{\tiny$\pm$.02} & 33.35{\tiny$\pm$.07} & 1.114{\tiny$\pm$.001} & 2.811{\tiny$\pm$.005} & 2.100{\tiny$\pm$.003} & 4.493{\tiny$\pm$.005} & 14.57{\tiny$\pm$.04} & 23.69{\tiny$\pm$.08} & 0.638{\tiny$\pm$.001} & 1.764{\tiny$\pm$.021} & 1.238{\tiny$\pm$.001} & \underline{3.080}{\tiny$\pm$.011}\\
\midrule
\rowcolor{decoshade}\textbf{DeCo} & MIMO & \textbf{19.36}{\tiny$\pm$.01} & \textbf{32.39}{\tiny$\pm$.01} & \textbf{1.077}{\tiny$\pm$.001} & \textbf{2.789}{\tiny$\pm$.004} & \textbf{2.021}{\tiny$\pm$.003} & \underline{4.445}{\tiny$\pm$.002} & \textbf{13.86}{\tiny$\pm$.00} & \textbf{22.89}{\tiny$\pm$.01} & \textbf{0.614}{\tiny$\pm$.000} & \underline{1.749}{\tiny$\pm$.006} & \textbf{1.218}{\tiny$\pm$.001} & 3.096{\tiny$\pm$.004}\\
\rowcolor{decoshade}\textbf{DeCo (RAGC)} & MIMO & \underline{19.68}{\tiny$\pm$.02} & \underline{32.64}{\tiny$\pm$.04} & \underline{1.083}{\tiny$\pm$.001} & \textbf{2.789}{\tiny$\pm$.003} & \underline{2.054}{\tiny$\pm$.001} & \textbf{4.441}{\tiny$\pm$.009} & 14.31{\tiny$\pm$.01} & \underline{23.39}{\tiny$\pm$.04} & \underline{0.625}{\tiny$\pm$.001} & \textbf{1.748}{\tiny$\pm$.002} & \underline{1.228}{\tiny$\pm$.003} & \textbf{3.064}{\tiny$\pm$.011}\\
\bottomrule
\end{tabular}
\end{table*}

%% file: sections/experiments.tex
\section{Experiments}
\label{sec:experiments}

We evaluate DeCo by answering seven research questions:
\begin{itemize}
\item RQ1: How accurate is DeCo compared with single-variable, auxiliary-input, and joint-output forecasters?
\item RQ2: What does each component of DeCo contribute, and what do the other variables add to the correction?
\item RQ3: How robust is DeCo to the number of fold forecasters and to the early-fusion weight?
\item RQ4: How does coordination perform with another forecaster?
\item RQ5: In which traffic states does coordination help most?
\item RQ6: Do the lower errors carry over to congestion warning and traffic state identification?
\item RQ7: How efficient is DeCo at inference and in training?
\end{itemize}

\subsection{Experimental Setting}
\label{sec:setting}

\textbf{Datasets and protocol.} We use the four three-variable benchmarks of Section~\ref{sec:data}, each split chronologically into training, validation, and test periods at 6:2:2. Every model receives the twelve most recent steps, forecasts the next twelve, and is evaluated on the same origins and valid targets. Invalid targets are excluded from training, model selection, and evaluation. We run three seeds, each an independent pipeline that retrains every forecaster and module, and report the mean $\pm$ standard deviation. Intervals are 95\% bootstrap intervals over forecast-origin days in the test period, with 10{,}000 resamples shared across methods and seeds.

\textbf{Baselines.} Table~\ref{tab:main} compares DeCo with five models. STAEformer~\cite{liu2023staeformer}, HimNet~\cite{dong2024himnet}, and D$^2$STGNN~\cite{shao2022d2stgnn} are each trained in the SISO, MISO, and MIMO forms of Section~\ref{sec:problem}. RAGC~\cite{ragc} is trained in the MISO form, with its official implementation extended to three input variables. DecoupleST is trained in the SISO form, which gives DeCo's single forecasters, and in the MISO form, where its input layers read all three variables. The table also lists DeCo~(RAGC), which replaces DeCo's forecasters with RAGC (Section~\ref{sec:ragc}). On PeMS04 and PeMS08, Table~\ref{tab:extra} adds 17 further baselines to these five models.

\textbf{Training.} All models of Table~\ref{tab:main}, including DeCo's forecasters, share one optimization recipe for their initial fits, so the three forms differ only in the variables a model reads and predicts. The loss is the MAE over valid nonzero targets in standardized units, following the masked loss common on these benchmarks. The optimizer is Adam with learning rate $10^{-3}$ and weight decay $10^{-4}$, without a learning-rate schedule, with gradient clipping at 5 and an effective batch size of 16. Training runs for at most 200 epochs with early stopping on the validation MAE at a patience of 30. Each baseline keeps one architecture configuration on all four datasets and all three variables. DeCo's forecasters then receive a second training phase of ten further epochs with the embedding-swap regularization of RAGC~\cite{ragc}. It replaces each sensor's embeddings with those of a randomly drawn sensor with probability 0.1, and the weights are averaged over these epochs. The coordination module is trained with AdamW at learning rate $5\times10^{-4}$ and weight decay $10^{-4}$ on mini-batches of 512 origin--sensor pairs. After every epoch, its corrected forecasts are scored on the validation split, for the fold average and for one reference forecaster per variable trained on the full training split. Training stops when neither score has improved for eight epochs or after 25 epochs, and the checkpoint with the best fold-average score is kept.

\textbf{Variants.} DeCo denotes the full system of Section~\ref{sec:method}, with $K=12$ fold forecasters per variable and an early-fusion weight of $w=0.2$. To isolate the coordination module, several analyses use DeCo-S, a single-forecaster variant built from the forecasters before their second training phase. Its module is trained on their out-of-fold forecasts and corrects the one forecaster per variable that is trained on the full training split, for which its checkpoint is selected. DeCo-S has no fold averaging and no early-fusion forecaster. The module is then the only difference between DeCo-S and the single forecasters, and the inputs from the other variables are the only difference between DeCo-S and its target-only counterpart.

\subsection{Overall Comparison (RQ1)}
\label{sec:main-results}

\input{tables/extra_baselines_table}

\input{figures/ablation_figure}
\input{figures/lead_figure}

Table~\ref{tab:main} reports test MAE and RMSE. We observe the following.

(1) DeCo attains the lowest MAE on every dataset and variable, up to 4.2\% below the best baseline. RAGC is the strongest baseline, and DeCo's MAE is up to 4.9\% below it.

(2) Reading the other variables helps, whereas predicting them jointly mostly does not. For STAEformer, HimNet, and D$^2$STGNN, the MISO form improves on the SISO form in most settings, whereas the MIMO form raises the error in most settings, by up to 8.2\% (Fig.~\ref{fig:intro}(c)). Coordinating the forecasts of single-variable forecasters gains more than the MISO form (DeCo-S, Fig.~\ref{fig:ablation}(b)), and DeCo is 3.2--5.6\% below DecoupleST MISO.

(3) DeCo also outperforms an ensemble of the strongest baseline. Against a fold average of RAGC of the same size (Section~\ref{sec:ragc}), its MAE is lower in almost every setting, by up to 3.3\%.

\input{figures/sensitivity_figure}

Table~\ref{tab:extra} extends the comparison on PeMS04 and PeMS08 with 17 further baselines, trained with the same splits, seeds, and targets: LightGBM~\cite{ke2017lightgbm}, DCRNN~\cite{li2018dcrnn}, N-BEATS~\cite{oreshkin2020nbeats}, DeepAR~\cite{salinas2020deepar}, AGCRN~\cite{bai2020agcrn}, MTGNN~\cite{wu2020mtgnn}, GTS~\cite{shang2021gts}, DSFormer~\cite{sun2025dsformer}, ST-MAN~\cite{he2023stman}, DST-Mamba~\cite{he2025dstmamba}, HyperD~\cite{shao2026hyperd}, and MTNet~\cite{zhang2026mtnet} in all three forms, and STGODE~\cite{fang2021stgode}, DMSTGCN~\cite{han2021dynamic}, MVC-STNet~\cite{miao2024mvcstnet}, AutoSTL~\cite{zhang2023autostl}, and MCST-Mamba~\cite{hamad2025mcstmamba} in the MISO and MIMO forms. DMSTGCN and MVC-STNet use batch size 64, and model-specific objectives, AutoSTL's bounded architecture search, and LightGBM's boosting protocol are retained. Each model is shown in its best form by test MAE, and none reaches DeCo's MAE on either dataset.

\subsection{Ablation Study (RQ2)}
\label{sec:ablation}

\textbf{Components.} Fig.~\ref{fig:ablation}(a) removes the components of DeCo. Each bar gives the increase in MAE over the full system on one dataset. The first two variants each remove one component, and the later ones remove components cumulatively. Without the coordination module, the MAE rises on every dataset and variable, by 1.35\% on average and most on PeMS07, and every interval excludes zero (Fig.~\ref{fig:ablation}(c)). Removing the early-fusion forecaster costs 0.34\% on average. The MAE increases with every removal on every dataset and variable.

\input{figures/states_figure}

\textbf{What the module uses.} Fig.~\ref{fig:ablation}(b) turns to the inputs of the module, with one forecaster per variable. (1) DeCo-S lowers the MAE of the single forecasters on every dataset and variable, by 0.7--3.2\%, and is also below their early-fusion form, DecoupleST MISO. (2) The target-only module recovers only part of this gain and raises the flow MAE on PeMS03 and PeMS08. DeCo-S improves on it in every setting, by 0.9--2.4\%, and every interval excludes zero (Fig.~\ref{fig:ablation}(c)). (3) To separate the two kinds of cross-variable input, we set either the histories or the forecasts of the other variables to zero in standardized coordinates and retrain the module. For DeCo-S, removing the forecasts raises the MAE by 0.99\% and removing the histories by 0.42\%, averaged over datasets and variables. On each dataset, removing the forecasts has the larger average cost.

\textbf{Forecast lead.} Fig.~\ref{fig:lead} resolves the comparison by lead. The reduction of DeCo-S grows with the lead. The target-only gain in occupancy and speed peaks at intermediate leads and then fades. The advantage of using the other variables therefore widens with the lead, to 1.4--3.2\% at 60 minutes.

\subsection{Sensitivity (RQ3)}
\label{sec:sensitivity}

Fig.~\ref{fig:sensitivity}(a) varies the number of fold forecasters. DeCo keeps $K$ of the twelve fold forecasters of each variable, trains the coordination module on their out-of-fold forecasts, and averages the $K$ corrected forecasts. DeCo stays accurate with fewer fold forecasters. Four keep the MAE within 1.1\% of $K=12$ at less than half of the training hours. Fig.~\ref{fig:sensitivity}(b) varies the early-fusion weight. For $w$ between 0.2 and 0.4, the MAE of every dataset changes by at most 0.11\%.

\subsection{Another Forecaster (RQ4)}
\label{sec:ragc}

\input{tables/ragc_table}

We replace DecoupleST with RAGC~\cite{ragc}, another group's forecaster, in its MISO form. DeCo~(RAGC) consists of RAGC fold forecasters and the coordination module alone (Table~\ref{tab:ragc}). It is below the single RAGC forecaster on every dataset and variable, by up to 3.3\%. (1) With one RAGC forecaster per variable and with the fold average of RAGC, the module lowers the MAE in most settings, by up to 1.9\%. (2) Coordination improves on the target-only module in every setting with one forecaster and in almost every setting with the fold average. (3) DeCo with DecoupleST forecasters remains below the fold average of RAGC in almost every setting.

\subsection{Where Coordination Helps Most (RQ5)}
\label{sec:states}

A sensor is congested at a step if its speed is at most the 5th percentile and its occupancy at least the 95th percentile of the training values of its dataset. A target is in free flow, at onset, in congestion, or in recovery when neither, only the target step, both, or only the last observed step is congested. We repeat the analysis under two looser definitions. One uses the 20th and 80th percentiles. The other uses the 35th and 65th percentiles, with fixed thresholds on PeMS04. Fig.~\ref{fig:states} shows how much DeCo-S lowers the speed MAE of the target-only module in each state, which is what the other variables add.

(1) The other variables help in every state. DeCo-S has a lower speed MAE than the target-only module in all four states, on every dataset and under every definition.

(2) Averaged over the datasets, the reduction at onset is 3.0--3.6\%. This is consistent with the fundamental diagram~\cite{li2011fundamental}. Below the critical density, speed stays near its free-flow level while occupancy rises, so the speed history alone carries little warning of a breakdown.

(3) Fig.~\ref{fig:intro}(d) compares both modules with the single forecasters under the strict definition. The gain is largest in congestion, where DeCo-S lowers the speed MAE by 8.2--12.2\% and the target-only module by 5.5--10.9\%. At onset and in free flow, DeCo-S lowers it on every dataset, whereas the target-only module raises it at onset on PeMS04 and PeMS08. In recovery, DeCo-S has the lower speed MAE of the two modules on every dataset.

\input{figures/sensors_figure}

\textbf{Across sensors.} The gain is broad across sensors (Fig.~\ref{fig:sensors}). DeCo-S lowers the speed MAE on 74--93\% of the sensors of each dataset, against 50--73\% for the target-only module.

\subsection{Downstream Use (RQ6)}
\label{sec:downstream}

\input{figures/downstream_figure}

We test whether the lower errors carry over to two uses of the forecasts. We apply the congestion rule of Section~\ref{sec:states} to each method's forecasts of speed and occupancy. This predicts, for every sensor and lead, whether the sensor will be congested. The rule reads both variables, so a SISO or MISO method uses its speed and occupancy models. An onset warning concerns a sensor that is not congested at the last observed step and predicts whether it will be congested at the lead. State identification assigns each target one of the four states of Section~\ref{sec:states}, with the predicted congestion in place of the future one. We report the F1 of the warning and the macro-F1 over the four states, onset included, at leads of 15, 30, and 60 minutes. Counts are pooled over the sensors and test days, and scores are averaged over the three seeds. Fig.~\ref{fig:downstream} uses the definition at the 20th and 80th percentiles. It compares DeCo on PeMS04 and PeMS08 with nine classical baselines of Table~\ref{tab:extra}.

(1) DeCo gives better warnings and better state identification than all nine baselines at all three leads. Its onset-warning F1 is 1.5--6.8 points above the best of them, and its state macro-F1 is 1.8--3.3 points above. This holds for every trained form of each baseline, on both datasets and at every lead.

(2) Coordination improves the warning, and target-only correction does not. With the same forecasters, DeCo-S raises the onset-warning F1 of the single forecasters on all four datasets and at every lead, by 1.1--1.5 points on average, whereas the target-only module lowers it. The gain of DeCo-S also holds under the strict and the loosest definition.

(3) The gain holds for recurrent and non-recurrent congestion. We call a five-minute slot of a sensor recurrent if the sensor was congested in it on at least half of the training days of the same type, weekday or weekend. Averaged over the datasets, DeCo-S adds about one to two points outside recurrent slots and about one point inside.

\subsection{Efficiency (RQ7)}
\label{sec:cost}

\input{figures/cost_figure}

On one NVIDIA H100, a network-wide forecast of all three variables takes 61\,ms on PeMS04 at batch size one, far below the five-minute sampling interval. The coordination module itself is light, with 0.11 million parameters. DeCo-S runs one forecaster per variable and one module call at inference. Most of the training time goes to the fold forecasters, which are independent and can be trained in parallel. Fewer fold forecasters lower the cost (Fig.~\ref{fig:cost}). With $K=4$, DeCo trains in about 60\% and 40\% of the hours of D$^2$STGNN in its MISO form on PeMS04 and PeMS08, while the MAE stays below every baseline on each variable. Latency is the median of 100 calls after warm-up, from GPU-resident inputs to the final forecast. Training hours sum epoch wall times, including validation and data access.

\subsection{Accuracy at Individual Leads}
\label{sec:leads}

\input{tables/lead_table}

Table~\ref{tab:lead} reports the test MAE at leads of 15, 30, and 60 minutes on PeMS04 and PeMS08 for the baselines of Table~\ref{tab:main} and DeCo. DeCo has the lowest MAE at every lead for every variable on both datasets. Its margin over the best baseline grows with the lead in every case, from 1.1--3.1\% at 15 minutes to 2.2--5.1\% at 60 minutes.

\subsection{Further Forecasters}
\label{sec:forecasters}

\input{tables/forecasters_table}

Section~\ref{sec:ragc} replaced DecoupleST with RAGC. Table~\ref{tab:forecasters} adds Graph WaveNet~\cite{wu2019graphwavenet}, STGCN~\cite{yu2018stgcn}, and STID~\cite{shao2022stid} on PeMS04 and PeMS08. As in~\eqref{eq:deploy}, the module is trained on the out-of-fold forecasts of twelve fold forecasters per variable and corrects the forecasts of every fold before they are averaged. It lowers the MAE of the fold average for every forecaster on every variable and dataset. These runs follow an earlier protocol, in which each forecaster is trained once with a shorter schedule, the forecasters and the module also read the windows one day, two days, and one week earlier, the module loss includes zero targets, and the three seeds vary the module.

\subsection{Case Study}
\label{sec:case}

\input{figures/case_figure}

Fig.~\ref{fig:case} follows the speed at one PeMS08 sensor on a test day. Between 07:00 and 07:40, the speed falls from above 62 to below 33\,mph while the occupancy rises from 5\% to 16\%. The single speed forecaster adapts slowly. Its 15-minute forecast stays more than 6\,mph above the measured speed until 09:00, and its 30-minute forecast more than 10\,mph until 09:20. DeCo, which also reads the occupancy and flow of the sensor, is within 3\,mph of the measured speed from 07:50 on at the 15-minute lead and within 5\,mph from 08:10 on at the 30-minute lead.

%% file: tables/extra_baselines_table.tex
\begin{table}[!htb]
\centering
\caption{Test MAE on PeMS04 and PeMS08 (mean over three seeds). Bold marks the lowest value in each column.}
\label{tab:extra}
\scriptsize
\setlength{\tabcolsep}{2.2pt}
\setlength{\aboverulesep}{1.0pt}%
\setlength{\belowrulesep}{1.0pt}
\renewcommand{\arraystretch}{0.965}%
\begin{tabular}{lcccccc}
\toprule
 & \multicolumn{3}{c}{PeMS04} & \multicolumn{3}{c}{PeMS08}\\
\cmidrule(lr){2-4}\cmidrule(lr){5-7}
Model & Flow & Occ. & Speed & Flow & Occ. & Speed\\
\midrule
STAEformer & 19.05$^b$ & 0.743$^b$ & 1.562$^b$ & 14.18$^a$ & 0.651$^b$ & 1.273$^b$\\
HimNet & 19.33$^a$ & 0.741$^b$ & 1.566$^b$ & 14.69$^b$ & 0.657$^b$ & 1.299$^b$\\
D$^2$STGNN & 19.51$^a$ & 0.759$^b$ & 1.582$^b$ & 15.19$^a$ & 0.666$^a$ & 1.310$^b$\\
DecoupleST & 18.95 & 0.736 & 1.560 & 14.49 & 0.651 & 1.289\\
RAGC & 19.00 & 0.737 & 1.529 & 14.57 & 0.638 & 1.238\\
\midrule
LightGBM & 21.26$^a$ & 0.818$^a$ & 1.695$^b$ & 15.77$^b$ & 0.688$^b$ & 1.335$^b$\\
DCRNN & 20.79$^b$ & 0.795$^b$ & 1.634$^b$ & 16.14$^b$ & 0.696$^b$ & 1.359$^b$\\
N-BEATS & 24.96$^b$ & 0.915$^b$ & 1.774$^b$ & 19.22$^b$ & 0.771$^b$ & 1.410$^b$\\
DeepAR & 19.77$^b$ & 0.803$^b$ & 1.667$^b$ & 14.58$^b$ & 0.688$^b$ & 1.355$^b$\\
AGCRN & 21.11$^b$ & 0.796$^c$ & 1.627$^b$ & 16.43$^c$ & 0.698$^c$ & 1.363$^c$\\
MTGNN & 20.42$^a$ & 0.784$^b$ & 1.618$^b$ & 15.94$^a$ & 0.698$^a$ & 1.366$^b$\\
MVC-STNet & 21.99$^b$ & 0.837$^c$ & 1.714$^b$ & 17.79$^c$ & 0.763$^c$ & 1.465$^b$\\
GTS & 22.39$^b$ & 0.819$^b$ & 1.660$^b$ & 16.70$^b$ & 0.721$^c$ & 1.368$^b$\\
STGODE & 21.61$^b$ & 0.808$^b$ & 1.651$^b$ & 16.64$^b$ & 0.717$^c$ & 1.369$^b$\\
DMSTGCN & 21.94$^b$ & 0.844$^b$ & 1.716$^b$ & 17.15$^b$ & 0.709$^b$ & 1.353$^b$\\
AutoSTL & 25.49$^b$ & 0.933$^c$ & 1.812$^b$ & 20.05$^b$ & 0.801$^b$ & 1.464$^b$\\
ST-MAN & 19.19$^b$ & 0.747$^b$ & 1.560$^b$ & 14.14$^b$ & 0.646$^b$ & 1.276$^b$\\
DST-Mamba & 22.52$^b$ & 0.835$^b$ & 1.743$^b$ & 17.66$^b$ & 0.723$^b$ & 1.419$^b$\\
MCST-Mamba & 20.77$^b$ & 0.829$^b$ & 1.886$^b$ & 15.81$^b$ & 0.744$^b$ & 1.567$^b$\\
DSFormer & 19.81$^b$ & 0.765$^b$ & 1.589$^b$ & 15.19$^b$ & 0.670$^b$ & 1.302$^b$\\
HyperD & 22.55$^b$ & 0.934$^b$ & 2.158$^b$ & 18.38$^b$ & 0.811$^b$ & 1.710$^b$\\
MTNet & 19.19$^b$ & 0.747$^a$ & 1.576$^b$ & 14.42$^b$ & 0.649$^b$ & 1.286$^b$\\
\midrule
\rowcolor{decoshade}\textbf{DeCo} & \textbf{18.34} & \textbf{0.709} & \textbf{1.491} & \textbf{13.86} & \textbf{0.614} & \textbf{1.218}\\
\rowcolor{decoshade}DeCo (RAGC) & 18.55 & 0.716 & 1.508 & 14.31 & 0.625 & 1.228\\
\bottomrule
\end{tabular}
\par\smallskip
\parbox{\columnwidth}{\scriptsize Each model is shown in its best trained form by test MAE ($^a$SISO, $^b$MISO, $^c$MIMO), three seeds. RAGC and DecoupleST are in the MISO form. Occupancy in \%.}
\end{table}

%% file: figures/ablation_figure.tex
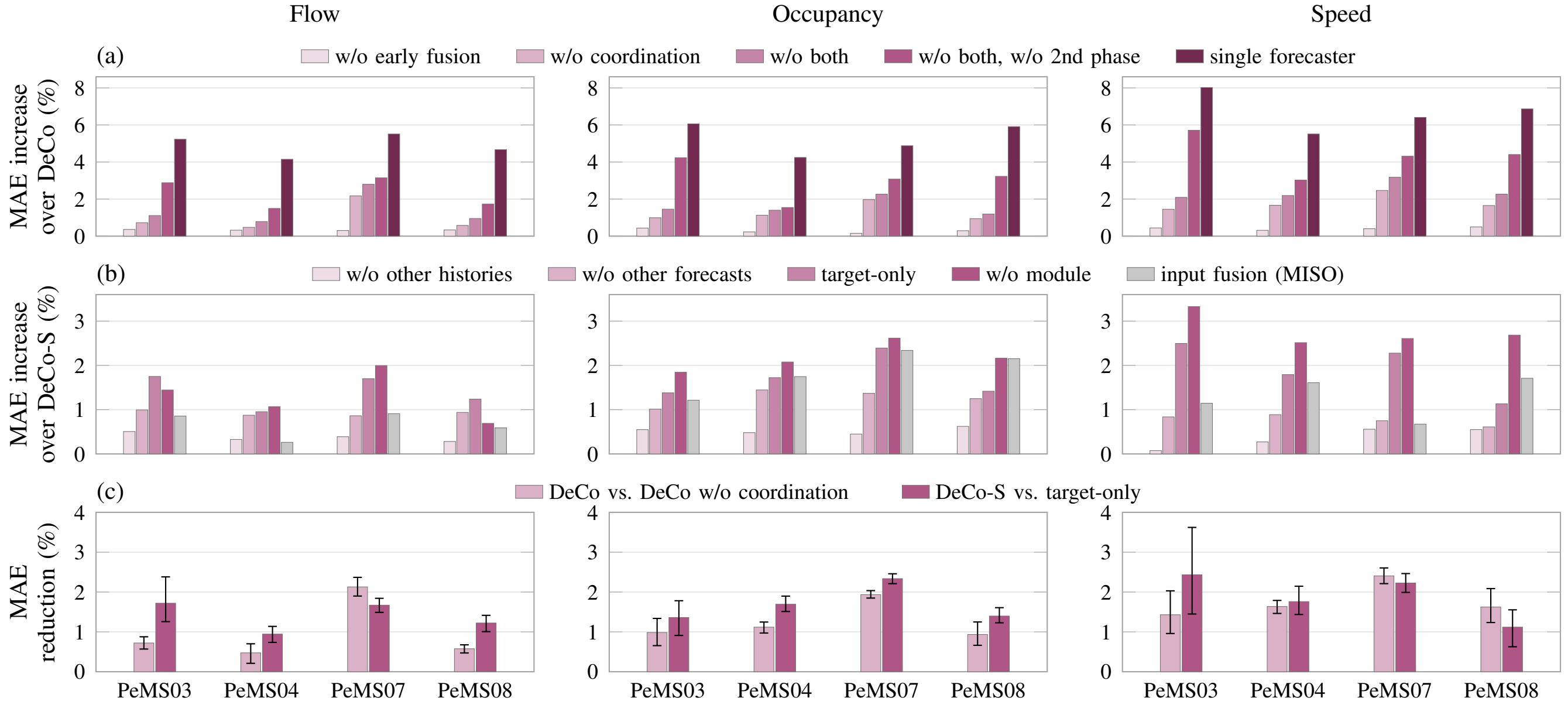
\begin{figure*}[t]
\centering
\def\dcsw#1{{\setlength{\fboxsep}{0pt}\setlength{\fboxrule}{0.2pt}\raisebox{0.012cm}{\fcolorbox{black!55}{#1}{\rule{0.30cm}{0pt}\rule{0pt}{0.14cm}}}}}
\begin{tikzpicture}
\begin{groupplot}[group style={group size=3 by 3, horizontal sep=0.85cm, vertical sep=0.66cm}, scale only axis, width=4.95cm, height=1.8cm,
  xmin=-0.55, xmax=3.55, xtick={0,1,2,3}, xticklabels=\empty, x tick style={draw=none}, ymin=0,
  title style={font=\footnotesize, yshift=-0.9ex, text depth=0.5ex}, ylabel style={align=center, yshift=-2pt}, ymajorgrids, grid style={black!9}, axis line style={black!35}, tick style={black!35}, tick label style={font=\scriptsize}, label style={font=\footnotesize}, clip=false]
\nextgroupplot[ymax=8.6, ytick={0,2,4,6,8}, yticklabels={0,2,4,6,8}, ylabel={MAE increase\\over DeCo (\%)}]
\addplot[ybar, bar width=3.7pt, bar shift=0pt, fill=cofacolor!18, draw=black!55, line width=0.2pt] coordinates {(-0.2387,0.361) (0.7613,0.322) (1.7613,0.305) (2.7613,0.335)};
\addplot[ybar, bar width=3.7pt, bar shift=0pt, fill=cofacolor!40, draw=black!55, line width=0.2pt] coordinates {(-0.1194,0.726) (0.8806,0.476) (1.8806,2.175) (2.8806,0.580)};
\addplot[ybar, bar width=3.7pt, bar shift=0pt, fill=cofacolor!62, draw=black!55, line width=0.2pt] coordinates {(0.0000,1.111) (1.0000,0.787) (2.0000,2.798) (3.0000,0.952)};
\addplot[ybar, bar width=3.7pt, bar shift=0pt, fill=cofacolor!86, draw=black!55, line width=0.2pt] coordinates {(0.1194,2.882) (1.1194,1.495) (2.1194,3.145) (3.1194,1.735)};
\addplot[ybar, bar width=3.7pt, bar shift=0pt, fill=cofacolor!70!black, draw=black!55, line width=0.2pt] coordinates {(0.2387,5.227) (1.2387,4.147) (2.2387,5.511) (3.2387,4.667)};
\nextgroupplot[ymax=8.6, ytick={0,2,4,6,8}, yticklabels={0,2,4,6,8}]
\addplot[ybar, bar width=3.7pt, bar shift=0pt, fill=cofacolor!18, draw=black!55, line width=0.2pt] coordinates {(-0.2387,0.436) (0.7613,0.230) (1.7613,0.154) (2.7613,0.294)};
\addplot[ybar, bar width=3.7pt, bar shift=0pt, fill=cofacolor!40, draw=black!55, line width=0.2pt] coordinates {(-0.1194,0.993) (0.8806,1.131) (1.8806,1.974) (2.8806,0.944)};
\addplot[ybar, bar width=3.7pt, bar shift=0pt, fill=cofacolor!62, draw=black!55, line width=0.2pt] coordinates {(0.0000,1.456) (1.0000,1.398) (2.0000,2.265) (3.0000,1.194)};
\addplot[ybar, bar width=3.7pt, bar shift=0pt, fill=cofacolor!86, draw=black!55, line width=0.2pt] coordinates {(0.1194,4.234) (1.1194,1.546) (2.1194,3.078) (3.1194,3.228)};
\addplot[ybar, bar width=3.7pt, bar shift=0pt, fill=cofacolor!70!black, draw=black!55, line width=0.2pt] coordinates {(0.2387,6.060) (1.2387,4.245) (2.2387,4.878) (3.2387,5.910)};
\nextgroupplot[ymax=8.6, ytick={0,2,4,6,8}, yticklabels={0,2,4,6,8}]
\addplot[ybar, bar width=3.7pt, bar shift=0pt, fill=cofacolor!18, draw=black!55, line width=0.2pt] coordinates {(-0.2387,0.444) (0.7613,0.315) (1.7613,0.401) (2.7613,0.498)};
\addplot[ybar, bar width=3.7pt, bar shift=0pt, fill=cofacolor!40, draw=black!55, line width=0.2pt] coordinates {(-0.1194,1.451) (0.8806,1.664) (1.8806,2.465) (2.8806,1.652)};
\addplot[ybar, bar width=3.7pt, bar shift=0pt, fill=cofacolor!62, draw=black!55, line width=0.2pt] coordinates {(0.0000,2.094) (1.0000,2.192) (2.0000,3.178) (3.0000,2.266)};
\addplot[ybar, bar width=3.7pt, bar shift=0pt, fill=cofacolor!86, draw=black!55, line width=0.2pt] coordinates {(0.1194,5.714) (1.1194,3.028) (2.1194,4.312) (3.1194,4.404)};
\addplot[ybar, bar width=3.7pt, bar shift=0pt, fill=cofacolor!70!black, draw=black!55, line width=0.2pt] coordinates {(0.2387,8.021) (1.2387,5.512) (2.2387,6.408) (3.2387,6.867)};
\nextgroupplot[ymax=3.6, ytick={0,1,2,3}, yticklabels={0,1,2,3}, ylabel={MAE increase\\over DeCo-S (\%)}]
\addplot[ybar, bar width=3.7pt, bar shift=0pt, fill=cofacolor!18, draw=black!55, line width=0.2pt] coordinates {(-0.2387,0.506) (0.7613,0.327) (1.7613,0.389) (2.7613,0.282)};
\addplot[ybar, bar width=3.7pt, bar shift=0pt, fill=cofacolor!40, draw=black!55, line width=0.2pt] coordinates {(-0.1194,0.994) (0.8806,0.874) (1.8806,0.862) (2.8806,0.938)};
\addplot[ybar, bar width=3.7pt, bar shift=0pt, fill=cofacolor!62, draw=black!55, line width=0.2pt] coordinates {(0.0000,1.750) (1.0000,0.953) (2.0000,1.699) (3.0000,1.238)};
\addplot[ybar, bar width=3.7pt, bar shift=0pt, fill=cofacolor!86, draw=black!55, line width=0.2pt] coordinates {(0.1194,1.444) (1.1194,1.068) (2.1194,1.996) (3.1194,0.690)};
\addplot[ybar, bar width=3.7pt, bar shift=0pt, fill=black!22, draw=black!55, line width=0.2pt] coordinates {(0.2387,0.856) (1.2387,0.262) (2.2387,0.909) (3.2387,0.589)};
\nextgroupplot[ymax=3.6, ytick={0,1,2,3}, yticklabels={0,1,2,3}]
\addplot[ybar, bar width=3.7pt, bar shift=0pt, fill=cofacolor!18, draw=black!55, line width=0.2pt] coordinates {(-0.2387,0.549) (0.7613,0.481) (1.7613,0.449) (2.7613,0.622)};
\addplot[ybar, bar width=3.7pt, bar shift=0pt, fill=cofacolor!40, draw=black!55, line width=0.2pt] coordinates {(-0.1194,1.012) (0.8806,1.445) (1.8806,1.369) (2.8806,1.249)};
\addplot[ybar, bar width=3.7pt, bar shift=0pt, fill=cofacolor!62, draw=black!55, line width=0.2pt] coordinates {(0.0000,1.380) (1.0000,1.723) (2.0000,2.392) (3.0000,1.417)};
\addplot[ybar, bar width=3.7pt, bar shift=0pt, fill=cofacolor!86, draw=black!55, line width=0.2pt] coordinates {(0.1194,1.845) (1.1194,2.077) (2.1194,2.616) (3.1194,2.163)};
\addplot[ybar, bar width=3.7pt, bar shift=0pt, fill=black!22, draw=black!55, line width=0.2pt] coordinates {(0.2387,1.214) (1.2387,1.747) (2.2387,2.338) (3.2387,2.152)};
\nextgroupplot[ymax=3.6, ytick={0,1,2,3}, yticklabels={0,1,2,3}]
\addplot[ybar, bar width=3.7pt, bar shift=0pt, fill=cofacolor!18, draw=black!55, line width=0.2pt] coordinates {(-0.2387,0.075) (0.7613,0.272) (1.7613,0.558) (2.7613,0.550)};
\addplot[ybar, bar width=3.7pt, bar shift=0pt, fill=cofacolor!40, draw=black!55, line width=0.2pt] coordinates {(-0.1194,0.836) (0.8806,0.887) (1.8806,0.750) (2.8806,0.611)};
\addplot[ybar, bar width=3.7pt, bar shift=0pt, fill=cofacolor!62, draw=black!55, line width=0.2pt] coordinates {(0.0000,2.495) (1.0000,1.792) (2.0000,2.277) (3.0000,1.132)};
\addplot[ybar, bar width=3.7pt, bar shift=0pt, fill=cofacolor!86, draw=black!55, line width=0.2pt] coordinates {(0.1194,3.329) (1.1194,2.514) (2.1194,2.607) (3.1194,2.684)};
\addplot[ybar, bar width=3.7pt, bar shift=0pt, fill=black!22, draw=black!55, line width=0.2pt] coordinates {(0.2387,1.145) (1.2387,1.609) (2.2387,0.673) (3.2387,1.709)};
\nextgroupplot[xticklabels={PeMS03,PeMS04,PeMS07,PeMS08}, ymax=4, ytick={0,1,2,3,4}, yticklabels={0,1,2,3,4}, ylabel={MAE\\reduction (\%)}]
\addplot[ybar, bar width=6.4pt, bar shift=0pt, fill=cofacolor!40, draw=black!55, line width=0.2pt, error bars/.cd, y dir=both, y explicit, error bar style={black, line width=0.4pt}, error mark options={rotate=90, mark size=1.4pt, line width=0.4pt, black}] coordinates {(-0.1019,0.7208) += (0,0.1561) -= (0,0.1518) (0.8981,0.4736) += (0,0.2272) -= (0,0.2630) (1.8981,2.1289) += (0,0.2397) -= (0,0.2296) (2.8981,0.5765) += (0,0.0983) -= (0,0.1058)};
\addplot[ybar, bar width=6.4pt, bar shift=0pt, fill=cofacolor!86, draw=black!55, line width=0.2pt, error bars/.cd, y dir=both, y explicit, error bar style={black, line width=0.4pt}, error mark options={rotate=90, mark size=1.4pt, line width=0.4pt, black}] coordinates {(0.1019,1.7200) += (0,0.6603) -= (0,0.4630) (1.1019,0.9445) += (0,0.1933) -= (0,0.2091) (2.1019,1.6708) += (0,0.1720) -= (0,0.1817) (3.1019,1.2229) += (0,0.1922) -= (0,0.2169)};
\nextgroupplot[xticklabels={PeMS03,PeMS04,PeMS07,PeMS08}, ymax=4, ytick={0,1,2,3,4}, yticklabels={0,1,2,3,4}]
\addplot[ybar, bar width=6.4pt, bar shift=0pt, fill=cofacolor!40, draw=black!55, line width=0.2pt, error bars/.cd, y dir=both, y explicit, error bar style={black, line width=0.4pt}, error mark options={rotate=90, mark size=1.4pt, line width=0.4pt, black}] coordinates {(-0.1019,0.9832) += (0,0.3545) -= (0,0.3289) (0.8981,1.1187) += (0,0.1282) -= (0,0.1476) (1.8981,1.9353) += (0,0.1006) -= (0,0.0869) (2.8981,0.9349) += (0,0.3142) -= (0,0.2722)};
\addplot[ybar, bar width=6.4pt, bar shift=0pt, fill=cofacolor!86, draw=black!55, line width=0.2pt, error bars/.cd, y dir=both, y explicit, error bar style={black, line width=0.4pt}, error mark options={rotate=90, mark size=1.4pt, line width=0.4pt, black}] coordinates {(0.1019,1.3608) += (0,0.4211) -= (0,0.4501) (1.1019,1.6941) += (0,0.2034) -= (0,0.1833) (2.1019,2.3363) += (0,0.1222) -= (0,0.1261) (3.1019,1.3975) += (0,0.2101) -= (0,0.1699)};
\nextgroupplot[xticklabels={PeMS03,PeMS04,PeMS07,PeMS08}, ymax=4, ytick={0,1,2,3,4}, yticklabels={0,1,2,3,4}]
\addplot[ybar, bar width=6.4pt, bar shift=0pt, fill=cofacolor!40, draw=black!55, line width=0.2pt, error bars/.cd, y dir=both, y explicit, error bar style={black, line width=0.4pt}, error mark options={rotate=90, mark size=1.4pt, line width=0.4pt, black}] coordinates {(-0.1019,1.4301) += (0,0.6000) -= (0,0.4710) (0.8981,1.6369) += (0,0.1525) -= (0,0.1742) (1.8981,2.4055) += (0,0.1997) -= (0,0.1949) (2.8981,1.6249) += (0,0.4615) -= (0,0.3902)};
\addplot[ybar, bar width=6.4pt, bar shift=0pt, fill=cofacolor!86, draw=black!55, line width=0.2pt, error bars/.cd, y dir=both, y explicit, error bar style={black, line width=0.4pt}, error mark options={rotate=90, mark size=1.4pt, line width=0.4pt, black}] coordinates {(0.1019,2.4346) += (0,1.1880) -= (0,0.9865) (1.1019,1.7600) += (0,0.3860) -= (0,0.3240) (2.1019,2.2266) += (0,0.2372) -= (0,0.2354) (3.1019,1.1193) += (0,0.4343) -= (0,0.4931)};
\end{groupplot}
\node[anchor=base, inner sep=0, font=\footnotesize] at ($(group c1r1.north)+(0cm,0.62cm)$) {Flow};
\node[anchor=base, inner sep=0, font=\footnotesize] at ($(group c2r1.north)+(0cm,0.62cm)$) {Occupancy};
\node[anchor=base, inner sep=0, font=\footnotesize] at ($(group c3r1.north)+(0cm,0.62cm)$) {Speed};
\node[anchor=base west, inner sep=0, font=\footnotesize] at ($(group c1r1.north west)+(0cm,0.15cm)$) {(a)};
\node[anchor=base, inner sep=0, font=\scriptsize] at ($(group c2r1.north)+(0cm,0.15cm)$) {\dcsw{cofacolor!18}\hspace{0.07cm}w/o early fusion\hspace{0.4cm}\dcsw{cofacolor!40}\hspace{0.07cm}w/o coordination\hspace{0.4cm}\dcsw{cofacolor!62}\hspace{0.07cm}w/o both\hspace{0.4cm}\dcsw{cofacolor!86}\hspace{0.07cm}w/o both, w/o 2nd phase\hspace{0.4cm}\dcsw{cofacolor!70!black}\hspace{0.07cm}single forecaster};
\node[anchor=base west, inner sep=0, font=\footnotesize] at ($(group c1r2.north west)+(0cm,0.15cm)$) {(b)};
\node[anchor=base, inner sep=0, font=\scriptsize] at ($(group c2r2.north)+(0cm,0.15cm)$) {\dcsw{cofacolor!18}\hspace{0.07cm}w/o other histories\hspace{0.4cm}\dcsw{cofacolor!40}\hspace{0.07cm}w/o other forecasts\hspace{0.4cm}\dcsw{cofacolor!62}\hspace{0.07cm}target-only\hspace{0.4cm}\dcsw{cofacolor!86}\hspace{0.07cm}w/o module\hspace{0.4cm}\dcsw{black!22}\hspace{0.07cm}input fusion (MISO)};
\node[anchor=base west, inner sep=0, font=\footnotesize] at ($(group c1r3.north west)+(0cm,0.15cm)$) {(c)};
\node[anchor=base, inner sep=0, font=\scriptsize] at ($(group c2r3.north)+(0cm,0.15cm)$) {\dcsw{cofacolor!40}\hspace{0.07cm}DeCo vs.\ DeCo w/o coordination\hspace{0.6cm}\dcsw{cofacolor!86}\hspace{0.07cm}DeCo-S vs.\ target-only};
\end{tikzpicture}
\caption{Ablation (test MAE, three-seed mean). (a)~MAE increase (\%) over DeCo as its components are removed, down to a single forecaster. (b)~MAE increase over DeCo-S, which uses one forecaster per variable, when the histories or the forecasts of the other variables are set to zero, when the module sees the target variable alone, and when it is removed. The gray bar is the early-fusion forecaster (MISO). (c)~MAE reduction of DeCo over DeCo without coordination and of DeCo-S over the target-only module, with 95\% intervals.}
\label{fig:ablation}
\end{figure*}

%% file: figures/lead_figure.tex
\begin{figure}[t]
\centering
\begin{tikzpicture}
\begin{groupplot}[group style={group size=3 by 1, horizontal sep=0.26cm}, scale only axis, width=2.34cm, height=2.35cm,
  xmin=2.5, xmax=62.5, ymin=-1, ymax=4.4, xtick={15,30,45,60}, xticklabels={15,30,45,60}, ytick={0,2,4}, yticklabels={0,2,4},
  xlabel style={yshift=2pt}, ylabel style={yshift=-2pt}, title style={font=\footnotesize, yshift=-0.9ex, text depth=0.5ex}, ymajorgrids, grid style={black!9}, axis line style={black!35}, tick style={black!35}, tick label style={font=\scriptsize}, label style={font=\footnotesize}, clip=false]
\nextgroupplot[title={Flow}, ylabel={MAE reduction (\%)}]
\addplot[black!45, line width=0.4pt, forget plot] coordinates {(2.5,0) (62.5,0)};
\addplot[name path=Uself0, draw=none, forget plot] coordinates {(5,0.084) (10,0.161) (15,0.152) (20,0.271) (25,0.291) (30,0.355) (35,0.383) (40,0.285) (45,0.248) (50,0.364) (55,0.407) (60,0.406)};
\addplot[name path=Lself0, draw=none, forget plot] coordinates {(5,-0.188) (10,-0.379) (15,-0.408) (20,-0.406) (25,-0.454) (30,-0.589) (35,-0.661) (40,-0.740) (45,-0.665) (50,-0.658) (55,-0.634) (60,-0.641)};
\addplot[black!13, forget plot] fill between[of=Uself0 and Lself0];
\addplot[name path=Udeco0, draw=none, forget plot] coordinates {(5,0.873) (10,1.147) (15,1.367) (20,1.680) (25,1.799) (30,1.928) (35,2.088) (40,2.114) (45,2.266) (50,2.359) (55,2.577) (60,2.769)};
\addplot[name path=Ldeco0, draw=none, forget plot] coordinates {(5,0.239) (10,0.234) (15,0.307) (20,0.736) (25,0.744) (30,0.699) (35,0.743) (40,0.777) (45,0.820) (50,0.863) (55,0.863) (60,0.997)};
\addplot[cofacolor!24, forget plot] fill between[of=Udeco0 and Ldeco0];
\addplot[black!58, line width=0.9pt, dash pattern=on 2.6pt off 1.5pt, forget plot] coordinates {(5,-0.031) (10,-0.050) (15,-0.118) (20,-0.060) (25,-0.093) (30,-0.082) (35,-0.105) (40,-0.170) (45,-0.165) (50,-0.148) (55,-0.133) (60,-0.138)};
\addplot[cofacolor, line width=1.15pt, forget plot] coordinates {(5,0.526) (10,0.689) (15,0.869) (20,1.117) (25,1.117) (30,1.221) (35,1.341) (40,1.433) (45,1.588) (50,1.656) (55,1.681) (60,1.777)};
\nextgroupplot[title={Occupancy}, yticklabels=\empty, xlabel={Lead (min)}]
\addplot[black!45, line width=0.4pt, forget plot] coordinates {(2.5,0) (62.5,0)};
\addplot[name path=Uself1, draw=none, forget plot] coordinates {(5,0.444) (10,0.875) (15,1.042) (20,1.239) (25,1.324) (30,1.112) (35,0.982) (40,0.748) (45,0.643) (50,0.343) (55,0.373) (60,0.415)};
\addplot[name path=Lself1, draw=none, forget plot] coordinates {(5,0.139) (10,0.156) (15,0.209) (20,0.221) (25,0.286) (30,0.251) (35,0.285) (40,0.246) (45,0.232) (50,0.185) (55,0.183) (60,0.057)};
\addplot[black!13, forget plot] fill between[of=Uself1 and Lself1];
\addplot[name path=Udeco1, draw=none, forget plot] coordinates {(5,1.284) (10,1.790) (15,2.064) (20,2.224) (25,2.386) (30,2.573) (35,2.640) (40,2.728) (45,2.837) (50,3.014) (55,3.093) (60,3.130)};
\addplot[name path=Ldeco1, draw=none, forget plot] coordinates {(5,0.958) (10,1.296) (15,1.546) (20,1.647) (25,1.659) (30,1.730) (35,1.787) (40,1.977) (45,2.260) (50,2.100) (55,2.094) (60,1.915)};
\addplot[cofacolor!24, forget plot] fill between[of=Udeco1 and Ldeco1];
\addplot[black!58, line width=0.9pt, dash pattern=on 2.6pt off 1.5pt, forget plot] coordinates {(5,0.273) (10,0.448) (15,0.541) (20,0.638) (25,0.661) (30,0.604) (35,0.577) (40,0.479) (45,0.414) (50,0.238) (55,0.237) (60,0.192)};
\addplot[cofacolor, line width=1.15pt, forget plot] coordinates {(5,1.110) (10,1.506) (15,1.875) (20,2.033) (25,2.115) (30,2.194) (35,2.235) (40,2.329) (45,2.439) (50,2.399) (55,2.407) (60,2.355)};
\nextgroupplot[title={Speed}, yticklabels=\empty]
\addplot[black!45, line width=0.4pt, forget plot] coordinates {(2.5,0) (62.5,0)};
\addplot[name path=Uself2, draw=none, forget plot] coordinates {(5,0.264) (10,0.594) (15,1.316) (20,1.871) (25,2.117) (30,2.099) (35,1.975) (40,1.764) (45,1.643) (50,1.463) (55,1.247) (60,1.152)};
\addplot[name path=Lself2, draw=none, forget plot] coordinates {(5,0.004) (10,0.071) (15,0.220) (20,0.265) (25,0.379) (30,0.389) (35,0.375) (40,0.386) (45,0.367) (50,0.358) (55,0.361) (60,0.325)};
\addplot[black!13, forget plot] fill between[of=Uself2 and Lself2];
\addplot[name path=Udeco2, draw=none, forget plot] coordinates {(5,1.169) (10,1.830) (15,2.488) (20,2.879) (25,3.082) (30,3.181) (35,3.278) (40,3.547) (45,3.632) (50,3.822) (55,3.912) (60,3.871)};
\addplot[name path=Ldeco2, draw=none, forget plot] coordinates {(5,0.549) (10,1.200) (15,1.987) (20,2.225) (25,2.359) (30,2.465) (35,2.598) (40,2.603) (45,2.701) (50,2.697) (55,2.710) (60,2.704)};
\addplot[cofacolor!24, forget plot] fill between[of=Udeco2 and Ldeco2];
\addplot[black!58, line width=0.9pt, dash pattern=on 2.6pt off 1.5pt, forget plot] coordinates {(5,0.147) (10,0.432) (15,0.769) (20,0.994) (25,1.130) (30,1.111) (35,1.049) (40,0.967) (45,0.892) (50,0.810) (55,0.708) (60,0.638)};
\addplot[cofacolor, line width=1.15pt, forget plot] coordinates {(5,0.880) (10,1.518) (15,2.141) (20,2.556) (25,2.755) (30,2.845) (35,2.880) (40,2.954) (45,3.023) (50,3.088) (55,3.119) (60,3.140)};
\end{groupplot}
\begin{scope}[shift={($(group c2r1.north)+(-1.95cm,0.745cm)$)}, font=\scriptsize]
\fill[cofacolor!24] (0,-0.09) rectangle (0.5,0.09); \draw[cofacolor, line width=1.15pt] (0,0) -- (0.5,0);
\node[anchor=base west,inner sep=0] at (0.58,-0.085) {DeCo-S};
\fill[black!13] (2.05,-0.09) rectangle (2.55,0.09); \draw[black!58, line width=0.9pt, dash pattern=on 2.6pt off 1.5pt] (2.05,0) -- (2.55,0);
\node[anchor=base west,inner sep=0] at (2.63,-0.085) {Target-only};
\end{scope}
\end{tikzpicture}
\caption{MAE reduction (\%) over the single forecasters by forecast lead. Solid lines are DeCo-S and dashed lines the target-only module. Each line is the mean over the four datasets, and its band spans the lowest and highest dataset.}
\label{fig:lead}
\end{figure}
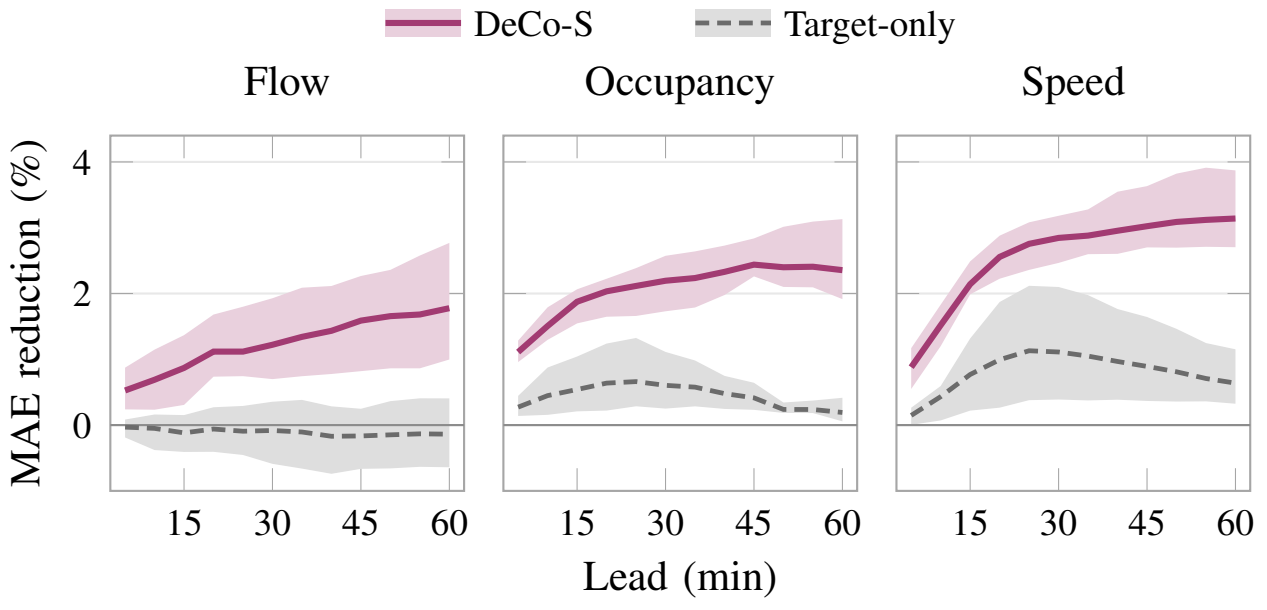

%% file: figures/sensitivity_figure.tex
\begin{figure}[t]
\centering
\begin{tikzpicture}
\begin{groupplot}[group style={group size=2 by 1, horizontal sep=1.0cm}, scale only axis, width=3.25cm, height=2.2cm,
  xlabel style={yshift=2pt}, ylabel style={yshift=-2pt}, title style={font=\footnotesize, yshift=-0.9ex, text depth=0.5ex}, ymajorgrids, grid style={black!9}, axis line style={black!35}, tick style={black!35}, tick label style={font=\scriptsize}, label style={font=\footnotesize}, clip=false]
\nextgroupplot[title={(a) Relative to $K{=}12$}, xmode=log, log basis x=2, xmin=0.84, xmax=14.3, xtick={1,2,3,4,6,12}, xticklabels={1,2,3,4,6,12}, minor tick style={draw=none},
  ymin=-0.23, ymax=3.3, ytick={0,1,2,3}, yticklabels={0,1,2,3}, xlabel={Fold forecasters, $K$}, ylabel={MAE change (\%)}]
\fill[cofacolor!12] (axis cs:10.4,-0.23) rectangle (axis cs:13.9,3.3);
\addplot[black!45, line width=0.4pt, forget plot] coordinates {(0.84,0) (14.3,0)};
\addplot[ds03, line width=0.9pt, mark=*, mark size=1.15pt, mark options={solid, fill=ds03}, forget plot] coordinates {(1,2.982) (2,1.626) (3,0.791) (4,0.681) (6,0.418) (12,0.000)};
\addplot[ds04, line width=0.9pt, mark=square*, mark size=1.05pt, mark options={solid, fill=ds04}, forget plot] coordinates {(1,2.354) (2,1.387) (3,1.144) (4,0.682) (6,0.416) (12,0.000)};
\addplot[ds07, line width=0.9pt, mark=triangle*, mark size=1.45pt, mark options={solid, fill=ds07}, forget plot] coordinates {(1,2.762) (2,1.836) (3,1.383) (4,1.135) (6,0.738) (12,0.000)};
\addplot[ds08, line width=0.9pt, mark=diamond*, mark size=1.45pt, mark options={solid, fill=ds08}, forget plot] coordinates {(1,1.989) (2,1.228) (3,0.846) (4,0.632) (6,0.500) (12,0.000)};
\nextgroupplot[title={(b) Relative to $w{=}0.2$}, xmin=-0.025, xmax=0.525, xtick={0,0.1,0.2,0.3,0.4,0.5}, xticklabels={0,0.1,0.2,0.3,0.4,0.5},
  ymin=-0.15, ymax=0.5, ytick={-0.1,0,0.1,0.2,0.3,0.4}, yticklabels={$-$0.1,0,0.1,0.2,0.3,0.4}, xlabel={Early-fusion weight $w$}]
\fill[cofacolor!12] (axis cs:0.18,-0.15) rectangle (axis cs:0.22,0.5);
\addplot[black!45, line width=0.4pt, forget plot] coordinates {(-0.025,0) (0.525,0)};
\addplot[ds03, line width=0.9pt, mark=*, mark size=1.15pt, mark options={solid, fill=ds03}, forget plot] coordinates {(0,0.414) (0.05,0.280) (0.1,0.166) (0.15,0.073) (0.2,0.000) (0.25,-0.052) (0.3,-0.084) (0.35,-0.095) (0.4,-0.086) (0.45,-0.056) (0.5,-0.007)};
\addplot[ds04, line width=0.9pt, mark=square*, mark size=1.05pt, mark options={solid, fill=ds04}, forget plot] coordinates {(0,0.289) (0.05,0.187) (0.1,0.104) (0.15,0.042) (0.2,0.000) (0.25,-0.022) (0.3,-0.025) (0.35,-0.008) (0.4,0.028) (0.45,0.084) (0.5,0.158)};
\addplot[ds07, line width=0.9pt, mark=triangle*, mark size=1.45pt, mark options={solid, fill=ds07}, forget plot] coordinates {(0,0.287) (0.05,0.177) (0.1,0.092) (0.15,0.034) (0.2,0.000) (0.25,-0.009) (0.3,0.006) (0.35,0.044) (0.4,0.106) (0.45,0.191) (0.5,0.298)};
\addplot[ds08, line width=0.9pt, mark=diamond*, mark size=1.45pt, mark options={solid, fill=ds08}, forget plot] coordinates {(0,0.376) (0.05,0.247) (0.1,0.142) (0.15,0.060) (0.2,0.000) (0.25,-0.037) (0.3,-0.051) (0.35,-0.043) (0.4,-0.013) (0.45,0.039) (0.5,0.114)};
\end{groupplot}
\begin{scope}[shift={($(group c1r1.north west)+(-0.55cm,0.745cm)$)}, font=\scriptsize]
\draw[ds03, line width=0.9pt] (0.00,0) -- (0.42,0);\node[ds03, mark size=1.15pt, inner sep=0pt] at (0.21,0) {\pgfuseplotmark{*}};
\node[anchor=base west, inner sep=0pt] at (0.50,-0.085) {PeMS03};
\draw[ds04, line width=0.9pt] (1.62,0) -- (2.04,0);\node[ds04, mark size=1.05pt, inner sep=0pt] at (1.83,0) {\pgfuseplotmark{square*}};
\node[anchor=base west, inner sep=0pt] at (2.12,-0.085) {PeMS04};
\draw[ds07, line width=0.9pt] (3.24,0) -- (3.66,0);\node[ds07, mark size=1.45pt, inner sep=0pt] at (3.45,0) {\pgfuseplotmark{triangle*}};
\node[anchor=base west, inner sep=0pt] at (3.74,-0.085) {PeMS07};
\draw[ds08, line width=0.9pt] (4.86,0) -- (5.28,0);\node[ds08, mark size=1.45pt, inner sep=0pt] at (5.07,0) {\pgfuseplotmark{diamond*}};
\node[anchor=base west, inner sep=0pt] at (5.36,-0.085) {PeMS08};
\fill[cofacolor!12] (6.48,-0.09) rectangle (6.76,0.09);
\node[anchor=base west, inner sep=0pt] at (6.84,-0.085) {used in DeCo};
\end{scope}
\end{tikzpicture}
\caption{Sensitivity of the MAE, as the change (\%) against the setting used in DeCo, marked by the shaded band. (a)~Number of fold forecasters $K$ kept per variable, against $K=12$. Each point is the mean of two evenly spaced choices of the $K$ forecasters. (b)~Early-fusion weight, against $w=0.2$. Each line is one dataset, averaged over the three variables.}
\label{fig:sensitivity}
\end{figure}
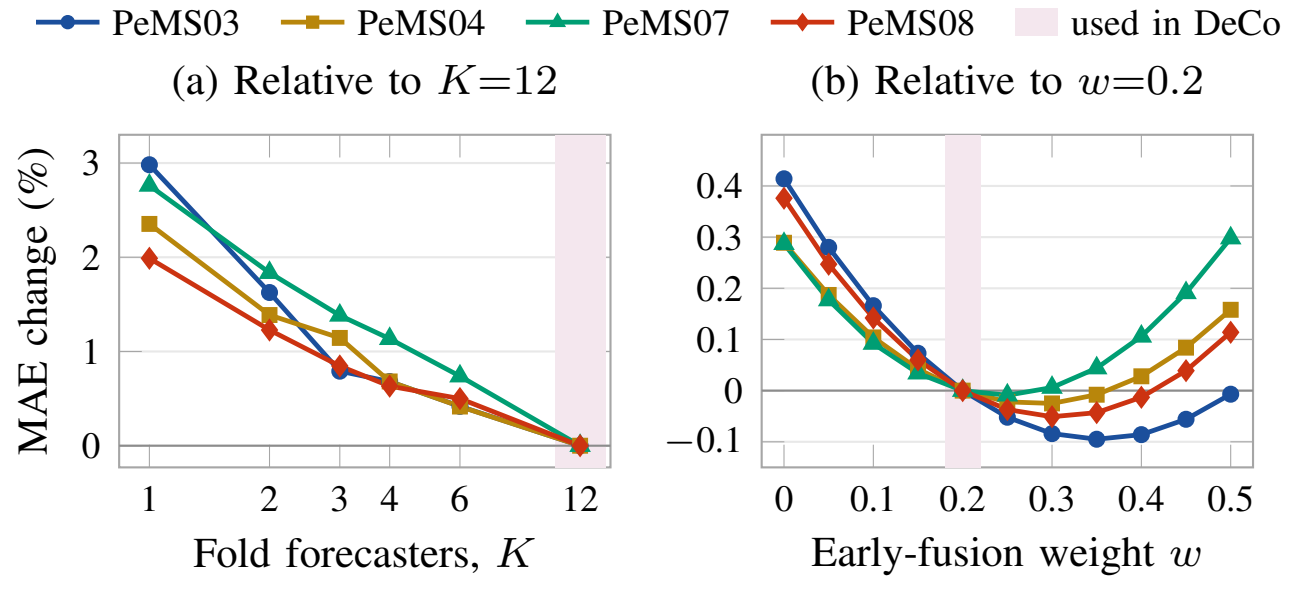

%% file: figures/states_figure.tex
\begin{figure}[t]
\centering
\def\dcsw#1{{\setlength{\fboxsep}{0pt}\setlength{\fboxrule}{0.2pt}\raisebox{0.012cm}{\fcolorbox{black!55}{#1}{\rule{0.30cm}{0pt}\rule{0pt}{0.14cm}}}}}
\begin{tikzpicture}
\begin{axis}[scale only axis, width=7.4cm, height=2.5cm, xmin=-0.5, xmax=3.5, ymin=0, ymax=6.2,
  xtick={0,1,2,3}, xticklabels={Free flow,Onset,In congestion,Recovery}, x tick style={draw=none}, ytick={0,2,4,6}, yticklabels={0,2,4,6},
  ylabel={Speed MAE reduction\\over target-only (\%)}, ylabel style={yshift=-2pt, align=center}, ymajorgrids, grid style={black!9}, axis line style={black!35}, tick style={black!35}, tick label style={font=\scriptsize}, label style={font=\footnotesize}, clip=false]
\addplot[ybar, bar width=8.4pt, bar shift=0pt, fill=cofacolor!88, draw=black!55, line width=0.2pt, error bars/.cd, y dir=both, y explicit, error bar style={black!85, line width=0.4pt}, error mark options={rotate=90, mark size=1.5pt, line width=0.4pt, black!85}] coordinates {(-0.1900,1.546) += (0,0.944) -= (0,0.955) (0.8100,3.032) += (0,1.469) -= (0,1.309) (1.8100,2.916) += (0,1.228) -= (0,1.412) (2.8100,2.519) += (0,1.300) -= (0,1.496)};
\addplot[ybar, bar width=8.4pt, bar shift=0pt, fill=cofacolor!55, draw=black!55, line width=0.2pt, error bars/.cd, y dir=both, y explicit, error bar style={black!85, line width=0.4pt}, error mark options={rotate=90, mark size=1.5pt, line width=0.4pt, black!85}] coordinates {(0.0000,1.428) += (0,1.351) -= (0,1.261) (1.0000,3.162) += (0,1.312) -= (0,1.565) (2.0000,1.366) += (0,0.409) -= (0,0.403) (3.0000,4.582) += (0,0.789) -= (0,0.979)};
\addplot[ybar, bar width=8.4pt, bar shift=0pt, fill=cofacolor!25, draw=black!55, line width=0.2pt, error bars/.cd, y dir=both, y explicit, error bar style={black!85, line width=0.4pt}, error mark options={rotate=90, mark size=1.5pt, line width=0.4pt, black!85}] coordinates {(0.1900,1.291) += (0,1.361) -= (0,1.185) (1.1900,3.566) += (0,1.162) -= (0,1.525) (2.1900,1.602) += (0,0.306) -= (0,0.592) (3.1900,4.520) += (0,0.441) -= (0,0.400)};
\end{axis}
\node[anchor=base, inner sep=0, font=\scriptsize] at ($(current axis.north)+(0cm,0.22cm)$) {\dcsw{cofacolor!88}\hspace{0.07cm}5th / 95th\hspace{0.45cm}\dcsw{cofacolor!55}\hspace{0.07cm}20th / 80th\hspace{0.45cm}\dcsw{cofacolor!25}\hspace{0.07cm}35th / 65th percentile};
\end{tikzpicture}
\caption{Speed MAE reduction (\%) of DeCo-S over the target-only module in each traffic state. The three bars of a state use the three congestion definitions, from strict to loose, named by their speed and occupancy percentiles. Bars are means over the four datasets, and whiskers span the lowest and highest dataset.}
\label{fig:states}
\end{figure}
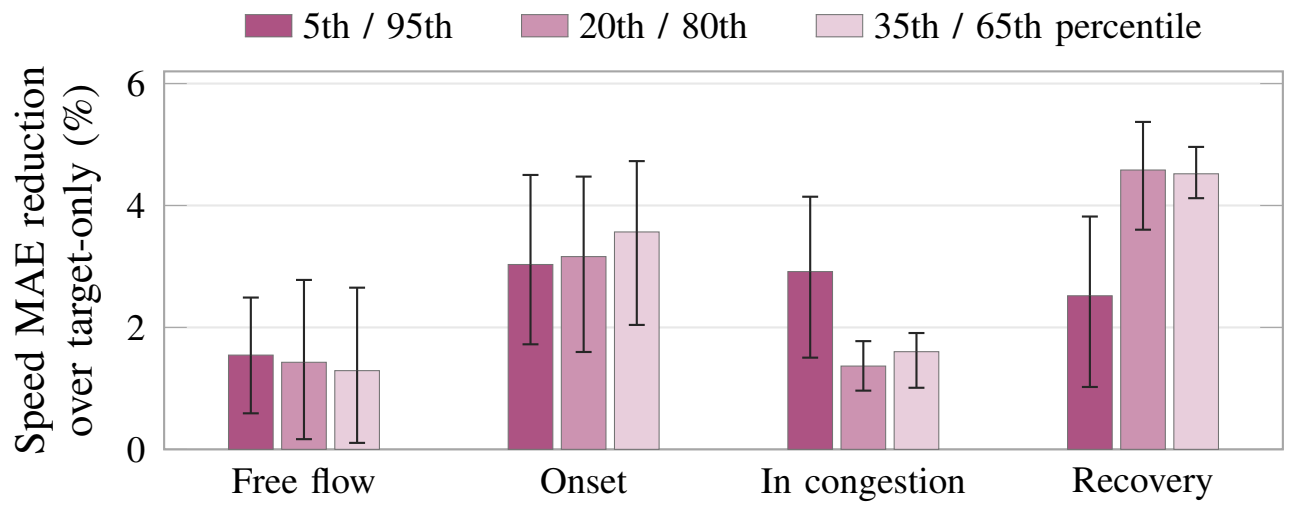

%% file: tables/ragc_table.tex
\begin{table}[t]
\centering
\caption{DeCo with RAGC forecasters (test MAE, three-seed mean).}
\label{tab:ragc}
\scriptsize
\setlength{\tabcolsep}{2.4pt}
\setlength{\aboverulesep}{0.3ex}
\setlength{\belowrulesep}{0.3ex}
\begin{tabular}{llccccc@{\hspace{7pt}}c}
\toprule
 & & \multicolumn{2}{c}{Single forecaster} & \multicolumn{3}{c}{Fold average} & \\
\cmidrule(lr){3-4}\cmidrule(lr){5-7}
Data & Var. & RAGC & + Coord. & RAGC & + Target-only & + Coord.$^{\ast}$ & DeCo\\
\midrule
\multirow{3}{*}{PeMS03} & Flow & 14.56 & 14.42 & 14.39 & 14.40 & 14.24 & \textbf{14.06}\\
 & Occ. & 0.888 & 0.884 & 0.878 & 0.880 & 0.877 & \textbf{0.875}\\
 & Speed & 1.798 & 1.814 & \textbf{1.779} & 1.793 & 1.797 & 1.791\\
\addlinespace[2pt]
\multirow{3}{*}{PeMS04} & Flow & 19.00 & 18.82 & 18.69 & 18.56 & 18.55 & \textbf{18.34}\\
 & Occ. & 0.737 & 0.724 & 0.727 & 0.722 & 0.716 & \textbf{0.709}\\
 & Speed & 1.529 & 1.524 & 1.515 & 1.515 & 1.508 & \textbf{1.491}\\
\addlinespace[2pt]
\multirow{3}{*}{PeMS07} & Flow & 20.35 & 20.01 & 20.00 & 19.84 & 19.68 & \textbf{19.36}\\
 & Occ. & 1.114 & 1.093 & 1.099 & 1.094 & 1.083 & \textbf{1.077}\\
 & Speed & 2.100 & 2.078 & 2.072 & 2.066 & 2.054 & \textbf{2.021}\\
\addlinespace[2pt]
\multirow{3}{*}{PeMS08} & Flow & 14.57 & 14.52 & 14.33 & 14.32 & 14.31 & \textbf{13.86}\\
 & Occ. & 0.638 & 0.632 & 0.628 & 0.629 & 0.625 & \textbf{0.614}\\
 & Speed & 1.238 & 1.242 & 1.224 & 1.229 & 1.228 & \textbf{1.218}\\
\bottomrule
\end{tabular}
\par\smallskip
\parbox{\columnwidth}{\scriptsize $^{\ast}$DeCo (RAGC) in Table~\ref{tab:main}. Bold marks the lowest value in each row.}
\end{table}

%% file: figures/sensors_figure.tex
\begin{figure}[t]
\centering
\begin{tikzpicture}
\begin{axis}[scale only axis, width=7.3cm, height=2.75cm, xmin=-0.6, xmax=3.6, ymin=-5.2, ymax=10.2, xtick={0,1,2,3},
  xticklabels={PeMS03,PeMS04,PeMS07,PeMS08}, x tick style={draw=none}, ytick={-4,0,4,8}, yticklabels={$-$4,0,4,8}, ylabel={Speed MAE reduction (\%)},
  ylabel style={yshift=-2pt}, ymajorgrids, grid style={black!9}, axis line style={black!35}, tick style={black!35}, tick label style={font=\scriptsize}, label style={font=\footnotesize}, clip=false]
\draw[black!50, line width=0.45pt] (axis cs:-0.6,0) -- (axis cs:3.6,0);
\draw[black!55, line width=0.55pt] (axis cs:-0.170,-0.881) -- (axis cs:-0.170,-0.050) (axis cs:-0.170,1.097) -- (axis cs:-0.170,3.306);
\draw[black!55, line width=0.55pt] (axis cs:-0.240,-0.881) -- (axis cs:-0.100,-0.881) (axis cs:-0.240,3.306) -- (axis cs:-0.100,3.306);
\draw[draw=black!55, fill=white, line width=0.55pt] (axis cs:-0.300,-0.050) rectangle (axis cs:-0.040,1.097);
\draw[black!55, line width=0.75pt] (axis cs:-0.300,0.493) -- (axis cs:-0.040,0.493);
\node[anchor=south, inner sep=1.2pt, font=\fontsize{6}{7}\selectfont, text=black!60] at (axis cs:-0.17,3.306) {73\%};
\draw[ds03, line width=0.65pt] (axis cs:0.170,-0.226) -- (axis cs:0.170,1.611) (axis cs:0.170,4.355) -- (axis cs:0.170,7.415);
\draw[ds03, line width=0.65pt] (axis cs:0.100,-0.226) -- (axis cs:0.240,-0.226) (axis cs:0.100,7.415) -- (axis cs:0.240,7.415);
\draw[draw=ds03, fill=ds03!28, line width=0.65pt] (axis cs:0.040,1.611) rectangle (axis cs:0.300,4.355);
\draw[ds03, line width=0.85pt] (axis cs:0.040,2.781) -- (axis cs:0.300,2.781);
\node[anchor=south, inner sep=1.2pt, font=\fontsize{6}{7}\selectfont, text=ds03] at (axis cs:0.17,7.415) {93\%};
\draw[black!55, line width=0.55pt] (axis cs:0.830,-1.950) -- (axis cs:0.830,-0.599) (axis cs:0.830,1.057) -- (axis cs:0.830,3.991);
\draw[black!55, line width=0.55pt] (axis cs:0.760,-1.950) -- (axis cs:0.900,-1.950) (axis cs:0.760,3.991) -- (axis cs:0.900,3.991);
\draw[draw=black!55, fill=white, line width=0.55pt] (axis cs:0.700,-0.599) rectangle (axis cs:0.960,1.057);
\draw[black!55, line width=0.75pt] (axis cs:0.700,0.179) -- (axis cs:0.960,0.179);
\node[anchor=south, inner sep=1.2pt, font=\fontsize{6}{7}\selectfont, text=black!60] at (axis cs:0.83,3.991) {55\%};
\draw[ds04, line width=0.65pt] (axis cs:1.170,-1.206) -- (axis cs:1.170,0.370) (axis cs:1.170,3.514) -- (axis cs:1.170,6.035);
\draw[ds04, line width=0.65pt] (axis cs:1.100,-1.206) -- (axis cs:1.240,-1.206) (axis cs:1.100,6.035) -- (axis cs:1.240,6.035);
\draw[draw=ds04, fill=ds04!28, line width=0.65pt] (axis cs:1.040,0.370) rectangle (axis cs:1.300,3.514);
\draw[ds04, line width=0.85pt] (axis cs:1.040,1.644) -- (axis cs:1.300,1.644);
\node[anchor=south, inner sep=1.2pt, font=\fontsize{6}{7}\selectfont, text=ds04] at (axis cs:1.17,6.035) {83\%};
\draw[black!55, line width=0.55pt] (axis cs:1.830,-1.200) -- (axis cs:1.830,-0.415) (axis cs:1.830,0.430) -- (axis cs:1.830,1.862);
\draw[black!55, line width=0.55pt] (axis cs:1.760,-1.200) -- (axis cs:1.900,-1.200) (axis cs:1.760,1.862) -- (axis cs:1.900,1.862);
\draw[draw=black!55, fill=white, line width=0.55pt] (axis cs:1.700,-0.415) rectangle (axis cs:1.960,0.430);
\draw[black!55, line width=0.75pt] (axis cs:1.700,-0.001) -- (axis cs:1.960,-0.001);
\node[anchor=south, inner sep=1.2pt, font=\fontsize{6}{7}\selectfont, text=black!60] at (axis cs:1.83,1.862) {50\%};
\draw[ds07, line width=0.65pt] (axis cs:2.170,-0.455) -- (axis cs:2.170,0.832) (axis cs:2.170,3.055) -- (axis cs:2.170,7.274);
\draw[ds07, line width=0.65pt] (axis cs:2.100,-0.455) -- (axis cs:2.240,-0.455) (axis cs:2.100,7.274) -- (axis cs:2.240,7.274);
\draw[draw=ds07, fill=ds07!28, line width=0.65pt] (axis cs:2.040,0.832) rectangle (axis cs:2.300,3.055);
\draw[ds07, line width=0.85pt] (axis cs:2.040,1.668) -- (axis cs:2.300,1.668);
\node[anchor=south, inner sep=1.2pt, font=\fontsize{6}{7}\selectfont, text=ds07] at (axis cs:2.17,7.274) {92\%};
\draw[black!55, line width=0.55pt] (axis cs:2.830,-1.359) -- (axis cs:2.830,-0.423) (axis cs:2.830,1.127) -- (axis cs:2.830,4.068);
\draw[black!55, line width=0.55pt] (axis cs:2.760,-1.359) -- (axis cs:2.900,-1.359) (axis cs:2.760,4.068) -- (axis cs:2.900,4.068);
\draw[draw=black!55, fill=white, line width=0.55pt] (axis cs:2.700,-0.423) rectangle (axis cs:2.960,1.127);
\draw[black!55, line width=0.75pt] (axis cs:2.700,0.189) -- (axis cs:2.960,0.189);
\node[anchor=south, inner sep=1.2pt, font=\fontsize{6}{7}\selectfont, text=black!60] at (axis cs:2.83,4.068) {55\%};
\draw[ds08, line width=0.65pt] (axis cs:3.170,-2.564) -- (axis cs:3.170,-0.132) (axis cs:3.170,2.596) -- (axis cs:3.170,7.142);
\draw[ds08, line width=0.65pt] (axis cs:3.100,-2.564) -- (axis cs:3.240,-2.564) (axis cs:3.100,7.142) -- (axis cs:3.240,7.142);
\draw[draw=ds08, fill=ds08!28, line width=0.65pt] (axis cs:3.040,-0.132) rectangle (axis cs:3.300,2.596);
\draw[ds08, line width=0.85pt] (axis cs:3.040,0.961) -- (axis cs:3.300,0.961);
\node[anchor=south, inner sep=1.2pt, font=\fontsize{6}{7}\selectfont, text=ds08] at (axis cs:3.17,7.142) {74\%};
\end{axis}
\begin{scope}[shift={($(current axis.north)+(-1.75cm,0.25cm)$)}, font=\scriptsize]
\draw[black!55, fill=white, line width=0.55pt] (0,-0.08) rectangle (0.28,0.08);
\node[anchor=base west,inner sep=0] at (0.36,-0.085) {Target-only};
\draw[black!70, fill=black!22, line width=0.65pt] (2.10,-0.08) rectangle (2.38,0.08);
\node[anchor=base west,inner sep=0] at (2.46,-0.085) {DeCo-S};
\end{scope}
\end{tikzpicture}
\caption{Speed MAE reduction (\%) over the single forecasters across the sensors of each dataset, for the target-only module and DeCo-S. Boxes span the quartiles with a line at the median, whiskers reach the 5th and 95th percentiles, and the number above a box is the share of sensors with a lower MAE.}
\label{fig:sensors}
\end{figure}
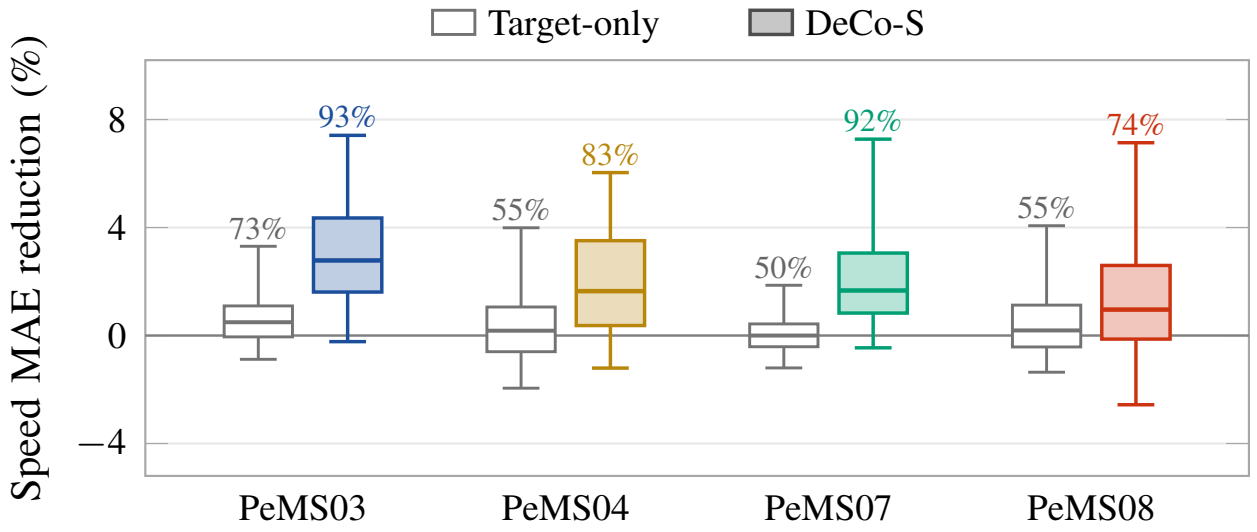

%% file: figures/downstream_figure.tex
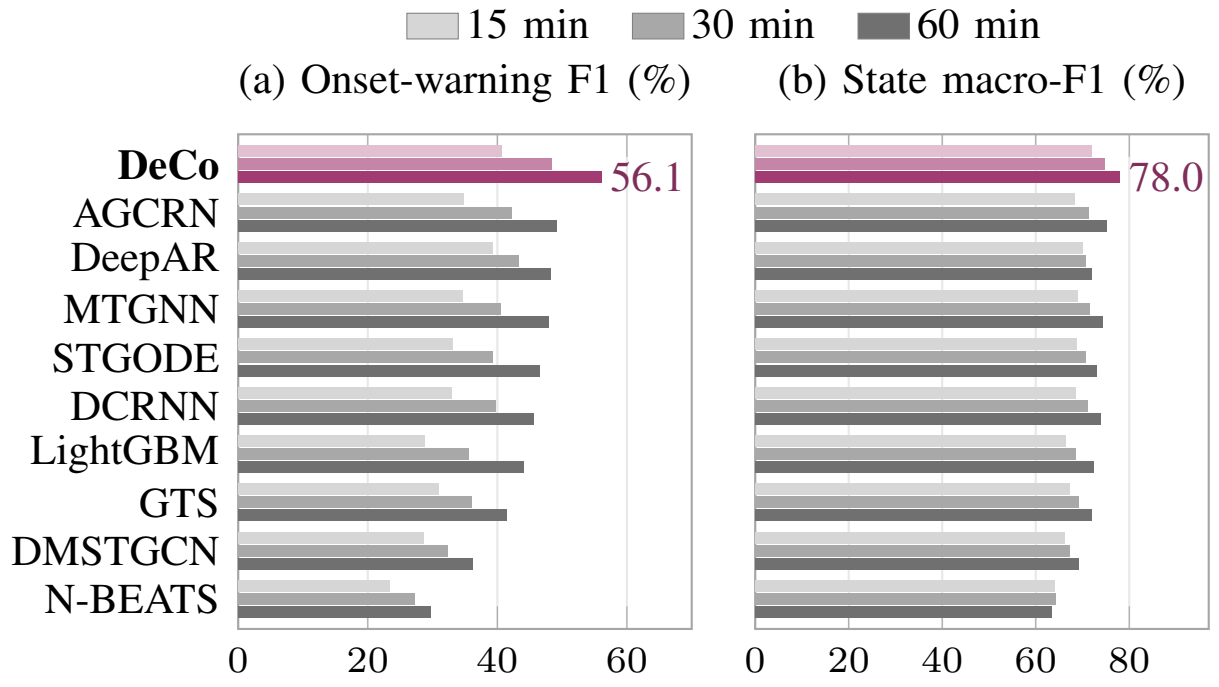
\begin{figure}[t]
\centering
\begin{tikzpicture}
\begin{groupplot}[group style={group size=2 by 1, horizontal sep=0.42cm}, scale only axis, width=3.0cm, height=3.272cm,
  title style={font=\footnotesize, yshift=-0.9ex, text depth=0.5ex}, xmajorgrids, grid style={black!9}, axis line style={black!35}, tick style={black!35},
  x tick label style={font=\scriptsize}, y tick label style={font=\footnotesize}, clip=false, y dir=reverse, ymin=-0.62, ymax=9.62, ytick={0,...,9}, ytick style={draw=none}, xmin=0]
\nextgroupplot[title={(a) Onset-warning F1 (\%)}, xmax=70, xtick={0,20,40,60}, yticklabels={\textbf{DeCo},AGCRN,DeepAR,MTGNN,STGODE,DCRNN,LightGBM,GTS,DMSTGCN,N-BEATS}]
\fill[cofacolor!32] (axis cs:0,-0.385) rectangle (axis cs:40.720,-0.155);
\fill[cofacolor!62] (axis cs:0,-0.115) rectangle (axis cs:48.381,0.115);
\fill[cofacolor] (axis cs:0,0.155) rectangle (axis cs:56.056,0.385);
\fill[black!16] (axis cs:0,0.615) rectangle (axis cs:34.767,0.845);
\fill[black!34] (axis cs:0,0.885) rectangle (axis cs:42.204,1.115);
\fill[black!56] (axis cs:0,1.155) rectangle (axis cs:49.235,1.385);
\fill[black!16] (axis cs:0,1.615) rectangle (axis cs:39.219,1.845);
\fill[black!34] (axis cs:0,1.885) rectangle (axis cs:43.250,2.115);
\fill[black!56] (axis cs:0,2.155) rectangle (axis cs:48.274,2.385);
\fill[black!16] (axis cs:0,2.615) rectangle (axis cs:34.683,2.845);
\fill[black!34] (axis cs:0,2.885) rectangle (axis cs:40.532,3.115);
\fill[black!56] (axis cs:0,3.155) rectangle (axis cs:47.881,3.385);
\fill[black!16] (axis cs:0,3.615) rectangle (axis cs:33.076,3.845);
\fill[black!34] (axis cs:0,3.885) rectangle (axis cs:39.301,4.115);
\fill[black!56] (axis cs:0,4.155) rectangle (axis cs:46.557,4.385);
\fill[black!16] (axis cs:0,4.615) rectangle (axis cs:32.927,4.845);
\fill[black!34] (axis cs:0,4.885) rectangle (axis cs:39.771,5.115);
\fill[black!56] (axis cs:0,5.155) rectangle (axis cs:45.637,5.385);
\fill[black!16] (axis cs:0,5.615) rectangle (axis cs:28.832,5.845);
\fill[black!34] (axis cs:0,5.885) rectangle (axis cs:35.632,6.115);
\fill[black!56] (axis cs:0,6.155) rectangle (axis cs:44.105,6.385);
\fill[black!16] (axis cs:0,6.615) rectangle (axis cs:30.930,6.845);
\fill[black!34] (axis cs:0,6.885) rectangle (axis cs:36.016,7.115);
\fill[black!56] (axis cs:0,7.155) rectangle (axis cs:41.514,7.385);
\fill[black!16] (axis cs:0,7.615) rectangle (axis cs:28.653,7.845);
\fill[black!34] (axis cs:0,7.885) rectangle (axis cs:32.321,8.115);
\fill[black!56] (axis cs:0,8.155) rectangle (axis cs:36.236,8.385);
\fill[black!16] (axis cs:0,8.615) rectangle (axis cs:23.441,8.845);
\fill[black!34] (axis cs:0,8.885) rectangle (axis cs:27.200,9.115);
\fill[black!56] (axis cs:0,9.155) rectangle (axis cs:29.636,9.385);
\node[font=\footnotesize, text=cofacolor!78!black, anchor=west, inner sep=1.5pt] at (axis cs:56.056,0.27) {56.1};
\nextgroupplot[title={(b) State macro-F1 (\%)}, xmax=97, xtick={0,20,40,60,80}, yticklabels={,,,,,,,,,}]
\fill[cofacolor!32] (axis cs:0,-0.385) rectangle (axis cs:71.927,-0.155);
\fill[cofacolor!62] (axis cs:0,-0.115) rectangle (axis cs:74.735,0.115);
\fill[cofacolor] (axis cs:0,0.155) rectangle (axis cs:78.016,0.385);
\fill[black!16] (axis cs:0,0.615) rectangle (axis cs:68.343,0.845);
\fill[black!34] (axis cs:0,0.885) rectangle (axis cs:71.338,1.115);
\fill[black!56] (axis cs:0,1.155) rectangle (axis cs:75.139,1.385);
\fill[black!16] (axis cs:0,1.615) rectangle (axis cs:70.132,1.845);
\fill[black!34] (axis cs:0,1.885) rectangle (axis cs:70.735,2.115);
\fill[black!56] (axis cs:0,2.155) rectangle (axis cs:72.029,2.385);
\fill[black!16] (axis cs:0,2.615) rectangle (axis cs:68.985,2.845);
\fill[black!34] (axis cs:0,2.885) rectangle (axis cs:71.484,3.115);
\fill[black!56] (axis cs:0,3.155) rectangle (axis cs:74.404,3.385);
\fill[black!16] (axis cs:0,3.615) rectangle (axis cs:68.717,3.845);
\fill[black!34] (axis cs:0,3.885) rectangle (axis cs:70.713,4.115);
\fill[black!56] (axis cs:0,4.155) rectangle (axis cs:73.014,4.385);
\fill[black!16] (axis cs:0,4.615) rectangle (axis cs:68.473,4.845);
\fill[black!34] (axis cs:0,4.885) rectangle (axis cs:71.203,5.115);
\fill[black!56] (axis cs:0,5.155) rectangle (axis cs:73.857,5.385);
\fill[black!16] (axis cs:0,5.615) rectangle (axis cs:66.478,5.845);
\fill[black!34] (axis cs:0,5.885) rectangle (axis cs:68.637,6.115);
\fill[black!56] (axis cs:0,6.155) rectangle (axis cs:72.447,6.385);
\fill[black!16] (axis cs:0,6.615) rectangle (axis cs:67.298,6.845);
\fill[black!34] (axis cs:0,6.885) rectangle (axis cs:69.271,7.115);
\fill[black!56] (axis cs:0,7.155) rectangle (axis cs:72.015,7.385);
\fill[black!16] (axis cs:0,7.615) rectangle (axis cs:66.269,7.845);
\fill[black!34] (axis cs:0,7.885) rectangle (axis cs:67.216,8.115);
\fill[black!56] (axis cs:0,8.155) rectangle (axis cs:69.266,8.385);
\fill[black!16] (axis cs:0,8.615) rectangle (axis cs:64.104,8.845);
\fill[black!34] (axis cs:0,8.885) rectangle (axis cs:64.263,9.115);
\fill[black!56] (axis cs:0,9.155) rectangle (axis cs:63.386,9.385);
\node[font=\footnotesize, text=cofacolor!78!black, anchor=west, inner sep=1.5pt] at (axis cs:78.016,0.27) {78.0};
\end{groupplot}
\node[anchor=south, inner sep=0pt, font=\footnotesize] at ($(group c1r1.north west)!0.5!(group c2r1.north east)+(0,0.62cm)$) {\tikz[baseline=-0.6ex]{\draw[black!50, line width=0.2pt, fill=black!16] (0,-0.085) rectangle (0.34,0.085);}\,15 min\hspace{1.0em}\tikz[baseline=-0.6ex]{\draw[black!50, line width=0.2pt, fill=black!34] (0,-0.085) rectangle (0.34,0.085);}\,30 min\hspace{1.0em}\tikz[baseline=-0.6ex]{\draw[black!50, line width=0.2pt, fill=black!56] (0,-0.085) rectangle (0.34,0.085);}\,60 min};
\end{tikzpicture}
\caption{Onset-warning F1 and state macro-F1 (\%) at leads of 15, 30, and 60 minutes on PeMS04 and PeMS08 (mean of the two datasets, three seeds). Each baseline is shown in its best trained form at each lead. The numbers are the scores of DeCo at 60 minutes.}
\label{fig:downstream}
\end{figure}

%% file: figures/cost_figure.tex
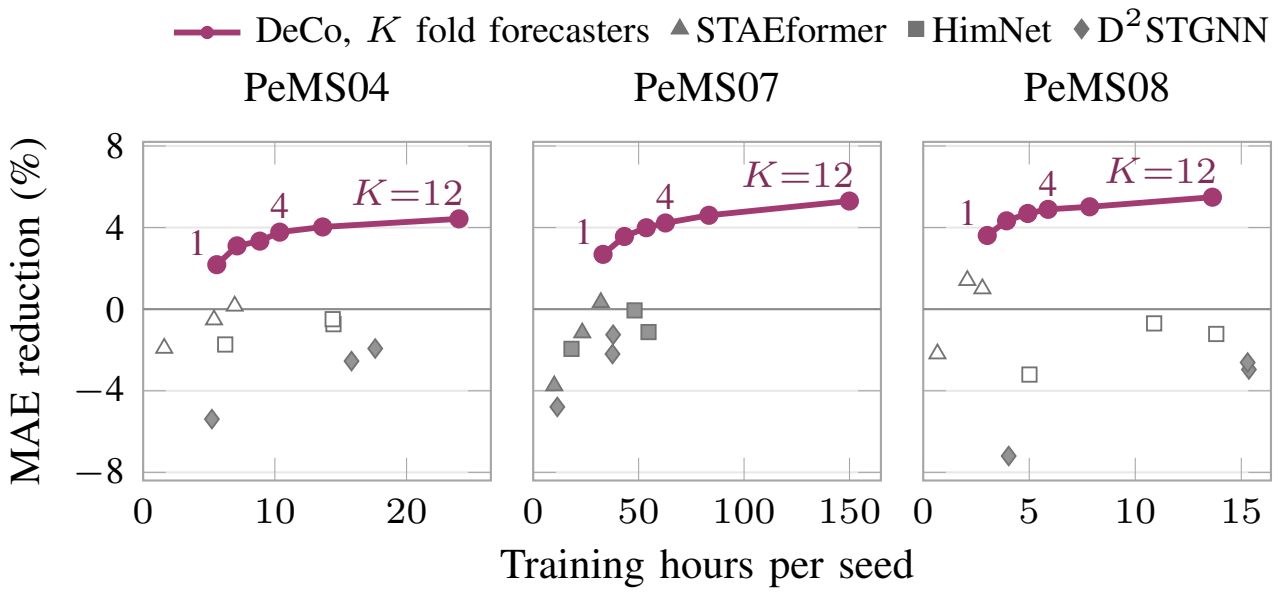
\begin{figure}[t]
\centering
\begin{tikzpicture}
\begin{groupplot}[group style={group size=3 by 1, horizontal sep=0.28cm}, scale only axis, width=2.3cm, height=2.24cm,
  xlabel style={yshift=2pt}, ylabel style={yshift=-2pt}, title style={font=\footnotesize, yshift=-0.9ex, text depth=0.5ex}, ymajorgrids, grid style={black!9}, axis line style={black!35}, tick style={black!35}, tick label style={font=\scriptsize}, label style={font=\footnotesize}, clip=false, xmin=0, ymin=-8.4, ymax=8.2, ytick={-8,-4,0,4,8}, yticklabels={$-$8,$-$4,0,4,8}]
\nextgroupplot[title={PeMS04}, xmax=26.4, xtick={0,10,20}, ylabel={MAE reduction (\%)}]
\addplot[black!45, line width=0.4pt, forget plot] coordinates {(0,0) (26.4,0)};
\addplot[only marks, mark=triangle*, mark size=1.75pt, mark options={draw=black!55, line width=0.35pt, fill=white}, forget plot] coordinates {(5.38,-0.521)};
\addplot[only marks, mark=triangle*, mark size=1.75pt, mark options={draw=black!55, line width=0.35pt, fill=white}, forget plot] coordinates {(6.95,0.155)};
\addplot[only marks, mark=triangle*, mark size=1.75pt, mark options={draw=black!55, line width=0.35pt, fill=white}, forget plot] coordinates {(1.60,-1.912)};
\addplot[only marks, mark=square*, mark size=1.35pt, mark options={draw=black!55, line width=0.35pt, fill=white}, forget plot] coordinates {(14.45,-0.736)};
\addplot[only marks, mark=square*, mark size=1.35pt, mark options={draw=black!55, line width=0.35pt, fill=white}, forget plot] coordinates {(14.39,-0.492)};
\addplot[only marks, mark=square*, mark size=1.35pt, mark options={draw=black!55, line width=0.35pt, fill=white}, forget plot] coordinates {(6.23,-1.732)};
\addplot[only marks, mark=diamond*, mark size=1.75pt, mark options={draw=black!55, line width=0.35pt, fill=black!42}, forget plot] coordinates {(15.82,-2.546)};
\addplot[only marks, mark=diamond*, mark size=1.75pt, mark options={draw=black!55, line width=0.35pt, fill=black!42}, forget plot] coordinates {(17.62,-1.929)};
\addplot[only marks, mark=diamond*, mark size=1.75pt, mark options={draw=black!55, line width=0.35pt, fill=black!42}, forget plot] coordinates {(5.22,-5.390)};
\addplot[cofacolor, line width=1.1pt, mark=*, mark size=1.25pt, mark options={solid, fill=cofacolor}, forget plot] coordinates {(5.59,2.180) (7.13,3.102) (8.86,3.335) (10.37,3.775) (13.63,4.029) (23.98,4.426)};
\node[font=\scriptsize, text=cofacolor!78!black, anchor=south east, inner sep=1.7pt] at (axis cs:5.59,2.180) {1};
\node[font=\scriptsize, text=cofacolor!78!black, anchor=south, inner sep=2.4pt] at (axis cs:10.37,3.775) {4};
\node[font=\scriptsize, text=cofacolor!78!black, anchor=south east, inner sep=2.4pt, xshift=3.5pt] at (axis cs:23.98,4.426) {$K{=}12$};
\nextgroupplot[title={PeMS07}, xmax=164.9, xtick={0,50,100,150}, xlabel={Training hours per seed}, yticklabels={,,,,}]
\addplot[black!45, line width=0.4pt, forget plot] coordinates {(0,0) (164.9,0)};
\addplot[only marks, mark=triangle*, mark size=1.75pt, mark options={draw=black!55, line width=0.35pt, fill=black!42}, forget plot] coordinates {(23.26,-1.158)};
\addplot[only marks, mark=triangle*, mark size=1.75pt, mark options={draw=black!55, line width=0.35pt, fill=black!42}, forget plot] coordinates {(32.07,0.330)};
\addplot[only marks, mark=triangle*, mark size=1.75pt, mark options={draw=black!55, line width=0.35pt, fill=black!42}, forget plot] coordinates {(9.98,-3.753)};
\addplot[only marks, mark=square*, mark size=1.35pt, mark options={draw=black!55, line width=0.35pt, fill=black!42}, forget plot] coordinates {(54.69,-1.123)};
\addplot[only marks, mark=square*, mark size=1.35pt, mark options={draw=black!55, line width=0.35pt, fill=black!42}, forget plot] coordinates {(48.07,-0.061)};
\addplot[only marks, mark=square*, mark size=1.35pt, mark options={draw=black!55, line width=0.35pt, fill=black!42}, forget plot] coordinates {(18.16,-1.946)};
\addplot[only marks, mark=diamond*, mark size=1.75pt, mark options={draw=black!55, line width=0.35pt, fill=black!42}, forget plot] coordinates {(37.61,-2.199)};
\addplot[only marks, mark=diamond*, mark size=1.75pt, mark options={draw=black!55, line width=0.35pt, fill=black!42}, forget plot] coordinates {(37.91,-1.254)};
\addplot[only marks, mark=diamond*, mark size=1.75pt, mark options={draw=black!55, line width=0.35pt, fill=black!42}, forget plot] coordinates {(11.44,-4.793)};
\addplot[cofacolor, line width=1.1pt, mark=*, mark size=1.25pt, mark options={solid, fill=cofacolor}, forget plot] coordinates {(33.13,2.686) (43.26,3.562) (53.63,3.991) (62.72,4.225) (83.34,4.601) (149.95,5.299)};
\node[font=\scriptsize, text=cofacolor!78!black, anchor=south east, inner sep=1.7pt] at (axis cs:33.13,2.686) {1};
\node[font=\scriptsize, text=cofacolor!78!black, anchor=south, inner sep=2.4pt] at (axis cs:62.72,4.225) {4};
\node[font=\scriptsize, text=cofacolor!78!black, anchor=south east, inner sep=2.4pt, xshift=3.5pt] at (axis cs:149.95,5.299) {$K{=}12$};
\nextgroupplot[title={PeMS08}, xmax=16.4, xtick={0,5,10,15}, yticklabels={,,,,}]
\addplot[black!45, line width=0.4pt, forget plot] coordinates {(0,0) (16.4,0)};
\addplot[only marks, mark=triangle*, mark size=1.75pt, mark options={draw=black!55, line width=0.35pt, fill=white}, forget plot] coordinates {(2.79,1.002)};
\addplot[only marks, mark=triangle*, mark size=1.75pt, mark options={draw=black!55, line width=0.35pt, fill=white}, forget plot] coordinates {(2.08,1.410)};
\addplot[only marks, mark=triangle*, mark size=1.75pt, mark options={draw=black!55, line width=0.35pt, fill=white}, forget plot] coordinates {(0.67,-2.187)};
\addplot[only marks, mark=square*, mark size=1.35pt, mark options={draw=black!55, line width=0.35pt, fill=white}, forget plot] coordinates {(13.82,-1.216)};
\addplot[only marks, mark=square*, mark size=1.35pt, mark options={draw=black!55, line width=0.35pt, fill=white}, forget plot] coordinates {(10.89,-0.698)};
\addplot[only marks, mark=square*, mark size=1.35pt, mark options={draw=black!55, line width=0.35pt, fill=white}, forget plot] coordinates {(5.01,-3.206)};
\addplot[only marks, mark=diamond*, mark size=1.75pt, mark options={draw=black!55, line width=0.35pt, fill=black!42}, forget plot] coordinates {(15.36,-2.957)};
\addplot[only marks, mark=diamond*, mark size=1.75pt, mark options={draw=black!55, line width=0.35pt, fill=black!42}, forget plot] coordinates {(15.30,-2.614)};
\addplot[only marks, mark=diamond*, mark size=1.75pt, mark options={draw=black!55, line width=0.35pt, fill=black!42}, forget plot] coordinates {(4.03,-7.198)};
\addplot[cofacolor, line width=1.1pt, mark=*, mark size=1.25pt, mark options={solid, fill=cofacolor}, forget plot] coordinates {(3.02,3.611) (3.93,4.330) (4.93,4.690) (5.89,4.893) (7.85,5.017) (13.64,5.488)};
\node[font=\scriptsize, text=cofacolor!78!black, anchor=south east, inner sep=1.7pt] at (axis cs:3.02,3.611) {1};
\node[font=\scriptsize, text=cofacolor!78!black, anchor=south, inner sep=2.4pt] at (axis cs:5.89,4.893) {4};
\node[font=\scriptsize, text=cofacolor!78!black, anchor=south east, inner sep=2.4pt, xshift=3.5pt] at (axis cs:13.64,5.488) {$K{=}12$};
\end{groupplot}
\node[anchor=south, inner sep=0pt, font=\scriptsize] at ($(group c1r1.north west)!0.5!(group c3r1.north east)+(0,0.62cm)$) {\hspace{0.7em}\tikz[baseline=-0.55ex]{\draw[cofacolor, line width=1.1pt] (0,0)--(0.44,0); \fill[cofacolor] (0.22,0) circle (1.25pt);}\,\,DeCo, $K$ fold forecasters\hspace{0.7em}\tikz[baseline=-0.55ex]{\node[mark size=1.75pt, inner sep=0pt, draw=black!55, fill=black!42, line width=0.35pt, color=black!55] at (0,0) {\pgfuseplotmark{triangle*}};}\,\,STAEformer\hspace{0.7em}\tikz[baseline=-0.55ex]{\node[mark size=1.35pt, inner sep=0pt, draw=black!55, fill=black!42, line width=0.35pt, color=black!55] at (0,0) {\pgfuseplotmark{square*}};}\,\,HimNet\hspace{0.7em}\tikz[baseline=-0.55ex]{\node[mark size=1.75pt, inner sep=0pt, draw=black!55, fill=black!42, line width=0.35pt, color=black!55] at (0,0) {\pgfuseplotmark{diamond*}};}\,\,D$^2$STGNN};
\end{tikzpicture}
\caption{MAE reduction (\%) over the single forecasters (mean of the three variables) against training hours per seed. DeCo keeps $K=1$, 2, 3, 4, 6, and 12 fold forecasters from left to right, and each baseline has its SISO, MISO, and MIMO forms. Hollow markers include runs on non-H100 GPUs.}
\label{fig:cost}
\end{figure}

%% file: tables/lead_table.tex
\begin{table}[t]
\centering
\caption{Test MAE at leads of 15, 30, and 60 minutes on PeMS04 and PeMS08 (mean over three seeds). Bold marks the best value and underlining the second best.}
\label{tab:lead}
\scriptsize
\setlength{\tabcolsep}{1.7pt}
\setlength{\aboverulesep}{0.25ex}
\setlength{\belowrulesep}{0.25ex}
\renewcommand{\arraystretch}{0.96}
\begin{tabular}{llccccccccc}
\toprule
 & & \multicolumn{3}{c}{Flow} & \multicolumn{3}{c}{Occ.} & \multicolumn{3}{c}{Speed}\\
\cmidrule(lr){3-5}\cmidrule(lr){6-8}\cmidrule(lr){9-11}
Model & Variant & 15 & 30 & 60 & 15 & 30 & 60 & 15 & 30 & 60\\
\midrule
\multicolumn{11}{c}{PeMS04}\\
\midrule
\multirow{3}{*}{STAEformer} & SISO & 18.27 & 19.10 & 20.68 & 0.684 & 0.753 & 0.832 & 1.356 & 1.640 & 1.918\\
 & MISO & 18.16 & 19.01 & 20.62 & 0.683 & 0.752 & 0.833 & 1.336 & 1.615 & 1.889\\
 & MIMO & 18.51 & 19.55 & 21.42 & 0.689 & 0.761 & 0.847 & 1.362 & 1.651 & 1.917\\
\midrule
\multirow{3}{*}{HimNet} & SISO & 18.29 & 19.32 & 20.92 & 0.681 & 0.755 & 0.837 & 1.334 & 1.624 & 1.913\\
 & MISO & 18.35 & 19.43 & 21.02 & 0.677 & 0.750 & 0.832 & 1.326 & 1.617 & 1.898\\
 & MIMO & 18.56 & 19.80 & 21.74 & 0.682 & 0.759 & 0.847 & 1.333 & 1.623 & 1.898\\
\midrule
\multirow{3}{*}{D$^2$STGNN} & SISO & 18.41 & 19.48 & 21.48 & 0.695 & 0.777 & 0.875 & 1.344 & 1.646 & 1.956\\
 & MISO & 18.40 & 19.56 & 21.72 & 0.690 & 0.767 & 0.865 & 1.337 & 1.635 & 1.927\\
 & MIMO & 19.18 & 20.63 & 23.58 & 0.694 & 0.775 & 0.883 & 1.361 & 1.675 & 1.994\\
\midrule
\multirow{2}{*}{DecoupleST} & SISO & 18.11 & 19.11 & 20.68 & 0.679 & 0.748 & \underline{0.828} & 1.336 & 1.630 & 1.908\\
 & MISO & \underline{18.01} & \underline{18.95} & \underline{20.41} & 0.674 & \underline{0.745} & 0.831 & 1.327 & 1.615 & 1.887\\
\midrule
RAGC & MISO & \underline{18.01} & 18.99 & 20.70 & \underline{0.672} & 0.746 & 0.835 & \underline{1.302} & \underline{1.582} & \underline{1.859}\\
\midrule
\rowcolor{decoshade}\textbf{DeCo} & MIMO & \textbf{17.57} & \textbf{18.36} & \textbf{19.59} & \textbf{0.653} & \textbf{0.718} & \textbf{0.791} & \textbf{1.276} & \textbf{1.543} & \textbf{1.795}\\
\midrule
\multicolumn{11}{c}{PeMS08}\\
\midrule
\multirow{3}{*}{STAEformer} & SISO & \underline{13.23} & \underline{14.10} & \underline{15.76} & 0.592 & 0.662 & 0.744 & 1.094 & 1.325 & 1.560\\
 & MISO & \underline{13.23} & 14.13 & 15.91 & 0.593 & 0.660 & 0.734 & 1.087 & 1.314 & 1.541\\
 & MIMO & 14.19 & 15.09 & 16.97 & 0.597 & 0.669 & 0.755 & 1.121 & 1.350 & 1.577\\
\midrule
\multirow{3}{*}{HimNet} & SISO & 13.86 & 14.61 & 16.14 & 0.601 & 0.671 & 0.747 & 1.111 & 1.356 & 1.585\\
 & MISO & 13.83 & 14.61 & 16.13 & 0.597 & 0.664 & 0.744 & 1.106 & 1.345 & 1.571\\
 & MIMO & 14.54 & 15.49 & 17.30 & 0.604 & 0.674 & 0.759 & 1.112 & 1.346 & 1.569\\
\midrule
\multirow{3}{*}{D$^2$STGNN} & SISO & 14.17 & 15.09 & 17.20 & 0.598 & 0.671 & 0.764 & 1.109 & 1.363 & 1.622\\
 & MISO & 14.12 & 15.13 & 17.12 & 0.597 & 0.669 & 0.769 & 1.104 & 1.344 & 1.606\\
 & MIMO & 15.09 & 16.34 & 18.85 & 0.614 & 0.688 & 0.796 & 1.126 & 1.373 & 1.645\\
\midrule
\multirow{2}{*}{DecoupleST} & SISO & 13.60 & 14.49 & 15.99 & 0.588 & 0.660 & 0.738 & 1.102 & 1.351 & 1.578\\
 & MISO & 13.61 & 14.46 & 15.97 & 0.590 & 0.663 & 0.737 & 1.103 & 1.339 & 1.555\\
\midrule
RAGC & MISO & 13.72 & 14.53 & 16.17 & \underline{0.578} & \underline{0.645} & \underline{0.728} & \underline{1.060} & \underline{1.278} & \underline{1.501}\\
\midrule
\rowcolor{decoshade}\textbf{DeCo} & MIMO & \textbf{13.08} & \textbf{13.84} & \textbf{15.07} & \textbf{0.560} & \textbf{0.622} & \textbf{0.691} & \textbf{1.039} & \textbf{1.259} & \textbf{1.468}\\
\bottomrule
\end{tabular}
\par\smallskip
\parbox{\columnwidth}{\scriptsize Leads in minutes. Occupancy in \%.}
\end{table}

%% file: tables/forecasters_table.tex
\begin{table}[t]
\centering
\caption{Test MAE of the fold average of three further forecasters, without and with the coordination module (three module seeds).}
\label{tab:forecasters}
\scriptsize
\setlength{\tabcolsep}{2.0pt}
\setlength{\aboverulesep}{0.25ex}
\setlength{\belowrulesep}{0.25ex}
\begin{tabular}{llcccccc}
\toprule
 & & \multicolumn{2}{c}{Flow} & \multicolumn{2}{c}{Occ.} & \multicolumn{2}{c}{Speed}\\
\cmidrule(lr){3-4}\cmidrule(lr){5-6}\cmidrule(lr){7-8}
Dataset & Forecaster & Avg. & + $g$ & Avg. & + $g$ & Avg. & + $g$\\
\midrule
\multirow{3}{*}{PeMS04} & Graph WaveNet & 19.26 & \textbf{19.03} & 0.748 & \textbf{0.737} & 1.567 & \textbf{1.540}\\
 & STGCN & 19.73 & \textbf{19.11} & 0.753 & \textbf{0.737} & 1.559 & \textbf{1.542}\\
 & STID & 18.71 & \textbf{18.58} & 0.732 & \textbf{0.727} & 1.549 & \textbf{1.537}\\
\midrule
\multirow{3}{*}{PeMS08} & Graph WaveNet & 14.36 & \textbf{14.18} & 0.630 & \textbf{0.624} & 1.239 & \textbf{1.231}\\
 & STGCN & 14.86 & \textbf{14.47} & 0.648 & \textbf{0.636} & 1.256 & \textbf{1.242}\\
 & STID & 14.12 & \textbf{14.00} & 0.634 & \textbf{0.627} & 1.248 & \textbf{1.239}\\
\bottomrule
\end{tabular}
\par\smallskip
\parbox{\columnwidth}{\scriptsize Avg.: fold average of the forecasters. + $g$: with the coordination module. Occupancy in \%.}
\end{table}

%% file: figures/case_figure.tex
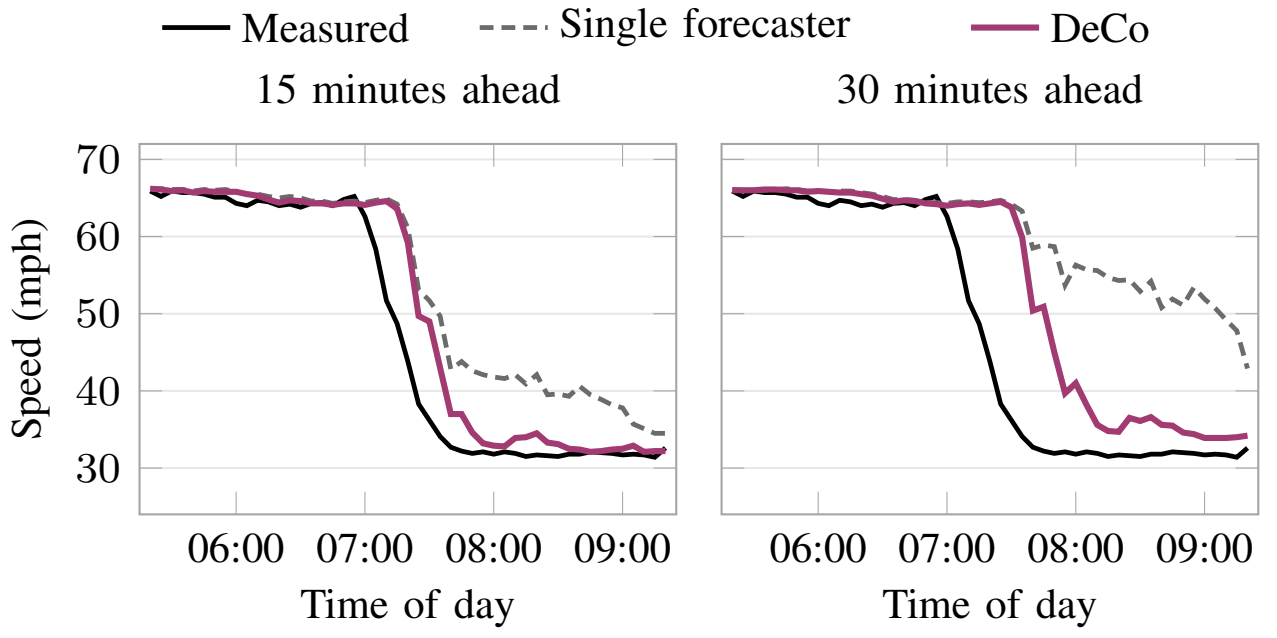
\begin{figure}[t]
\centering
\begin{tikzpicture}
\begin{groupplot}[group style={group size=2 by 1, horizontal sep=0.30cm}, scale only axis, width=3.55cm, height=2.45cm,
  xmin=-125, xmax=125, ymin=24, ymax=72, xtick={-80,-20,40,100}, xticklabels={06:00,07:00,08:00,09:00}, ytick={30,40,50,60,70},
  title style={font=\footnotesize, yshift=-0.9ex, text depth=0.5ex}, ymajorgrids, grid style={black!9}, axis line style={black!35}, tick style={black!35},
  tick label style={font=\footnotesize}, label style={font=\footnotesize}, xlabel style={yshift=2pt}, ylabel style={yshift=-2pt}]
\nextgroupplot[title={15 minutes ahead}, xlabel={Time of day}, ylabel={Speed (mph)}]
\addplot[black, line width=0.9pt, forget plot] coordinates {(-120,65.9) (-115,65.2) (-110,65.9) (-105,65.7) (-100,65.7) (-95,65.5) (-90,65.1) (-85,65.1) (-80,64.3) (-75,64.0) (-70,64.7) (-65,64.5) (-60,64.0) (-55,64.2) (-50,63.8) (-45,64.3) (-40,64.4) (-35,64.0) (-30,64.8) (-25,65.2) (-20,62.6) (-15,58.4) (-10,51.7) (-5,48.7) (0,43.8) (5,38.3) (10,36.2) (15,34.1) (20,32.7) (25,32.2) (30,31.9) (35,32.1) (40,31.8) (45,32.1) (50,31.9) (55,31.5) (60,31.7) (65,31.6) (70,31.5) (75,31.8) (80,31.8) (85,32.1) (90,32.0) (95,31.9) (100,31.7) (105,31.8) (110,31.7) (115,31.4) (120,32.6)};
\addplot[black!58, line width=0.9pt, dash pattern=on 2.6pt off 1.5pt, forget plot] coordinates {(-120,66.2) (-115,66.2) (-110,66.1) (-105,66.1) (-100,65.9) (-95,66.1) (-90,66.0) (-85,66.1) (-80,65.8) (-75,65.5) (-70,65.5) (-65,65.2) (-60,65.0) (-55,65.2) (-50,65.0) (-45,64.6) (-40,64.6) (-35,64.3) (-30,64.5) (-25,64.6) (-20,64.4) (-15,64.7) (-10,64.9) (-5,64.2) (0,61.1) (5,53.1) (10,51.7) (15,49.7) (20,42.8) (25,43.8) (30,42.6) (35,42.1) (40,41.8) (45,41.6) (50,42.1) (55,40.9) (60,42.1) (65,39.5) (70,39.6) (75,39.3) (80,40.6) (85,39.5) (90,38.9) (95,38.2) (100,37.8) (105,35.7) (110,35.1) (115,34.5) (120,34.5)};
\addplot[cofacolor, line width=1.15pt, forget plot] coordinates {(-120,66.2) (-115,66.1) (-110,65.9) (-105,66.0) (-100,65.7) (-95,65.9) (-90,65.8) (-85,65.8) (-80,65.8) (-75,65.5) (-70,65.3) (-65,64.8) (-60,64.4) (-55,64.7) (-50,64.6) (-45,64.3) (-40,64.3) (-35,64.1) (-30,64.3) (-25,64.3) (-20,64.1) (-15,64.4) (-10,64.6) (-5,63.6) (0,59.2) (5,49.7) (10,49.0) (15,43.0) (20,37.0) (25,37.0) (30,34.6) (35,33.2) (40,32.9) (45,32.8) (50,33.9) (55,34.0) (60,34.5) (65,33.3) (70,33.1) (75,32.5) (80,32.4) (85,32.1) (90,32.2) (95,32.4) (100,32.5) (105,32.9) (110,32.1) (115,32.2) (120,32.2)};
\nextgroupplot[title={30 minutes ahead}, xlabel={Time of day}, yticklabels=\empty]
\addplot[black, line width=0.9pt, forget plot] coordinates {(-120,65.9) (-115,65.2) (-110,65.9) (-105,65.7) (-100,65.7) (-95,65.5) (-90,65.1) (-85,65.1) (-80,64.3) (-75,64.0) (-70,64.7) (-65,64.5) (-60,64.0) (-55,64.2) (-50,63.8) (-45,64.3) (-40,64.4) (-35,64.0) (-30,64.8) (-25,65.2) (-20,62.6) (-15,58.4) (-10,51.7) (-5,48.7) (0,43.8) (5,38.3) (10,36.2) (15,34.1) (20,32.7) (25,32.2) (30,31.9) (35,32.1) (40,31.8) (45,32.1) (50,31.9) (55,31.5) (60,31.7) (65,31.6) (70,31.5) (75,31.8) (80,31.8) (85,32.1) (90,32.0) (95,31.9) (100,31.7) (105,31.8) (110,31.7) (115,31.4) (120,32.6)};
\addplot[black!58, line width=0.9pt, dash pattern=on 2.6pt off 1.5pt, forget plot] coordinates {(-120,66.1) (-115,66.0) (-110,66.0) (-105,66.0) (-100,66.1) (-95,66.2) (-90,66.1) (-85,65.9) (-80,65.9) (-75,65.8) (-70,65.9) (-65,65.9) (-60,65.7) (-55,65.5) (-50,65.2) (-45,64.8) (-40,64.8) (-35,64.7) (-30,64.4) (-25,64.4) (-20,64.3) (-15,64.5) (-10,64.5) (-5,64.4) (0,64.5) (5,64.7) (10,64.2) (15,63.3) (20,58.5) (25,58.9) (30,58.7) (35,53.7) (40,56.3) (45,55.7) (50,55.6) (55,54.7) (60,54.3) (65,54.4) (70,52.9) (75,54.2) (80,50.8) (85,51.9) (90,51.1) (95,53.4) (100,51.9) (105,50.7) (110,49.2) (115,47.8) (120,42.9)};
\addplot[cofacolor, line width=1.15pt, forget plot] coordinates {(-120,66.0) (-115,66.0) (-110,66.0) (-105,66.1) (-100,66.1) (-95,66.0) (-90,66.0) (-85,65.8) (-80,65.9) (-75,65.8) (-70,65.7) (-65,65.7) (-60,65.5) (-55,65.3) (-50,64.9) (-45,64.5) (-40,64.7) (-35,64.6) (-30,64.3) (-25,64.2) (-20,64.0) (-15,64.2) (-10,64.3) (-5,64.1) (0,64.3) (5,64.5) (10,63.8) (15,59.9) (20,50.4) (25,50.9) (30,45.0) (35,39.7) (40,41.0) (45,38.2) (50,35.6) (55,34.8) (60,34.7) (65,36.5) (70,36.1) (75,36.6) (80,35.6) (85,35.5) (90,34.6) (95,34.4) (100,33.9) (105,33.9) (110,33.9) (115,34.0) (120,34.2)};
\end{groupplot}
\begin{scope}[shift={(0.15,3.22)}, font=\footnotesize]
\draw[black, line width=0.9pt] (0,0) -- (0.45,0); \node[anchor=west, inner sep=0pt] at (0.52,0) {Measured};
\draw[black!58, line width=0.9pt, dash pattern=on 2.6pt off 1.5pt] (2.10,0) -- (2.55,0); \node[anchor=west, inner sep=0pt] at (2.62,0) {Single forecaster};
\draw[cofacolor, line width=1.15pt] (5.35,0) -- (5.80,0); \node[anchor=west, inner sep=0pt] at (5.87,0) {DeCo};
\end{scope}
\end{tikzpicture}
\caption{Speed at one PeMS08 sensor during a congestion onset in the test period. Each forecast is drawn at the time it refers to.}
\label{fig:case}
\end{figure}

%% file: sections/conclusion.tex
\section{Conclusion}
\label{sec:conclusion}

This paper presented DeCo, which trains a forecaster for each of flow, occupancy, and speed and coordinates their forecasts at each sensor with a module trained on out-of-fold forecasts. We extended PeMS03 and PeMS07 to three variables on their original sensors and released the aligned data, so that all four standard PeMS benchmarks support three-variable forecasting. DeCo attains the lowest MAE on every dataset and variable and is up to 4.9\% below RAGC, the strongest baseline. The coordination module lowers the error on every dataset and variable. With one forecaster per variable, cross-variable correction improves on target-only correction in every setting, and its largest speed gains fall inside congestion. Coordinating the variables after forecasting gains more than fusing them in the input. Future work will extend DeCo to longer horizons~\cite{yin2025xxltraffic}, other road networks and transport modes~\cite{yin2023mtmgnn,ma2022multimodal}, sensor sets that change over time~\cite{yin2026evoxxltraffic}, probabilistic forecasts~\cite{yin2024curriculum}, and coordination across sensors. Stage~1 can also adopt other forecasters, such as models built on decomposition or normalization~\cite{deng2024scnn,deng2024sscnn,deng2023ttsnorm}.